\documentclass[final,5p,times]{elsarticle}

\usepackage{graphicx}
\usepackage{amssymb}
\usepackage{amsthm}
\usepackage{amsmath}
\usepackage{lineno}
\usepackage{subfigure}
\usepackage{multirow}
\usepackage{booktabs}
\usepackage[colorinlistoftodos]{todonotes}
\usepackage{url}

\newcommand{\review}[1]{\textcolor{black}{#1}}

\usepackage{scrlayer-scrpage}
\clearscrheadfoot
\begin{document}

\begin{frontmatter}

\title{A study on topological descriptors for the analysis of 3D surface texture}


\author[mm]{Matthias Zeppelzauer}
\ead{m.zeppelzauer@fhstp.ac.at}
\author[bm]{Bartosz Zieli\'{n}ski}
\ead{bartosz.zielinski@uj.edu.pl}
\author[bm]{Mateusz Juda}
\author[mm]{Markus Seidl}

\address[mm]{Media Computing Group, Institute of Creative Media Technologies,\\
St. P\"{o}lten University of Applied Sciences,\\
Matthias-Corvinus Strasse 15, 3100 St. P\"{o}lten, Austria}
\address[bm]{The Institute of Computer Science and Computer Mathematics,\\ Faculty of Mathematics and Computer Science, Jagiellonian University,\\
ul. {\L{}}ojasiewicza 6, 30-348 Krak{\'o}w, Poland}

\begin{abstract}
Methods from computational topology are becoming more and more popular in computer vision and have shown to improve the state-of-the-art in several tasks. In this paper, we investigate the applicability of topological descriptors in the context of 3D surface analysis for the classification of different surface textures. We present a comprehensive study on topological descriptors, investigate their robustness and expressiveness and compare them with state-of-the-art methods including Convolutional Neural Networks (CNNs). Results show that class-specific information is reflected well in topological descriptors. The investigated descriptors can directly compete with non-topological descriptors and capture complementary information. As a consequence they improve the state-of-the-art when combined with non-topological descriptors. 
\end{abstract}


\end{frontmatter}



\section{Introduction}

In recent years, methods for sparse and dense reconstruction of 3D scenes have progressed strongly due to the availability of inexpensive off-the-shelf hardware like Microsoft Kinect and the development of robust 3D reconstruction algorithms (e.g. structure from motion techniques, SfM)~\cite{crandall2011,wu2013}. Thus, the amount of available 3D data is rising constantly. Furthermore, the reconstruction accuracy is increasing strongly, which enables 3D reconstructions with sub-millimeter resolution~\cite{wohlfeil2013}. The high resolution enables the accurate description of the geometric micro-structure of a surface, which opens up new opportunities for search and retrieval in 3D scenes, such as the recognition of objects by their specific surface properties as well as the distinction of different types of materials for improved scene understanding.

The geometric micro-structure determines the haptic appearance of a surface (e.g. in terms of roughness, waviness, and lay) and is also referred to as \emph{surface texture}\footnote{Note that surface texture is different from the \emph{apparent texture} or \emph{visual texture} of a surface which refers to the visual (radiometric) appearance of the surface in terms of color variations, brightness, and reflectivity~\cite{tuceryan1998}.}~\cite{ansiSurfaceTexture1996,blunt2003,barcelo2012}. The geometric structure of a surface is closely related to its topology and thus, topological descriptors are promising candidates for its representation. In this article, we investigate the capabilities of topological analysis of describing and classifying 3D surfaces according to their geometric micro-structure and present a first extensive study on topological descriptors in this context.

\begin{figure}%
\centering
	\subfigure[]{\label{sfig:pc}
	\includegraphics[width=0.98\linewidth]{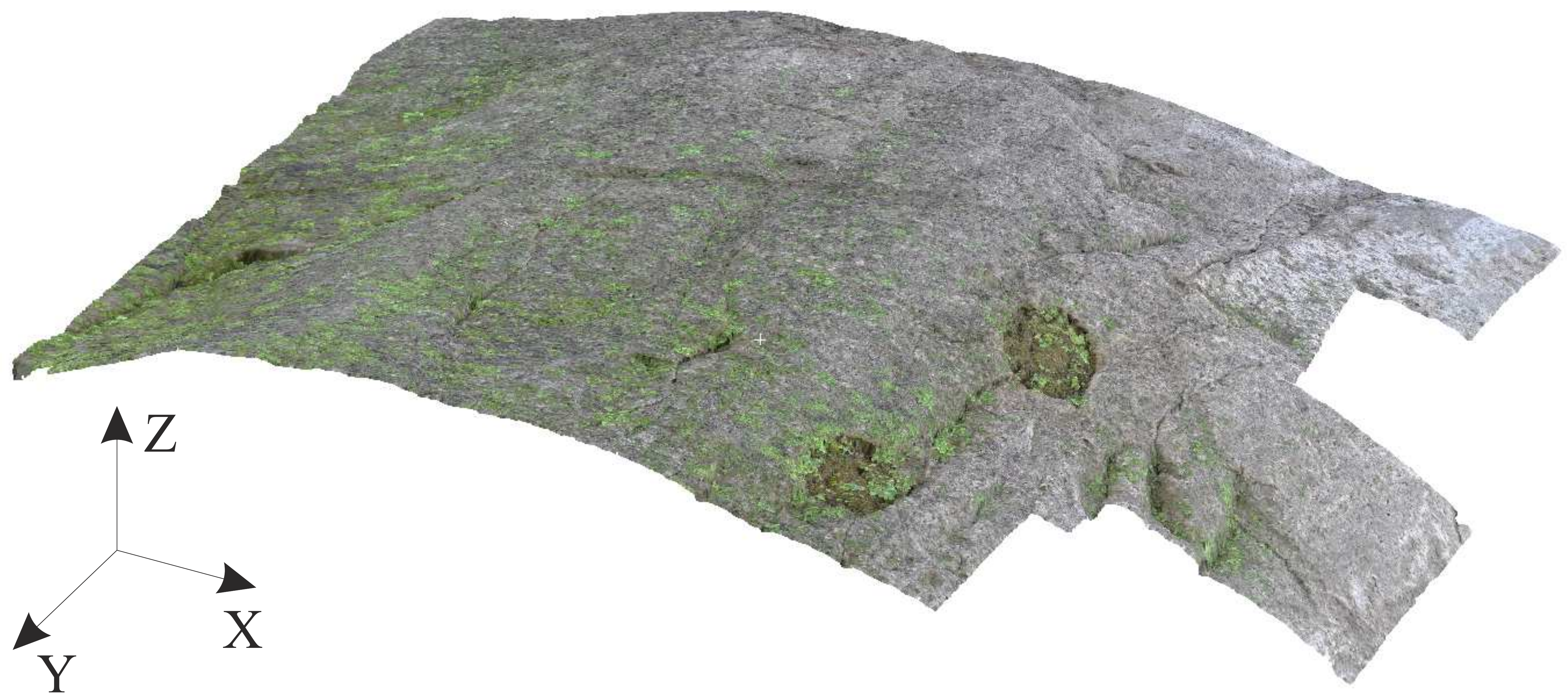}}
	\subfigure[]{\label{sfig:flatDM}
	\includegraphics[width=0.45\linewidth]{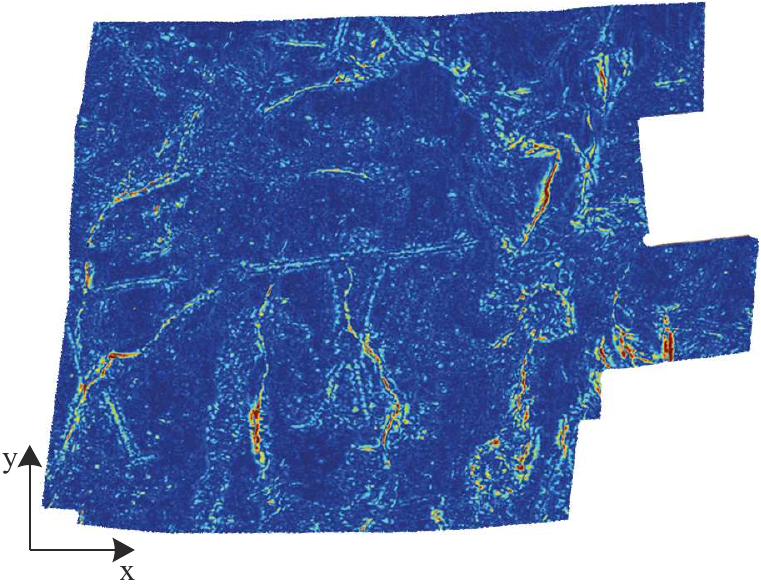}}
	\subfigure[]{\label{sfig:anno}
	\includegraphics[width=0.45\linewidth]{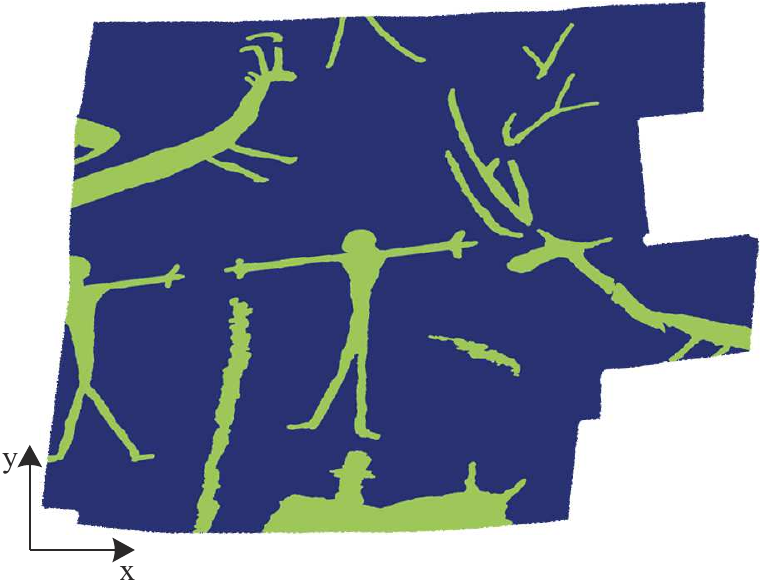}}
	\caption{3D surface data: (a) the 3D point cloud of the surface; (b) the depth projection of the surface; (c) ground truth labeling that specifies areas with different surface texture (blue and green).}
	\label{fig:depthMap}
\end{figure}

For our study we employ high-resolution 3D reconstructions of natural rock surfaces from the archaeological domain, see Fig.~\ref{fig:depthMap}a which exhibit different surface textures. We map the surfaces to depth maps (Fig.~\ref{fig:depthMap}b)~\cite{zeppelzauer2015efficient} and analyze the maps in a patch-wise manner. For each surface patch we extract topological descriptors and traditional non-topological descriptors. Next, we train a classifier and try to predict the class of texture from each patch. The result is a class label for each surface patch. We employ precise maps with ground-truth labels (see Fig.~\ref{fig:depthMap}c) to evaluate the classification accuracy.

This article is an extended version of the work in~\cite{zeppelzauer2016topological} where we reported first preliminary results for topological surface texture analysis. 
This article significantly extends this work to a comprehensive and in-depth study on topological descriptors for 3D surface analysis. We evaluate different extensions and variants to the proposed topological descriptors and study in detail their robustness and discriminativity. \review{In particular the contribution of this paper over the previous one is an investigation of}: (i) the sensitivity of the topological descriptors to their computational parameters; (ii) the influence of different normalization variants; (iii) the role of outliers in persistence diagrams~\cite{fasy2014confidence}; (iv) the extension of topological descriptors with additional pre-filtering operations (v) the robustness and the explanatory power of the descriptors and (vi) strategies to reduce the dimension of the descriptors' feature vectors. Moreover, we verify the validity of our results on an extended large-scale dataset \cite{poier2016petrosurf3d}. \review{To provide the complete picture of our analysis we include the results from~\cite{zeppelzauer2016topological} in this paper.} 

Our study reveals that topological descriptors are robust 3D surface descriptors which can compete and partly outperform traditional (non-topolo\-gical) descriptors. Moreover our investigations clearly show that topological descriptors capture complementary information to non-topological descriptors and that this information is \emph{necessary} to obtain peak performance in classification. Remarkably, our results on the extended dataset show that our combinations of topological and non-topological descriptors yield a similar performance level (and partly even higher performance) as deep learning in \cite{poier2016petrosurf3d}.

The remaining article is organized as follows. In Section~\ref{sec:RW} we review related work on 3D surface analysis and relevant topological approaches in this context. Section~\ref{sec:topoApproach} introduces the investigated topological descriptors together with their properties and outlines our topological approach for surface texture analysis. Section~\ref{sec:setup} summarizes the experimental setup and Section~\ref{sec:experimentalResults} presents the study on topological descriptors together with the obtained results. We summarize the major conclusions of our study in Section~\ref{sec:concl}.

\section{Related work}
\label{sec:RW}

Approaches for 3D surface analysis can be partitioned into two classes: methods that operate in 3D space directly and methods that operate on derived image-space representations. The following sections review related work on both approaches.

\subsection{Surface analysis in 3D space}

The natural domain for the analysis of surfaces is the 3D domain. A popular approach is to analyze the mesh or point cloud of a given surface reconstruction with local 3D surface descriptors that describe the local geometry around a 3D point. Numerous descriptors have been proposed for this purpose in recent years. 
Johnson and Hebert propose \emph{spin images} as local descriptors for the dense description of meshes for object recognition in 3D scenes~\cite{johnson1999using}. Darom and Keller extract SIFT features from local depth images of a points' neighborhood to model the surface geometry around it~\cite{darom2012scale}. Zaharescu et al. introduce a local surface descriptor for meshes (MeshHOG) that can be computed from an arbitrary scalar function (e.g. curvature) defined over the surface~\cite{zaharescu2009surface}. Steder et al. propose the \emph{normal-aligned radial feature} (NARF) which captures local depth variations in range data~\cite{steder2011point}. Frome et al. extend the well-known 2D \emph{shape context} descriptor~\cite{belongie2002shape} to 3D point clouds (3DSC) and show that it outperforms spin images~\cite{frome_recognizing_2004}. Rusu et al. propose two local 3D point cloud descriptors, namely \emph{persistent point feature histogram} (PFH)~\cite{rusu_persistent_2008} and an accelerated version \emph{fast PFH} (FPFH)~\cite{rusu_fast_2009}. Both build upon the relations between \textit{surfels}, i.e. the combination of a point and its surface normal to describe the surface geometry~\cite{wahl_surflet-pair-relation_2003}. Tombari et al. propose a 3D descriptor (SHOT) based on the point normals that is defined in a robust local reference frame~\cite{tombari_unique_2010}. 

Local descriptors are particularly well-suited for the \emph{sparse} description of a surface, i.e. descriptors are computed at salient feature points only. The analysis of 3D surfaces by their surface texture - as investigated in this paper -- requires the \emph{dense} analysis and representation of a surface. Previous experiments have shown that local 3D descriptors are impractical for this purpose, especially due to computational reasons, because descriptors have to be computed at each surface point to obtain a dense representation, which becomes highly demanding when a point cloud has several millions of points~\cite{zeppelzauer2015efficient}.

\subsection{Surface analysis in image-space}

To circumvent the problems raised by 3D descriptors in the context of dense surface analysis, Othmani et al. propose to map 3D surfaces to image-space and then apply efficient image analysis techniques there to process the projected surfaces~\cite{othmani2013single}. Usually a 2D depth map is derived from the 3D surface (piece-wise or globally, depending on the amount of curvature of the surface) to obtain an image-space representation. If the surface does not contain self-occlusions, the depth map completely represents the 3D surface, except for inaccuracies or losses in resolution due to rasterization~\cite{zeppelzauer2015efficient}.

The analysis of 3D surface texture in image-space can be considered as texture analysis task on the depth map. Thus, approaches of image texture analysis become applicable. Popular methods for image texture analysis include histograms of vector quantized filter responses~\cite{leung1996detecting} and later generalizations such as the bag-of-visual-words model for textures~\cite{csurka2004visual} and the Fisher vector~\cite{perronnin2007fisher}. Recently, deep learning-based approaches for texture analysis have been introduced for the problem of image texture analysis~\cite{cimpoi2016deep}, which outperform many existing methods.

An alternative approach to texture analysis (both for image texture and surface texture) is to consider the underlying data (depth map in our case) as a geometric object and to analyze its topological properties. For this purpose \emph{topological data analysis} (TDA) is a promising approach which currently gains increasing importance in data science. TDA provides methods to study qualitative properties of objects based on the theory of size functions introduced by Frosini et~al.~\cite{frosini1992measuring,Verri1993} and  persistent homology by Edelsbrunner et~al.~\cite{EdLeZo2002}. TDA methods have recently been successfully applied to computer vision problems, such as shape and texture analysis. Li et al. combine the bag-of-features approach with persistence diagrams (PDs) for shape retrieval~\cite{li2014persistence}. Reininghaus et al. proposes an SVM kernel for persistence diagrams to make them compatible with statistical machine learning methods~\cite{reininghaus2014stable}. The authors apply topological descriptors together with the novel kernel to shape retrieval and texture classification in images. Another approach to leverage topological information for machine learning tasks is the persistent image (PI) proposed by Adams et al.~\cite{adams2015persistent}. The PI is a vector-based representation of PD which is built upon earlier work on size functions~\cite{Ferri1997,Frosini1998}. The PI can be employed directly as a feature vector in conventional machine learning techniques and can easily be combined with other descriptors. 

Our work mainly builds upon the approach of Adams et al. In our study we investigate different topological descriptors, including PI, for the domain of surface texture analysis. \review{The difference to the investigations in ~\cite{reininghaus2014stable} and ~\cite{adams2015persistent} is that we try to enrich the PI representation by pre-filtering, normalization, and feature selection and that we analyze in-depth the information captured by the PI descriptor, its redundancy and discriminativity. Furthermore, we investigate the sensitivity of the representation to its computational parameters. Beyond this we combine them with non-topological state-of-the-art descriptors to investigate synergy effects.}

\section{Topological approach}
\label{sec:topoApproach}

\subsection{Overview}
\label{sec:approachOverview}

\begin{figure*} [ht]%
\centering
	\includegraphics[width=0.70\linewidth]{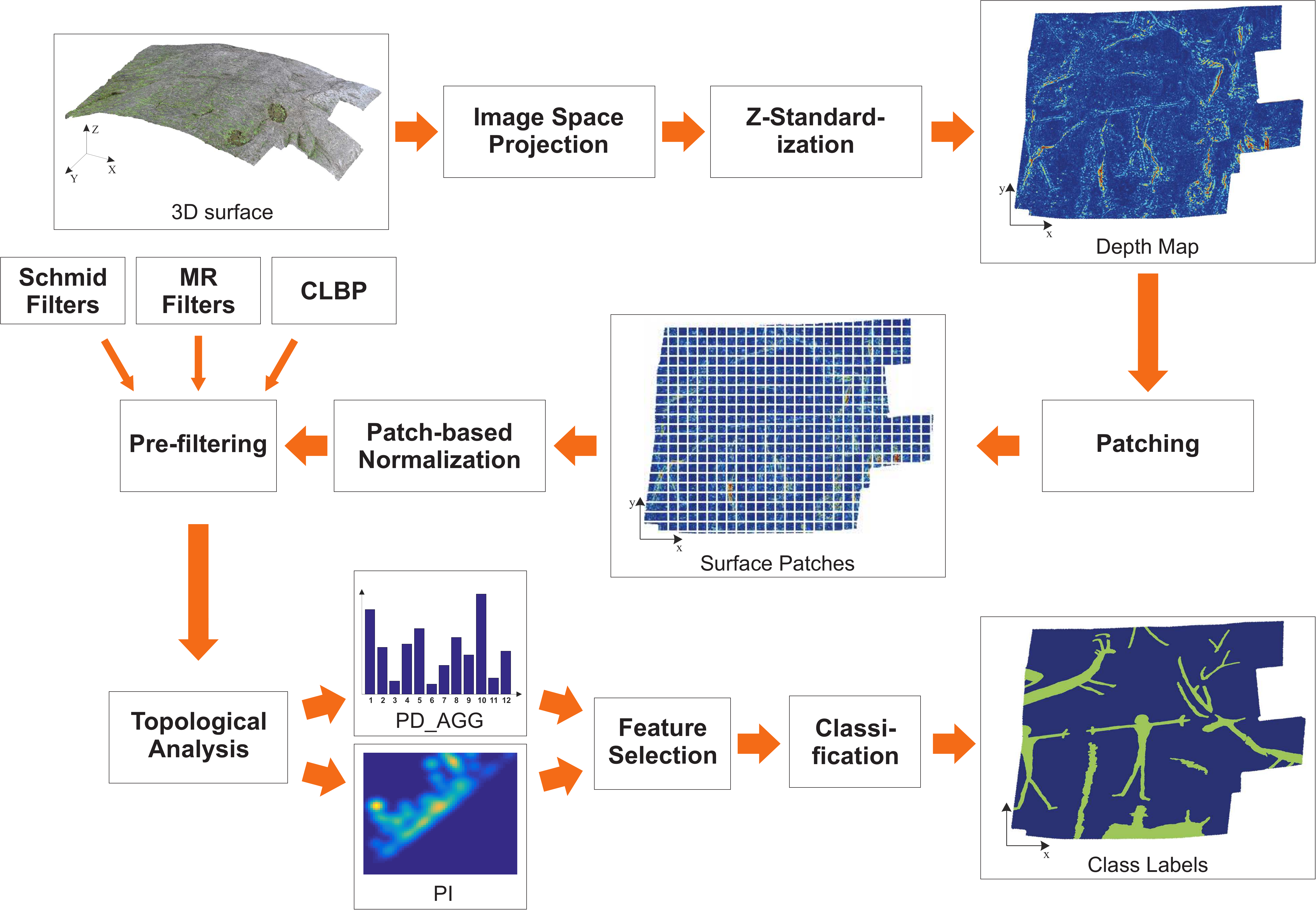}

	\caption{Overview of the proposed approach.}
	\label{fig:overview}
\end{figure*}

Fig.~\ref{fig:overview} provides an overview of the approach that we propose for the analysis of 3D surfaces by persistence homology features. The input 3D surfaces are first projected to image-space to obtain depth maps as proposed in~\cite{zeppelzauer2015efficient}. Next, depth maps are z-standardized (see Section~\ref{sec:normalization}) and split into square patches. Patches are normalized (optional step, see Section~\ref{sec:normalization}) and pre-filtered by Schmid or MR filters or CLBP (optional step, see Section~\ref{sec:preprocessing}). Next, topological descriptors are extracted as described in Sections~\ref{subsec:PDAGG} and~\ref{subsec:PI}. \review{Finally feature selection is performed (optionally) before patch-wise classification to decrease the dimensionality of the feature vector and to reduce redundancies}. The result is a class label for each surface patch, which were mapped back to the surface in Fig.~\ref{fig:overview} for illustration.

\review{In the following, we describe the individual components of the approach and start with its core component, namely the persistent homology.}

\subsection{Persistent homology}
\label{subsec:PDAGG}

By mathematical standards, topology with its 120 years of history, is a relatively young discipline. It grew out of Poincare’s seminal work on the stability of the solar system as a qualitative tool to study the dynamics of differential equations without explicit formulas for solutions~\cite{Poinc1890,Poinc1899,Poinc1895}. Due to the lack of useful analytic methods, topology soon became a purely theoretical discipline. However, in the last few years we observe a rapid development of topological data analysis tools, which open new applications for topology.

Topological spaces appearing in data analysis are typically constructed from small pieces called cells.
A natural tool in the study of multidimensional images with topological methods are hypercubes (points, edges, squares, cubes etc.), e.g. a pixel in a $2$-dimensional image is equivalent to a square, a voxel in a $3$ dimensional volume is equivalent to a cube. Hypercubes are building blocks for structures called cubical complexes. Such representations give topology a combinatorial flavour and make it a natural tool in the study of multi-dimensional data sets.

Intuitively, the rank of the $n$th homology group, the so called $n$th Betti
number denoted by $\beta_n$, counts the number of $n$-dimensional holes in the topological space.
In particular, $\beta_0$ counts the number of connected components.
As an example consider the image of the digit “8”. In this image
there is one connected component and two holes, hence $\beta_0=1$ and $\beta_1 =2$.
For a hollow sphere we have $\beta_0 =1$, $\beta_1 =0$, $\beta_2 =1$.
For a tube in a tire (torus) we have $\beta_0 =1$, $\beta_1 =2$, $\beta_2 =1$.

Betti numbers do not differentiate between small and large holes.
In consequence, the holes resulting from the noise in the data cannot be
distinguished from the holes indicative for the nature of the data.
For instance, in a noisy image of the digit “8” one can get easily $\beta_0 > 1$.
A remedy for this drawback is persistent homology,
a tool invented at the beginning of the 21st century~\cite{EdLeZo2002}.
\review{Persistent homology tracks the holes from birth to death when
the topological space is gradually built by adding cubes in some prescribed order.} 

If $X$ is a cubical complex, one can add cubes step by step.
Typically, the construction goes through different scales, starting from the smallest pieces.
However, in general an arbitrary function $f:X\to\mathbb{R}$, called the Morse function or the measurement function,
may be used to control the order in which the complex is built, starting from low values of $f$ and increasing subsequently.
This way we obtain a sequence of topological spaces, called a filtration,
\[
    \emptyset=X_{r_0}\subset X_{r_1}\subset X_{r_2}\subset\cdots\subset X_{r_n}=X,
\]
where $X_r:=f^{-1}((-\infty,r])$ and $r_i$ is a growing sequence of values of $f$
at which the complex changes.
As the space is gradually constructed, holes are born, persist for some time
and eventually may die. The length of the associated birth-death intervals (persistence intervals) indicates if the holes are relevant or merely noise.
The lifetime of holes is usually visualized by the \emph{persistence diagram} (PD). Persistence diagrams constitute the main tool of topological data analysis. They visualize geometric properties of a multidimensional object $X$ in a simple two-dimensional diagram.

\begin{figure}%
\centering
	\subfigure[]{\label{patch:org}
	\includegraphics[width=0.31\linewidth]{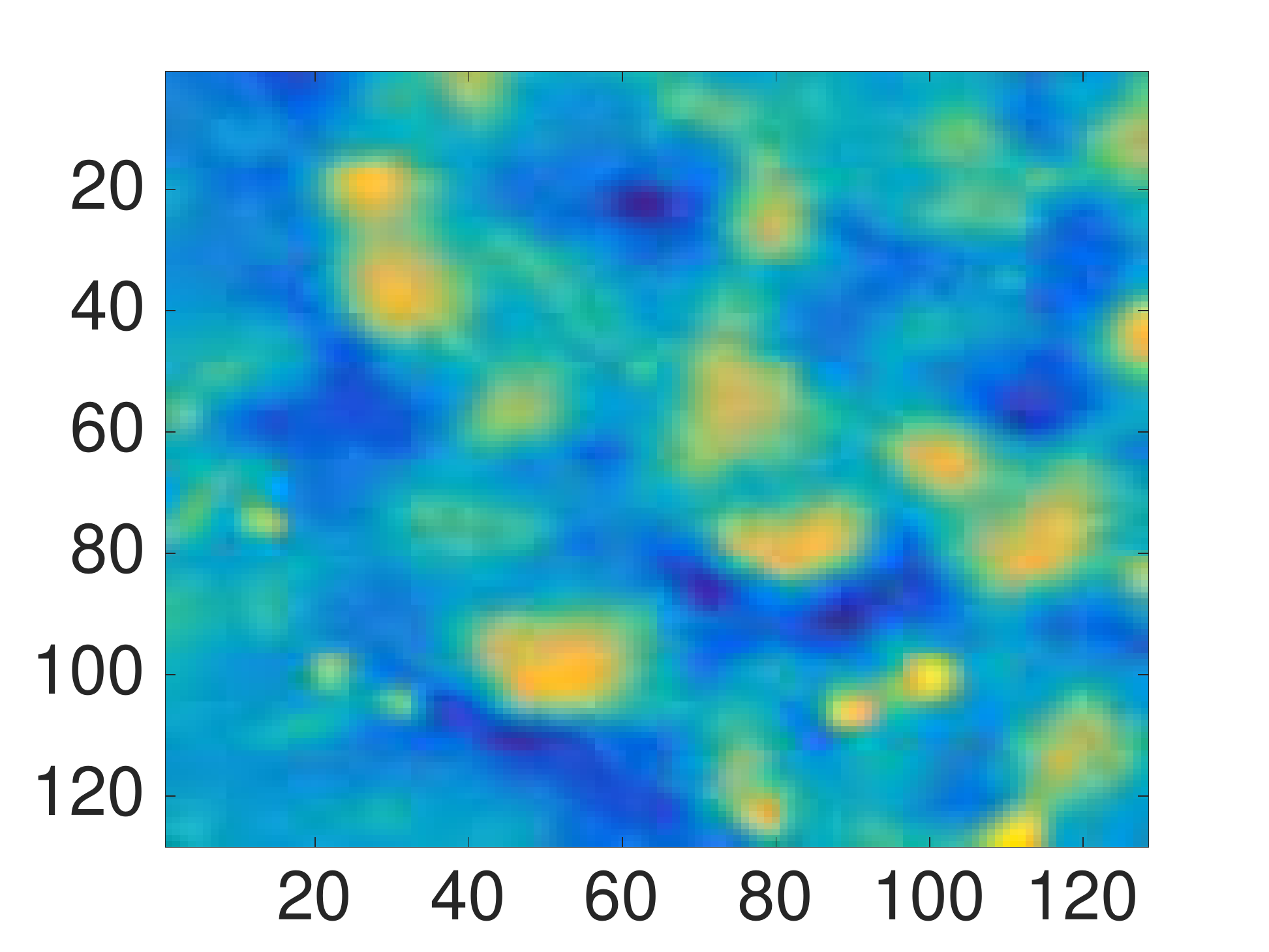}}
	\subfigure[]{\label{patch:pd}
	\includegraphics[width=0.31\linewidth]{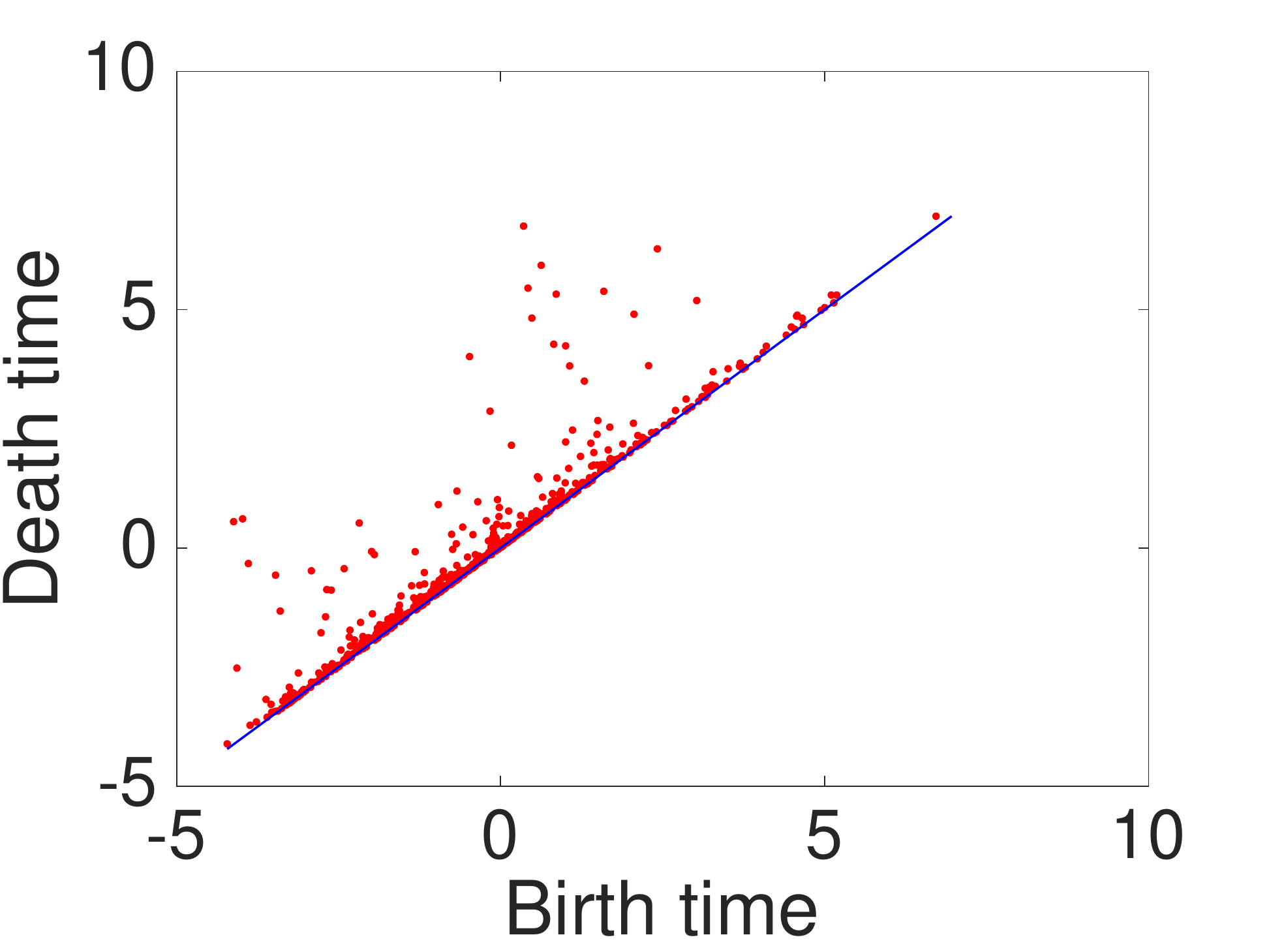}}
	\subfigure[]{\label{patch:pi}
	\includegraphics[width=0.31\linewidth]{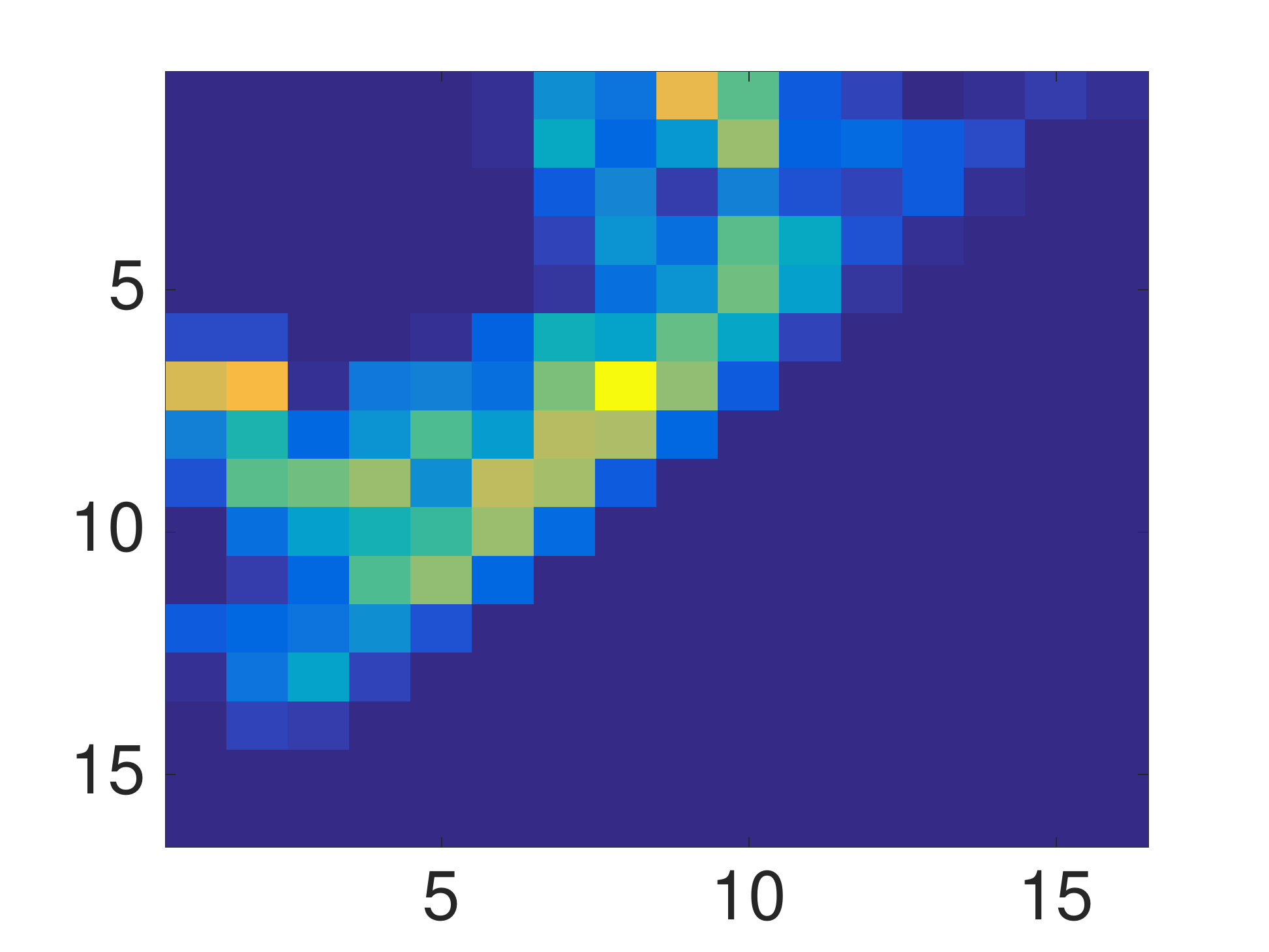}}
	\caption{Example surface patch with derived topological descriptors: (a) the depth map of the original 3D surface; (b) the corresponding persistent diagram; (c) and the persistence image (resolution $16\times16$ pixels, $\sigma_x=\sigma_y=0.001$ and exponential weighting function $g$).}
	\label{fig:patch}
\end{figure}

Fig.~\ref{fig:patch}a shows the depth map of a 3D surface where colors correspond to different depths (blue refers to low depth, yellow to high depth).
In this case pixels are represented as 2-dimensional cells of a cubical complex. For the complex we can obtain a filtration $X_r$ using a measuring function which has a value for a 2-dimensional cube equal to depth (pixel color). For a lower dimensional cell (a vertex or an edge) we can set the function value as a minimum from the higher-dimensional neighborhoods of the cell. Fig.~\ref{fig:patch}b shows the persistence diagram for $X_r$.

The PD is an expressive representation of the surface topography. Unfortunately, the data structure of the PD is complex and cannot easily be combined with traditional machine learning methods which require fixed-length input vectors. This hinders the application of computational topology to computer vision and machine learning tasks. There is still no specific answer on how and when the tools of computational topology and machine learning should be used together.

A first attempt to obtain a fixed-length descriptor of a topological space filtration is to compute elementary statistics of persistence intervals (or equivalently persistence diagrams). Let
\[
   I := \{[b_1,e_1], [b_2,e_2], \ldots, [b_n,e_n] \}
\]
be a set of persistence intervals. Let $D := \{ d_i := (e_i - b_i)\}_{i=1}^n$ be a set of the interval lengths. 
We build an aggregated descriptor of $D$, denoted by PD\_AGG, using following measures:
number of elements, minimum, maximum, mean, standard deviation, variance, $1$-quartile, median, $3$-quartile, and norms $\sum\sqrt{d_i}$, $\sum d_i$, and $\sum(d_i)^2$. The result is a 12-dimensional feature vector.

\subsection{Persistence image}
\label{subsec:PI}
Aside from the PD\_AGG descriptor described above, which can be used with standard classification methods, there are also attempts to use PD directly with appropriately modified classifiers. Reininghaus et al.~\cite{reininghaus2014stable} proposed a multi-scale kernel for PDs, which can be used with a support vector machine (SVM). \review{While this kernel is well-defined in theory, in practice it becomes highly inefficient when the number of training vectors becomes large, as the entire kernel matrix must be computed explicitly (note that this applies to all kernel-based methods). There are alternative methods that aim at representing the PD as a vectorial representation. \review{An example is a persistence landscape (PL)~\cite{bubenik2015statistical} which is a functional representation of a PD. The representation lies in a Banach space and is stable.} Another approach to obtain a vectorial representation from PD is persistent image introduced by Adams et al.~\cite{adams2015persistent}, which according to Makarenko et al. \cite{makarenko2016texture} is especially well-suited for texture data.}. \emph{Persistence image} (PI) is based on concepts known from the theory of size functions~\cite{Ferri1997,Frosini1998}. \review{The method scales well to large training data. In contrast to PL, PI lives in Euclidean space, which makes is directly compatible to a broad set of machine learning techniques.}

The PI is derived by mapping a PD to an integrable function $G_p: \mathbb{R}^2 \rightarrow \mathbb{R}$, which is a sum of Gaussian functions centered at each point of the PD. Taking a discretization of a subdomain of $G_p$ defines a grid. An image can be created by computing the integral of $G_p$ on each grid box, thus defining a matrix of pixel values. Formally, the value of each pixel $p=(x,y)$ within a PI is defined by the following equation:
\begin{equation}
PI(p) = \iint\limits_{p} \sum_{[b_i, e_i] \in I}{g(b_i, e_i) \, \dfrac{1}{2 \pi \sigma_x \sigma_y} \, e^{-\frac{1}{2} \left( \frac{(x - b_i)}{\sigma_x^2} + \frac{(y - e_i)}{\sigma_y^2} \right) }} \,dy\,dx,
\label{eq:PI}
\end{equation}
\noindent
where $g(b_i, e_i)$ is a weighting function, which depends on the distance from the diagonal. Points close to the diagonal are usually considered as noise and therefore get lower weights by the function $g$. Variables $\sigma_x$ and $\sigma_y$ are the standard deviations of the Gaussians in $x$ and $y$ direction. The resulting image (see Fig.~\ref{fig:patch}c) is vectorized to achieve a standardized vectorial representation which is compatible to a broad range of machine learning techniques. The pixels under the diagonal are not taken into account in case of PI (they all are zeros).

\begin{figure}%
\centering
	\subfigure[]{\label{patch:pi:lim}
	\includegraphics[width=0.45\linewidth]{./figures/patch_fg_pi-eps-converted-to.pdf}}
	\subfigure[]{\label{patch:pi:res}
	\includegraphics[width=0.45\linewidth]{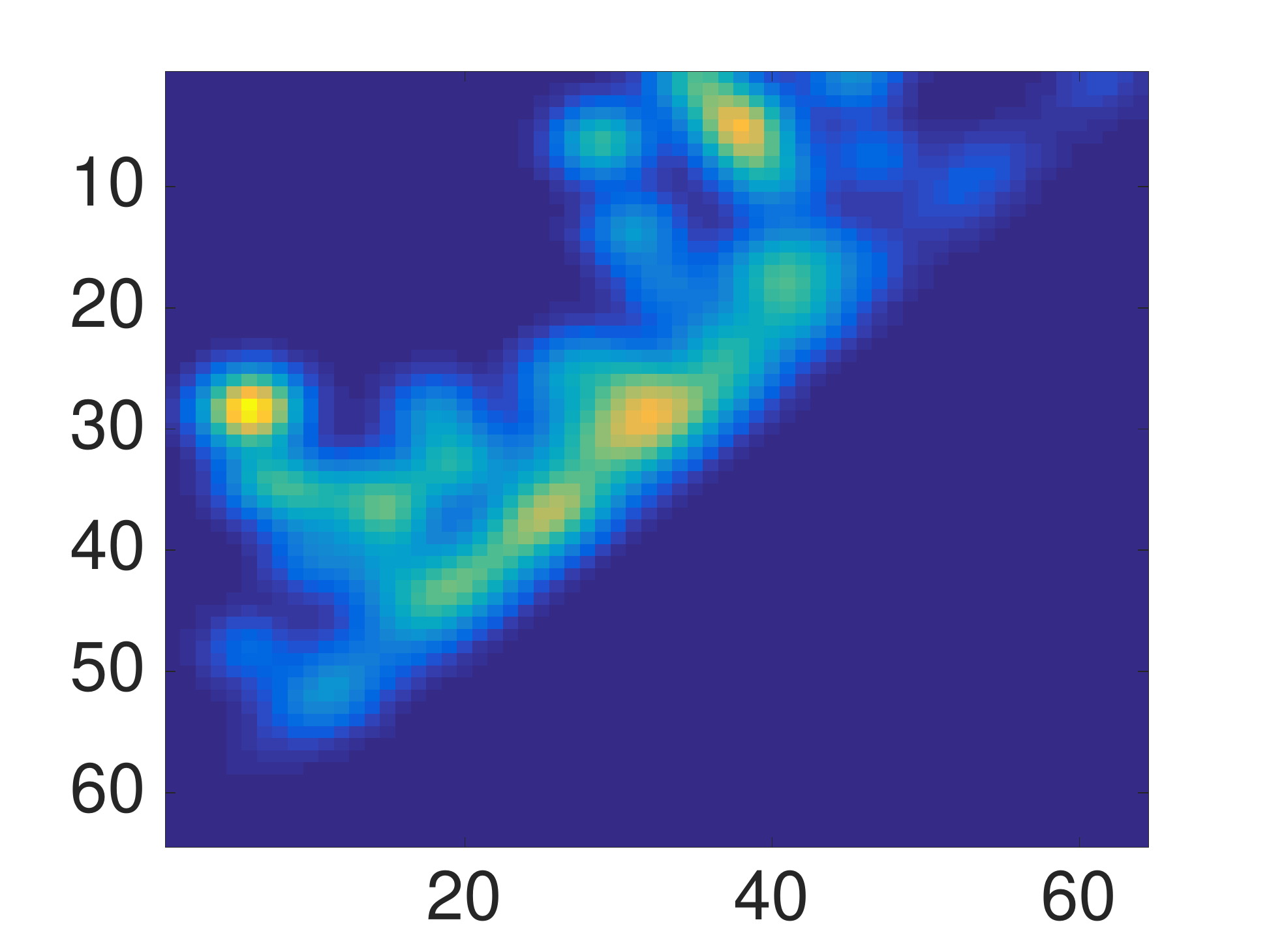}}
    \\
	\subfigure[]{\label{patch:pi:sigma}
	\includegraphics[width=0.45\linewidth]{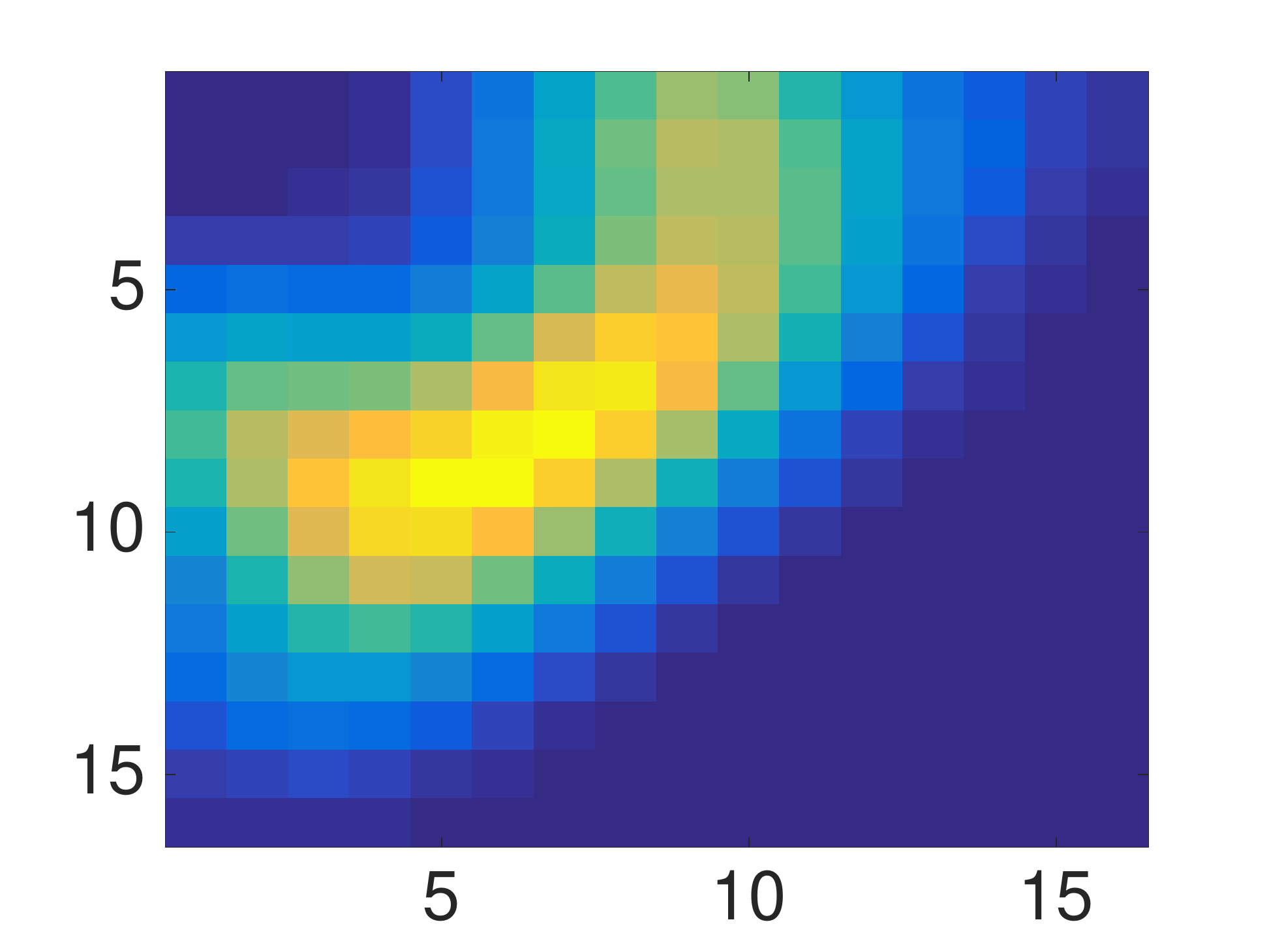}}
	\subfigure[]{\label{patch:pi:noWeight}
	\includegraphics[width=0.45\linewidth]{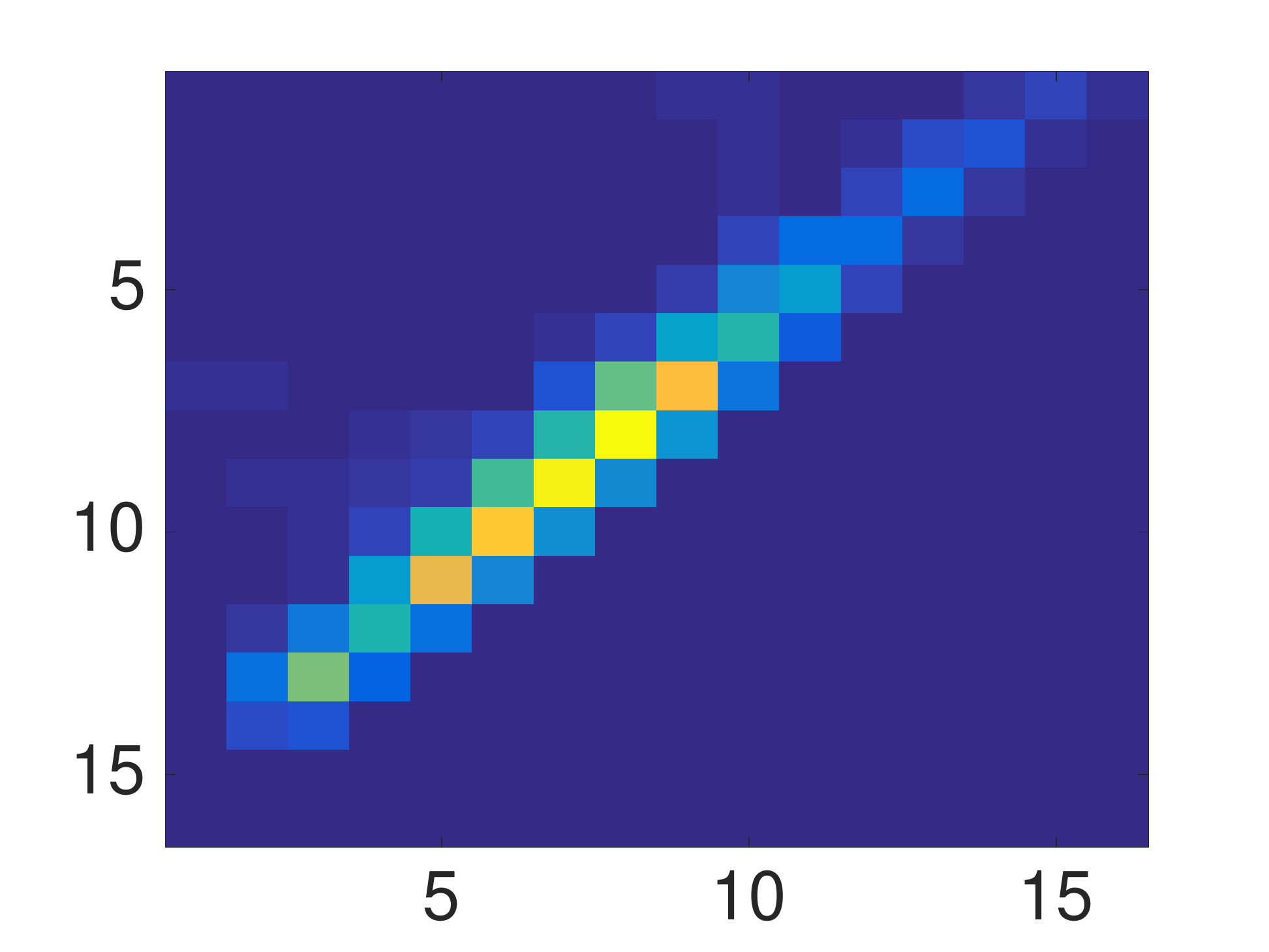}}
	\caption{PIs for patch from Fig.~\ref{fig:patch}a computed with different parameter values: (a) original PI from Fig.~\ref{fig:patch}b; (b) resolution $64\times64$ pixels instead of $16\times16$; (c) $\sigma_x=\sigma_y=0.01$ instead of $0.001$; (d) no weighting function.}
	\label{fig:differentParams}
\end{figure}

The advantage of PIs compared to aggregated PD descriptors is that the spatial information from the PD are preserved, which may lead to higher classification accuracies~\cite{adams2015persistent}. However, the computation of PI requires numerous parameters like the weighting function $g$, $\sigma_x$ and $\sigma_y$ of the employed Gaussians and the final image resolution. Changes in values of the parameters may result in strongly different PIs, which is illustrated in Fig.~\ref{fig:differentParams}. It is a priori not clear which parameters yield the best representation for a specific task and how robust PI is to these parameters in practice.

\subsection{Pre-filtering}
\label{sec:preprocessing}

Different types of pre-processing can be applied to the input data before computing topological descriptors, in order to extract more complex topological information. We take into consideration the Schmid filter bank~\cite{schmid2001constructing} and the Maximum Response (MR) filter bank~\cite{geusebroek2003fast}, as they are very popular and rotation invariant. Moreover, robust texture descriptors such as Local Binary Patterns (LBP) or Complete LBP (CLBP)\cite{guo2010completed} may be applied before topological analysis as frequently observed in related work~\cite{reininghaus2014stable}. 

\begin{figure}%
\centering
	\subfigure[]{\label{patch:schmid}
	\includegraphics[width=0.31\linewidth]{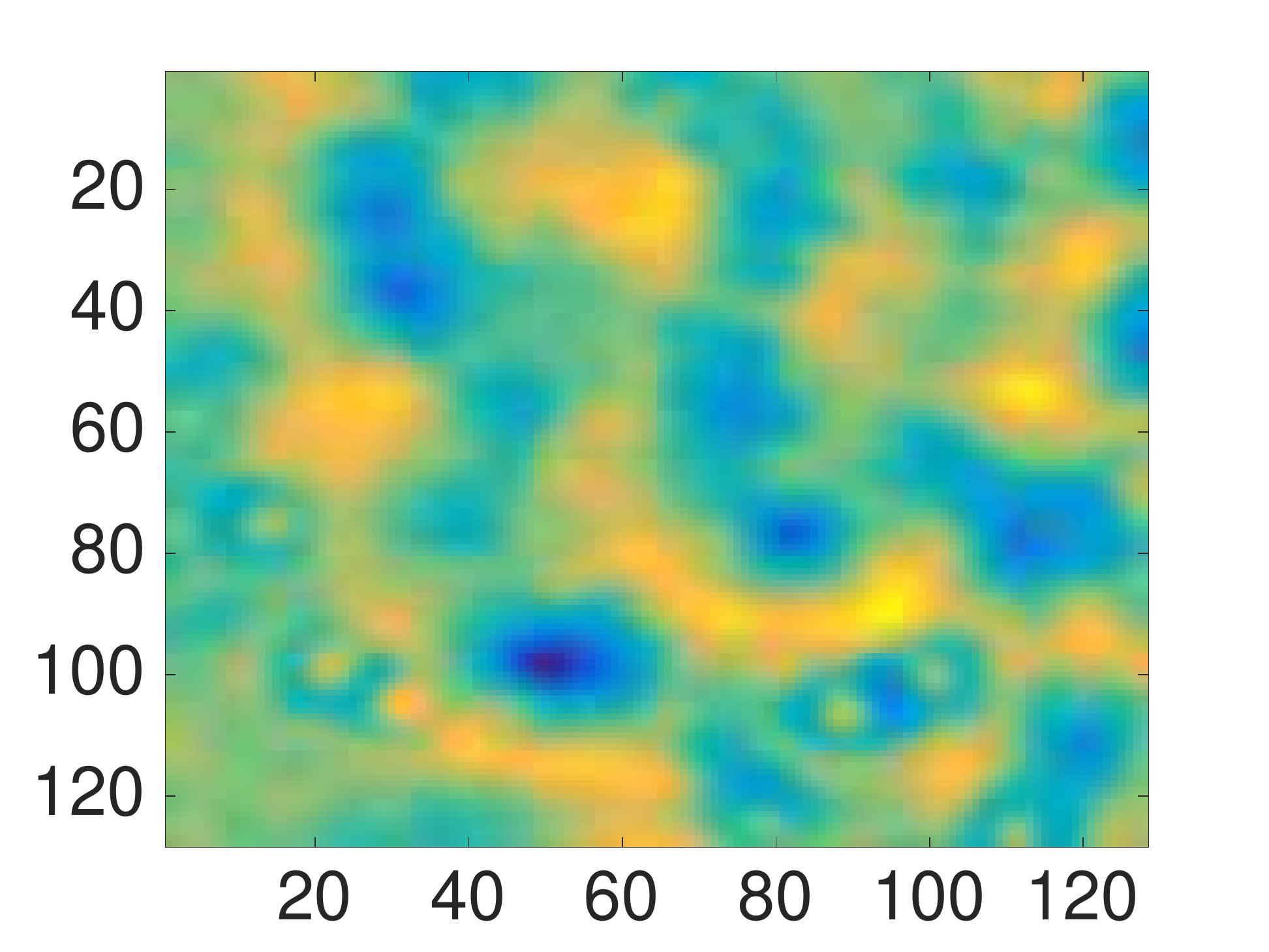}}
	\subfigure[]{\label{patch:mr1}
	\includegraphics[width=0.31\linewidth]{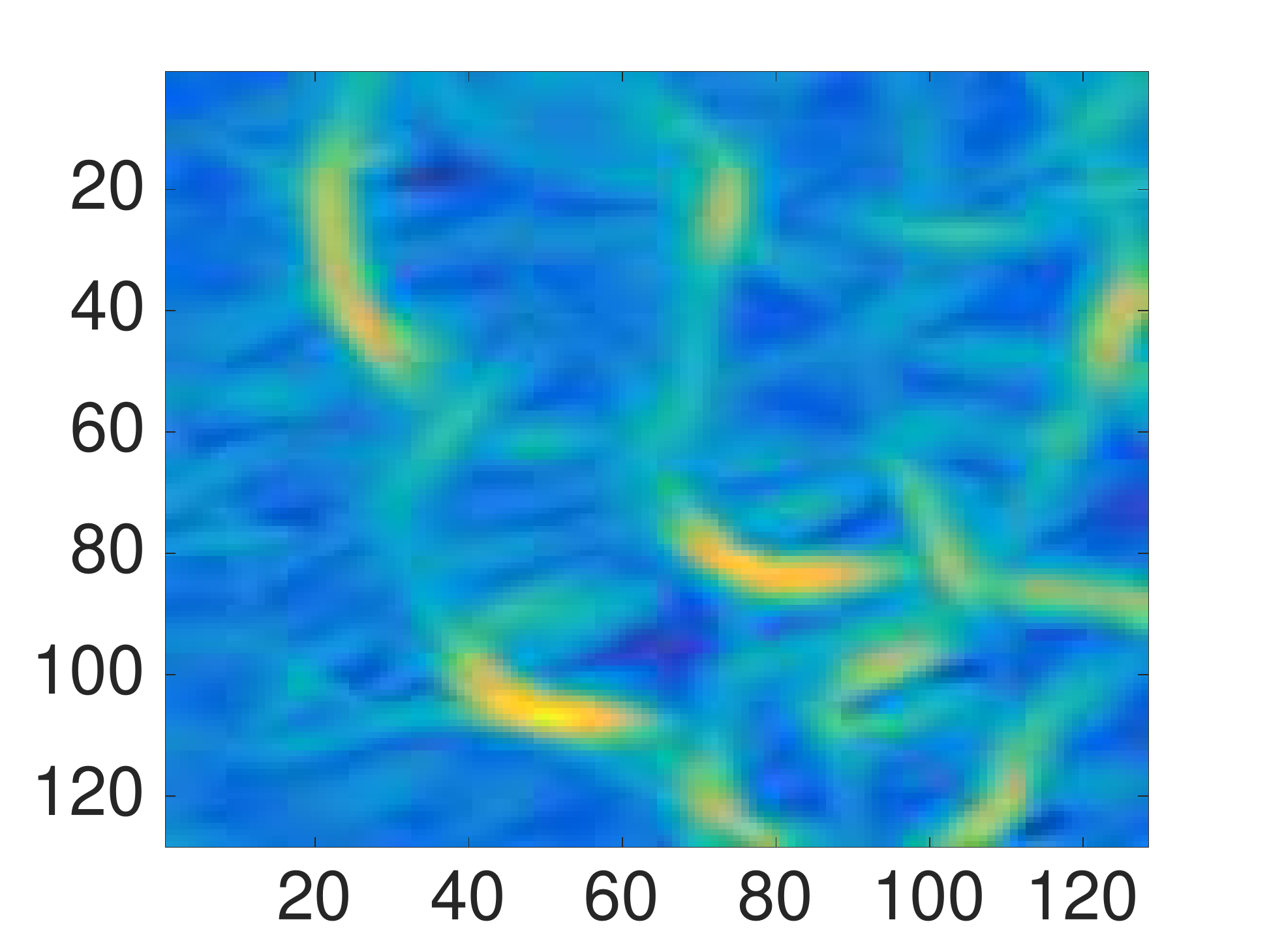}}
	\subfigure[]{\label{patch:clbpS}
	\includegraphics[width=0.31\linewidth]{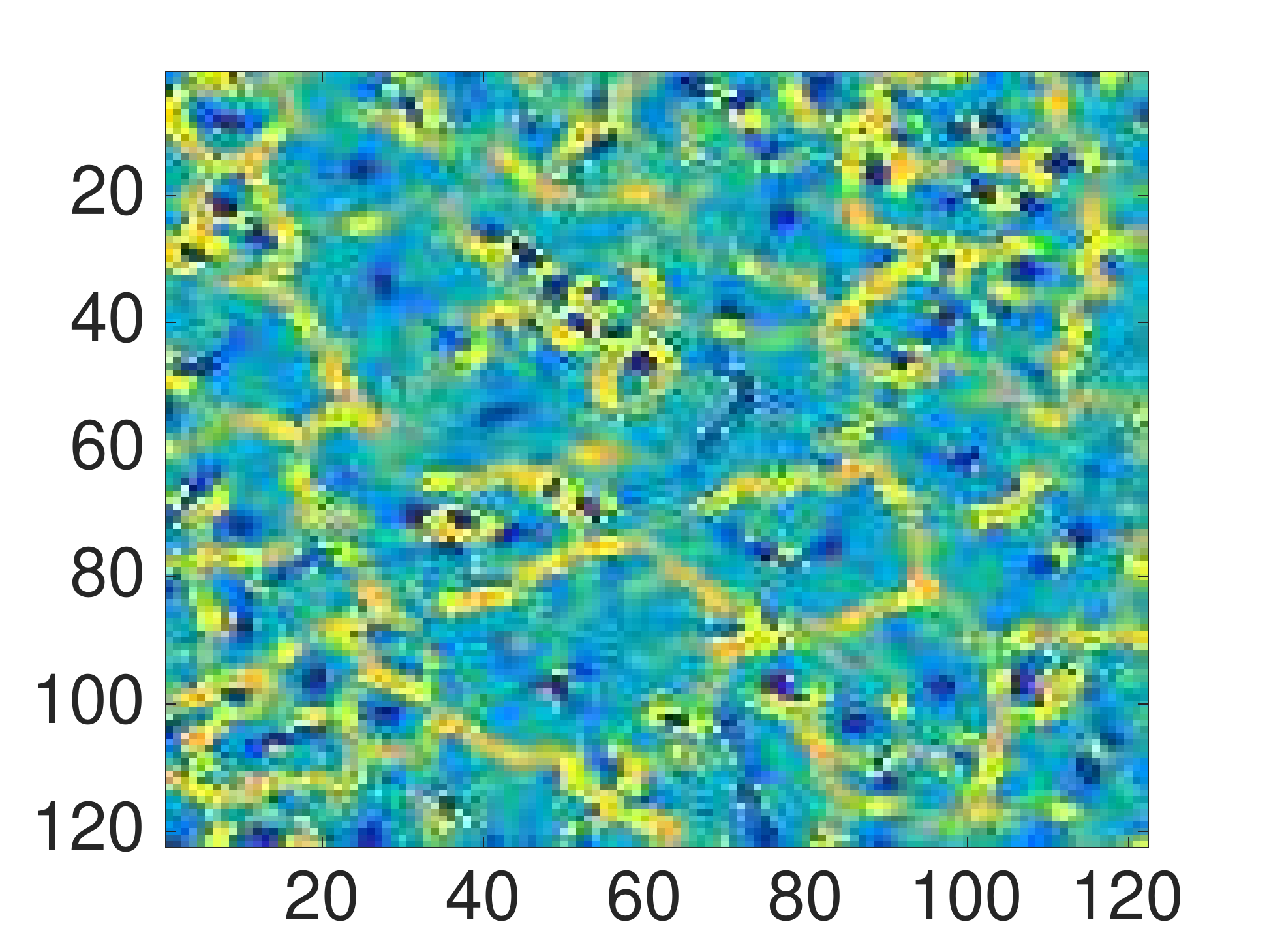}}\\
	\subfigure[]{\label{patch:pi:schmid}
	\includegraphics[width=0.31\linewidth]{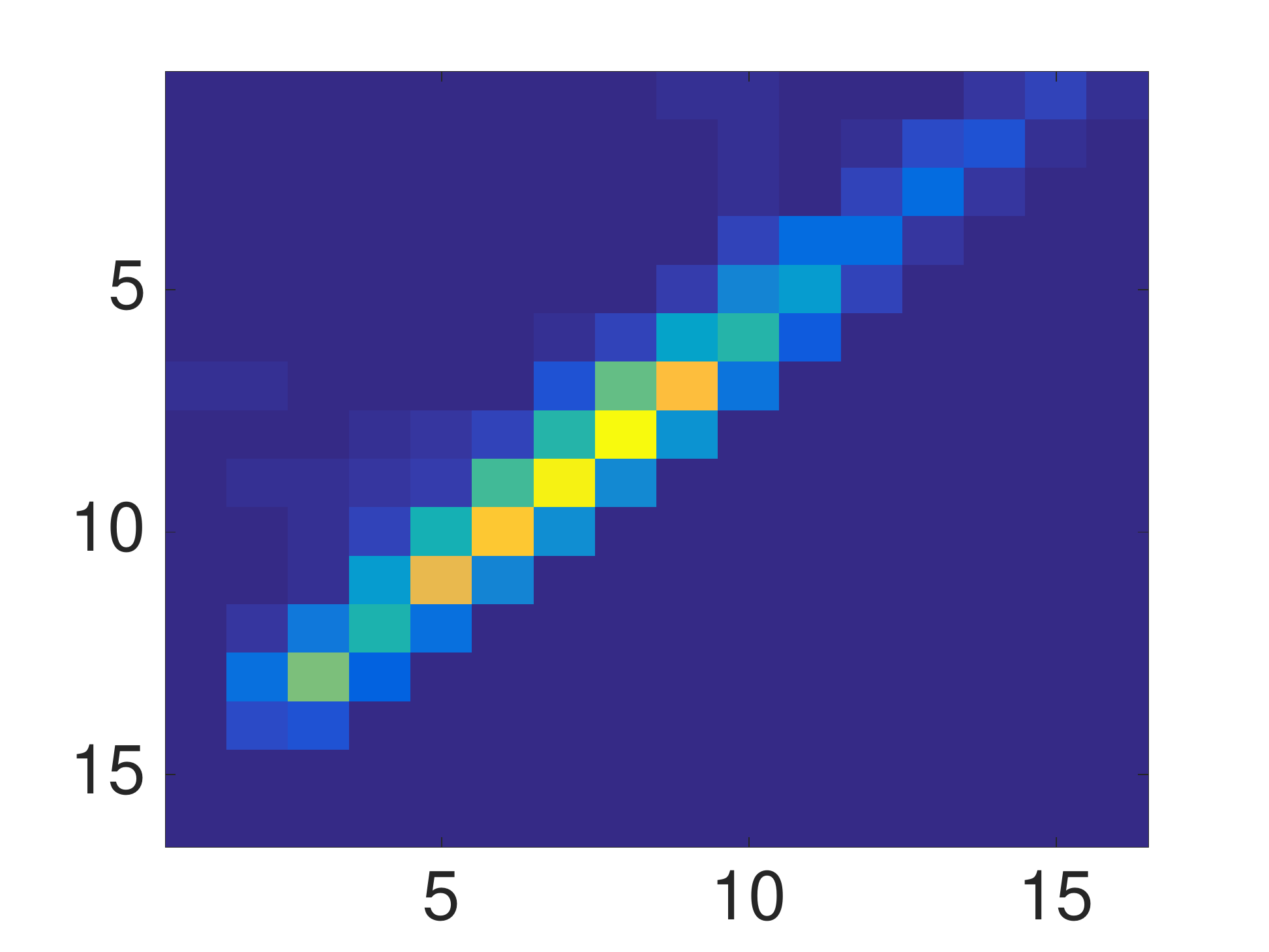}}
	\subfigure[]{\label{patch:pi:mr1}
	\includegraphics[width=0.31\linewidth]{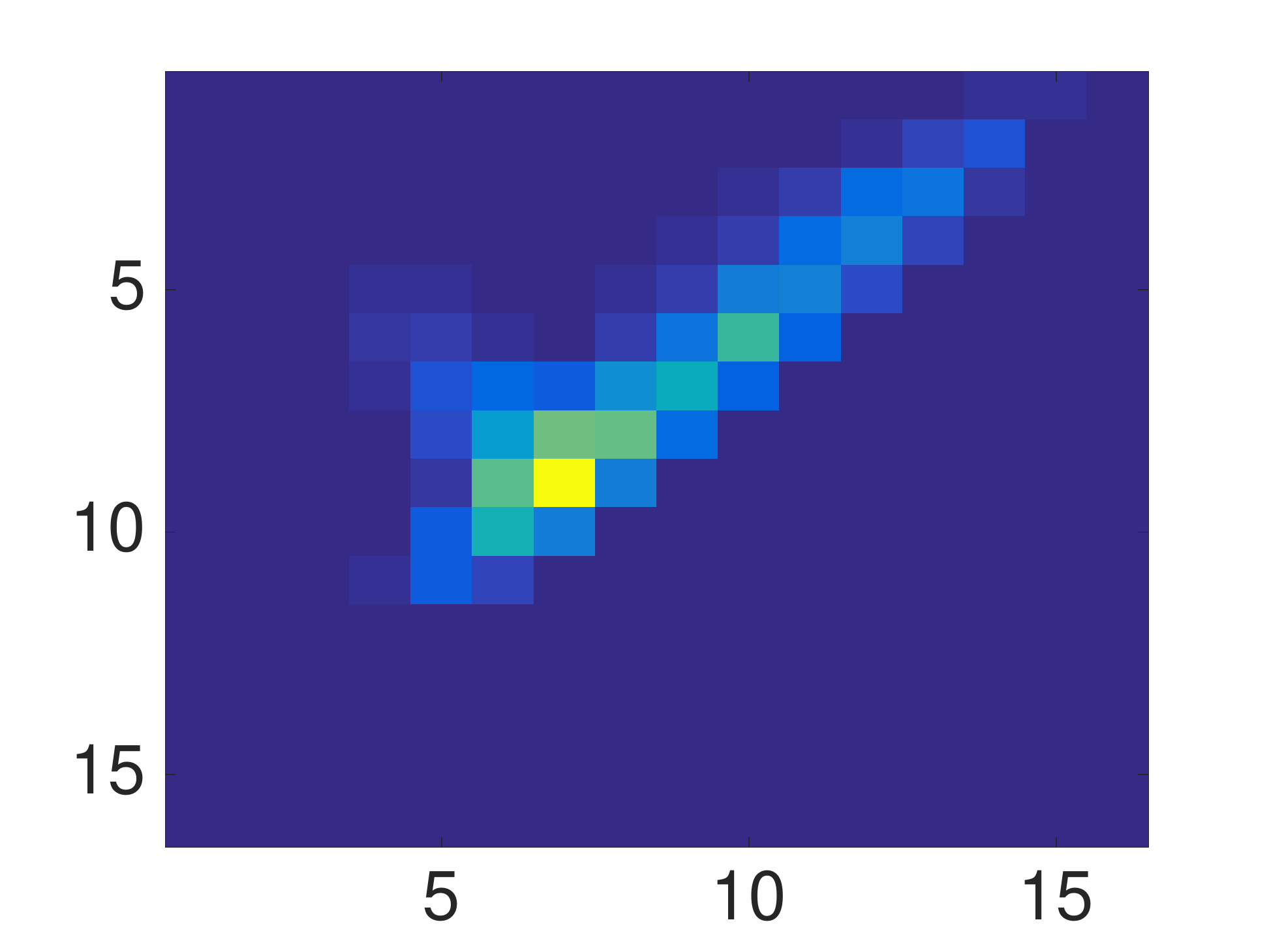}}
	\subfigure[]{\label{patch:pi:clbpS}
	\includegraphics[width=0.31\linewidth]{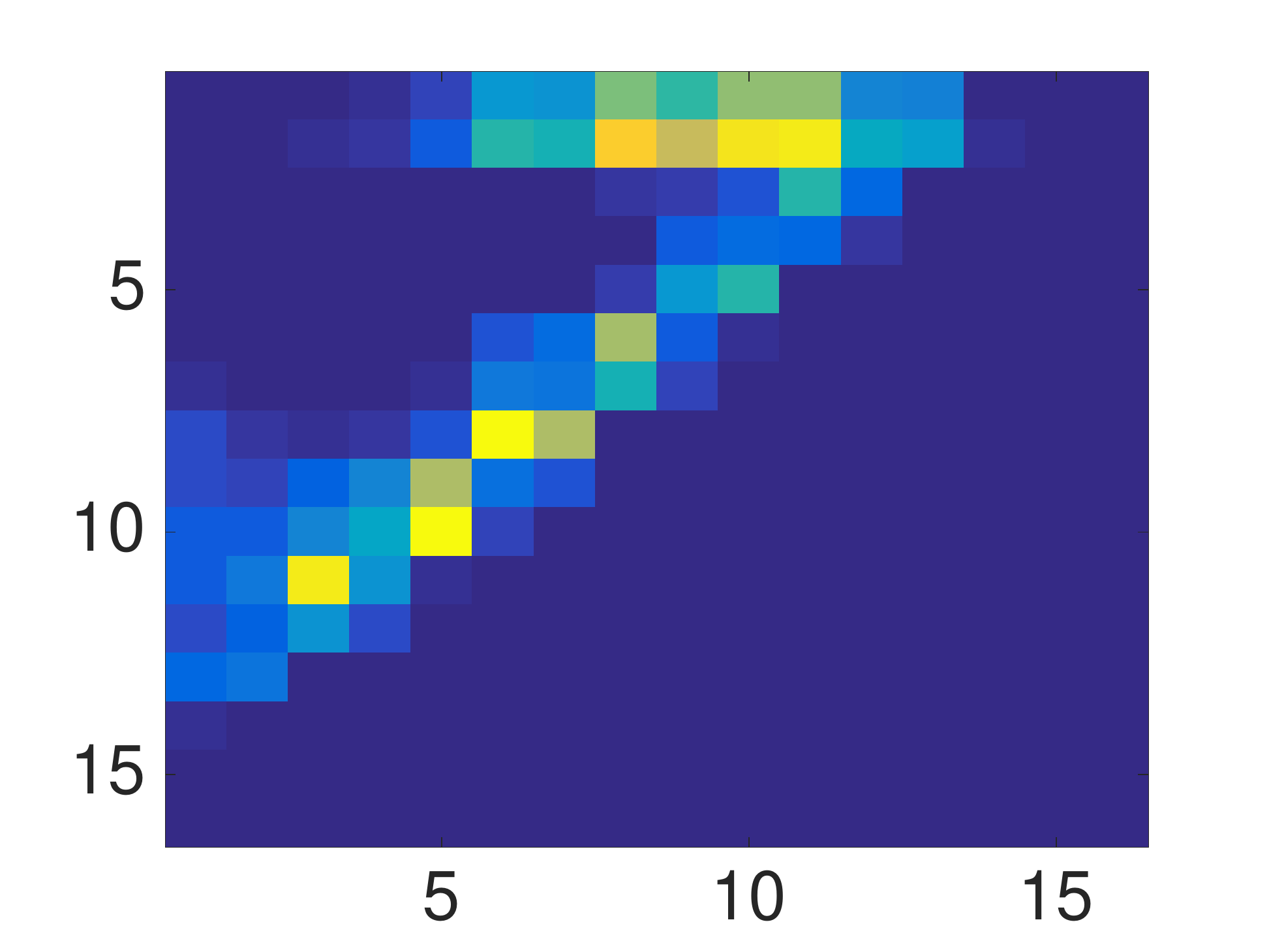}}
	\caption{The patch from Fig.~\ref{fig:patch}a preprocessed with: (a) Schmid filter of $size=49$, $\sigma=6$ and $\tau=1$; (b) MR filter of $size=49$, $\sigma_x=2$ and $\sigma_y=6$; (c) CLBP of type riu2, $n=8$ and $r=3$; together with their PIs (d-f).}
	\label{fig:differentFilters}
\end{figure}

The patch from Fig.~\ref{fig:patch}a processed with Schmid filters, MR filters, and with CLBP is presented in Fig.~\ref{fig:differentFilters}, together with the respective PIs. It can be observed that different pre-processing leads to strongly different representations. Again, it is not clear which representation is most robust and beneficial with respect to classification accuracy.

\subsection{Normalization}
\label{sec:normalization}

Especially when working with depth maps normalization plays an important role for persistent homology analysis. The input depth maps have different depth value ranges depending on the characteristics of the 3D surfaces (e.g. deep valleys and high hills versus flat surfaces). Normalization can be performed at a global or local level.

\paragraph{Global normalization:} To compute comparable topological descriptors the depth maps need to be mapped to a similar value range first, i.e. the distribution of the depth values needs to be standardized.  We apply z-standardization to the depth values of each depth map separately. As a result all depth maps have zero mean and unit variance and thus cover a similar value range (except for outliers).

\begin{figure} [t]%
\centering
	\subfigure[]{\label{patch:fig1up}
	\includegraphics[width=0.4\linewidth]{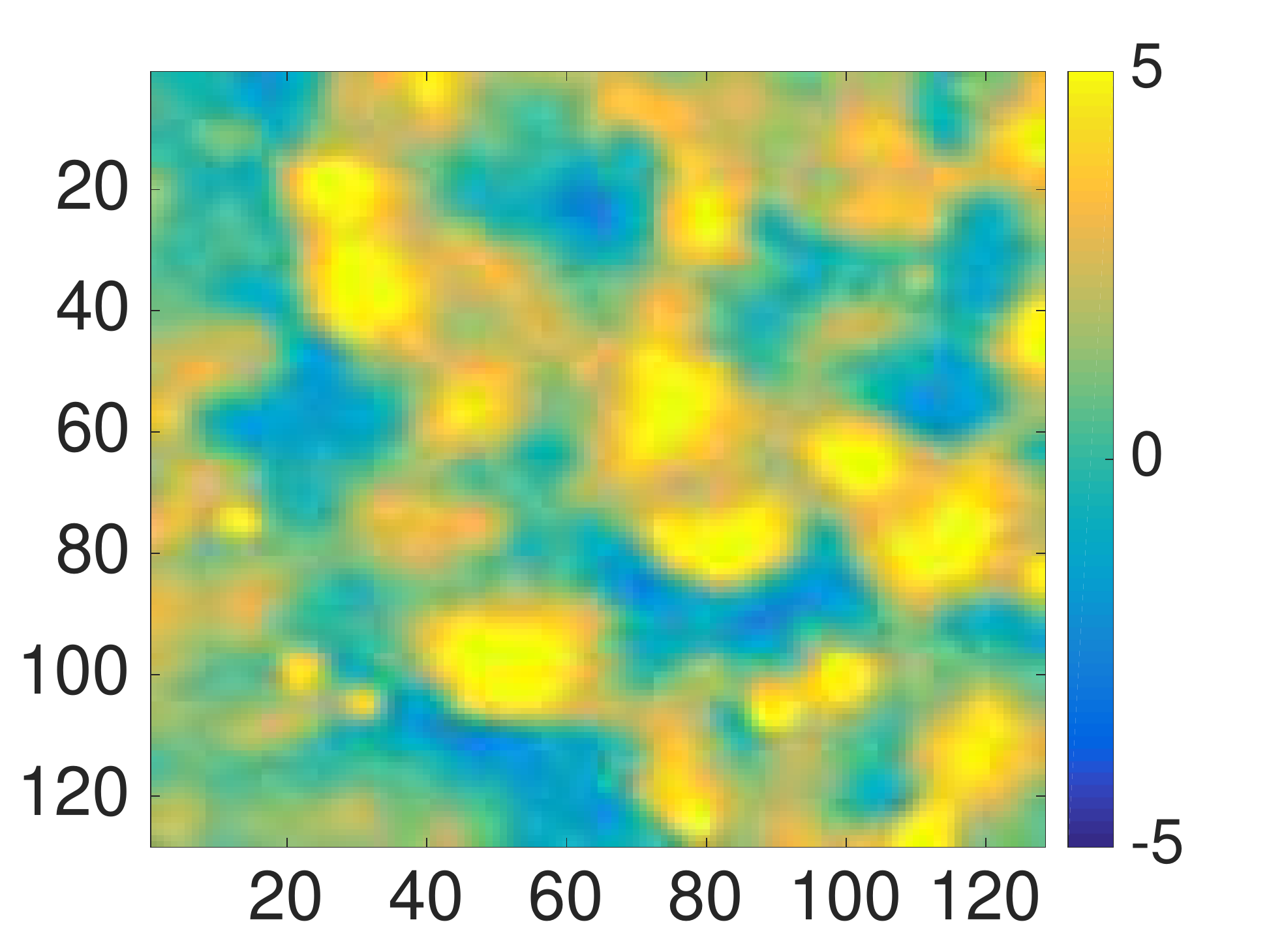}}
	\subfigure[]{\label{patch:fig2down}
	\includegraphics[width=0.4\linewidth]{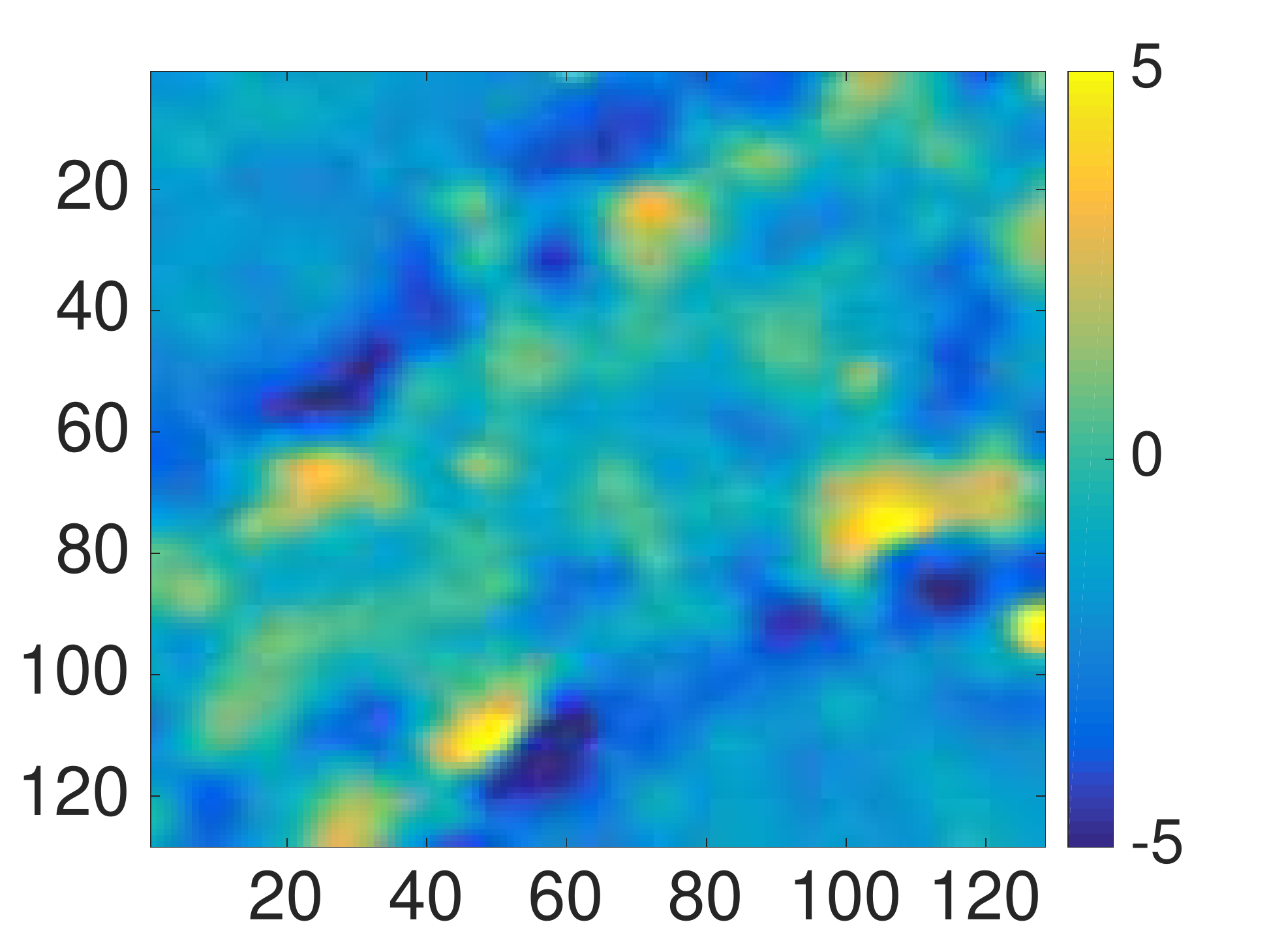}}
	\subfigure[]{\label{patch:fig1upPI}
	\includegraphics[width=0.4\linewidth]{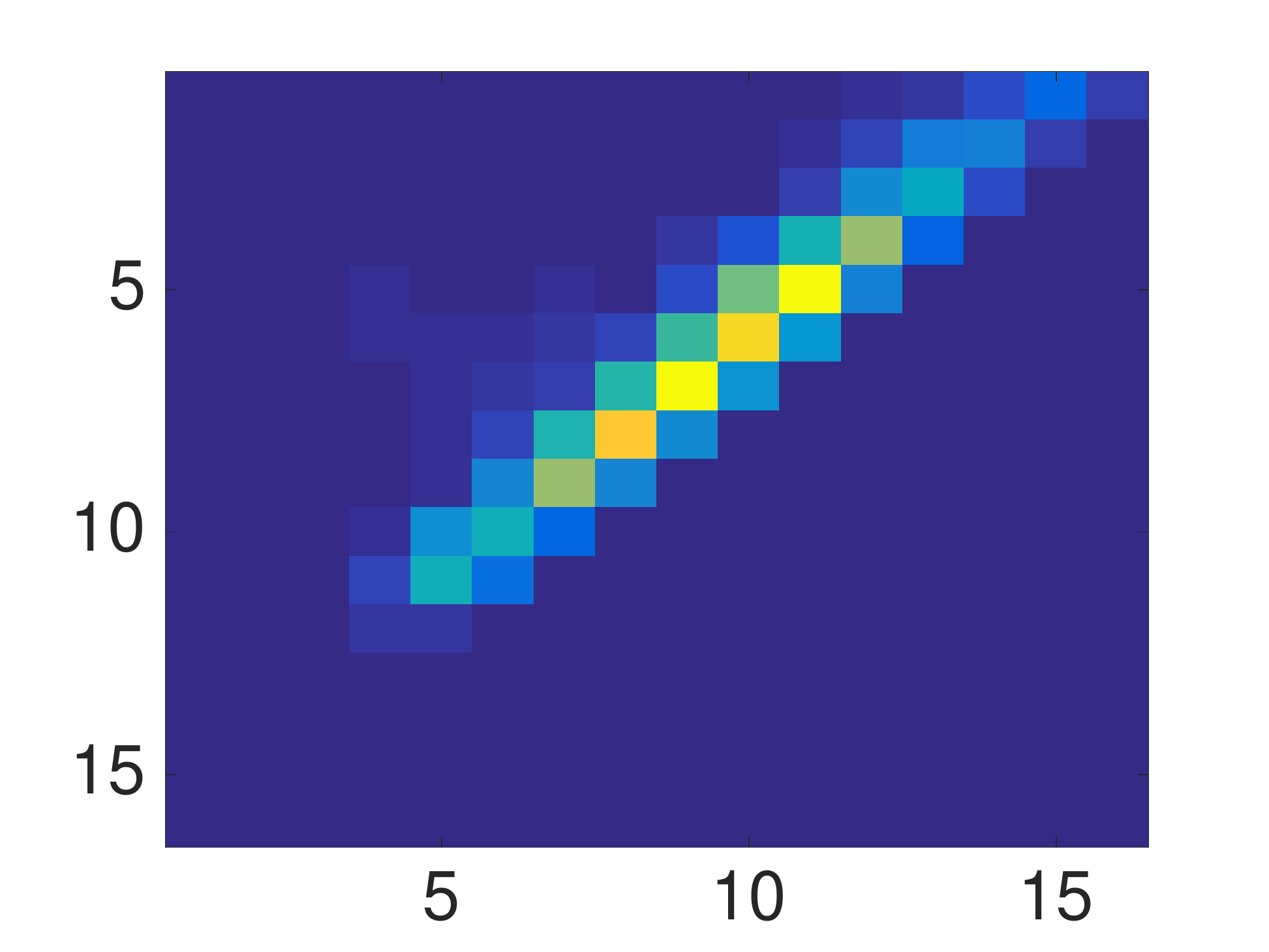}}
	\subfigure[]{\label{patch:fig2downPI}
	\includegraphics[width=0.4\linewidth]{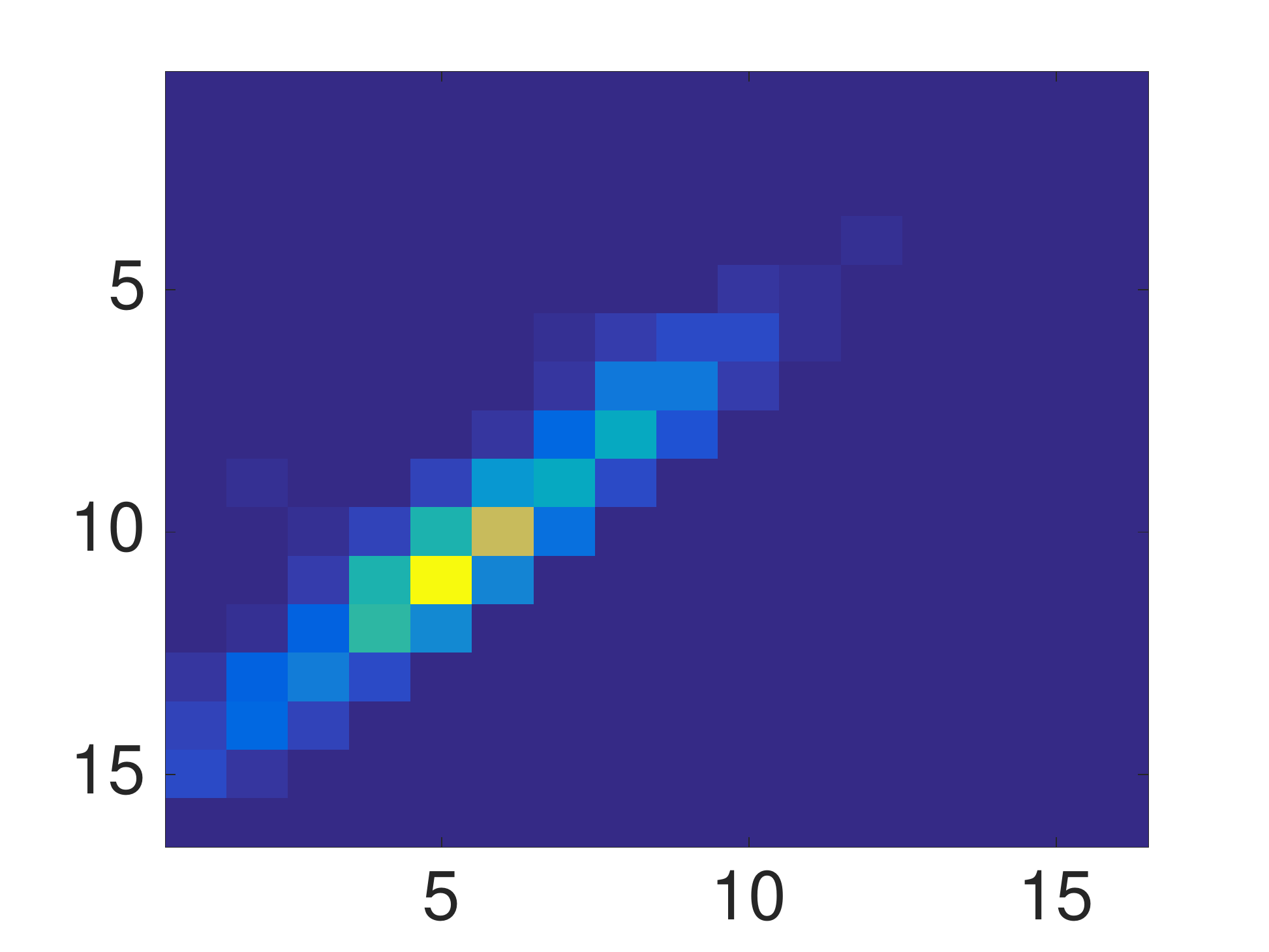}}\\
	\subfigure[]{\label{patch:fig1upNorm}
	\includegraphics[width=0.4\linewidth]{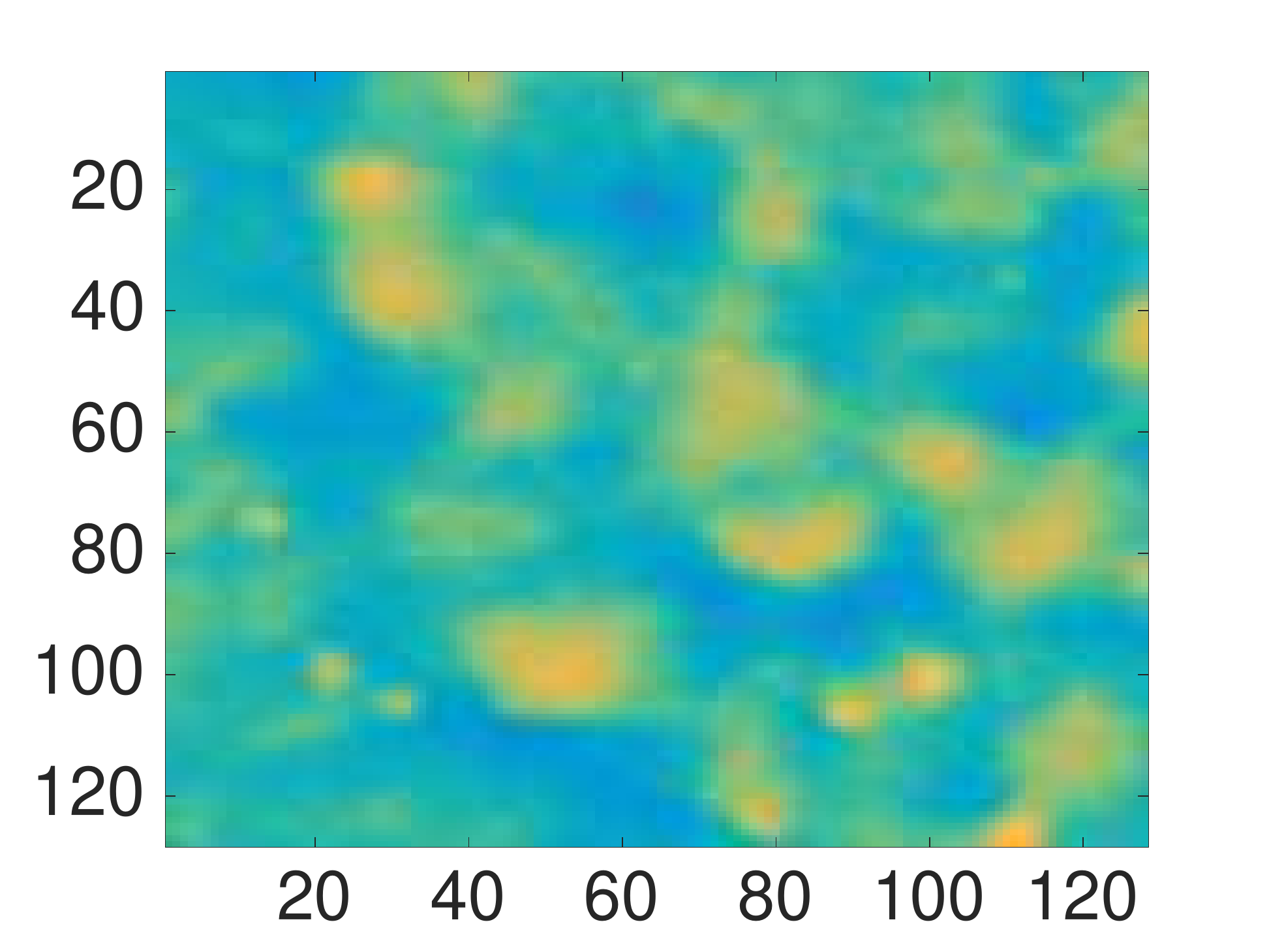}}
	\subfigure[]{\label{patch:fig2downNorm}
	\includegraphics[width=0.4\linewidth]{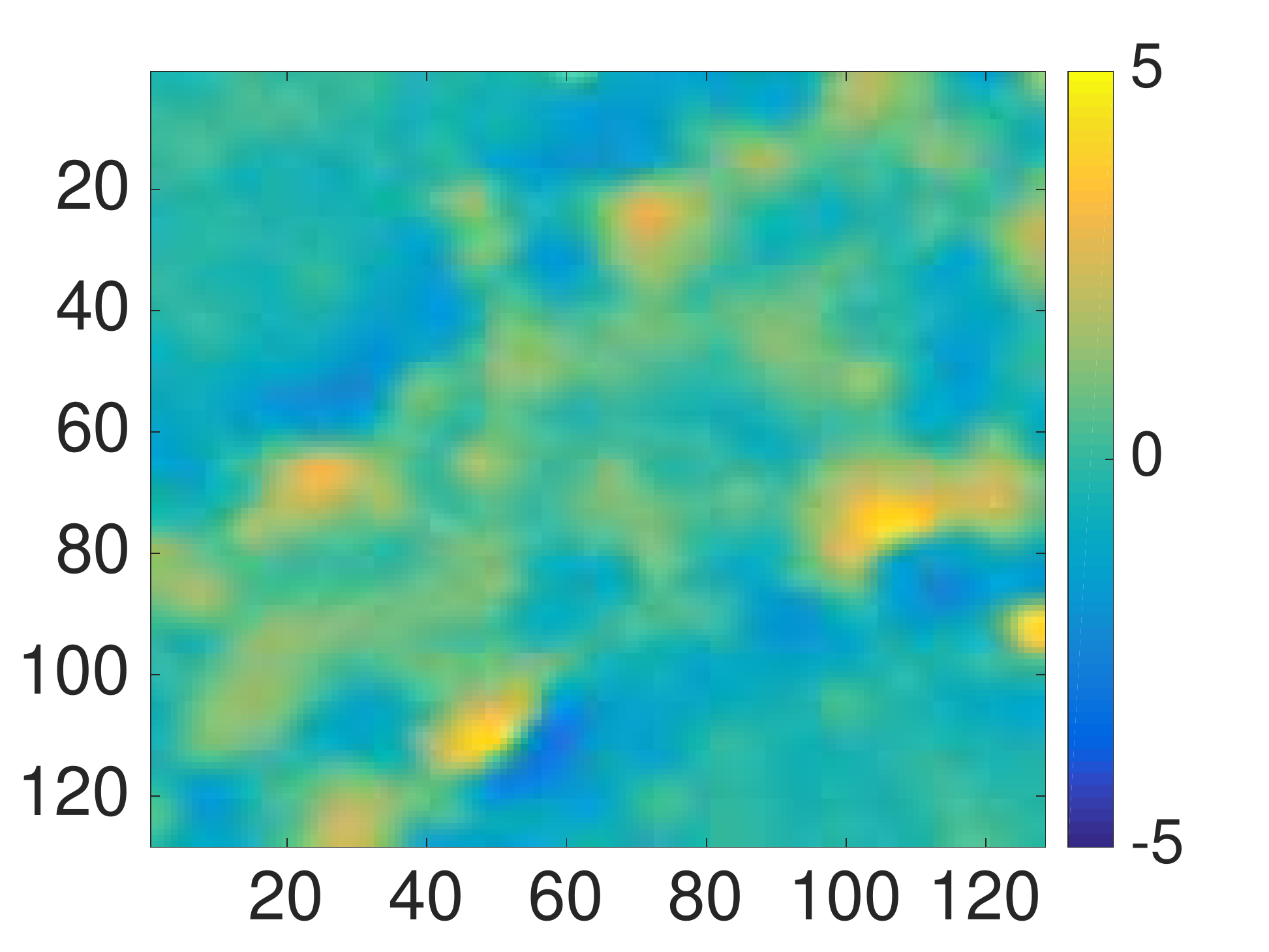}}
	\subfigure[]{\label{patch:fig1upNormPI}
	\includegraphics[width=0.4\linewidth]{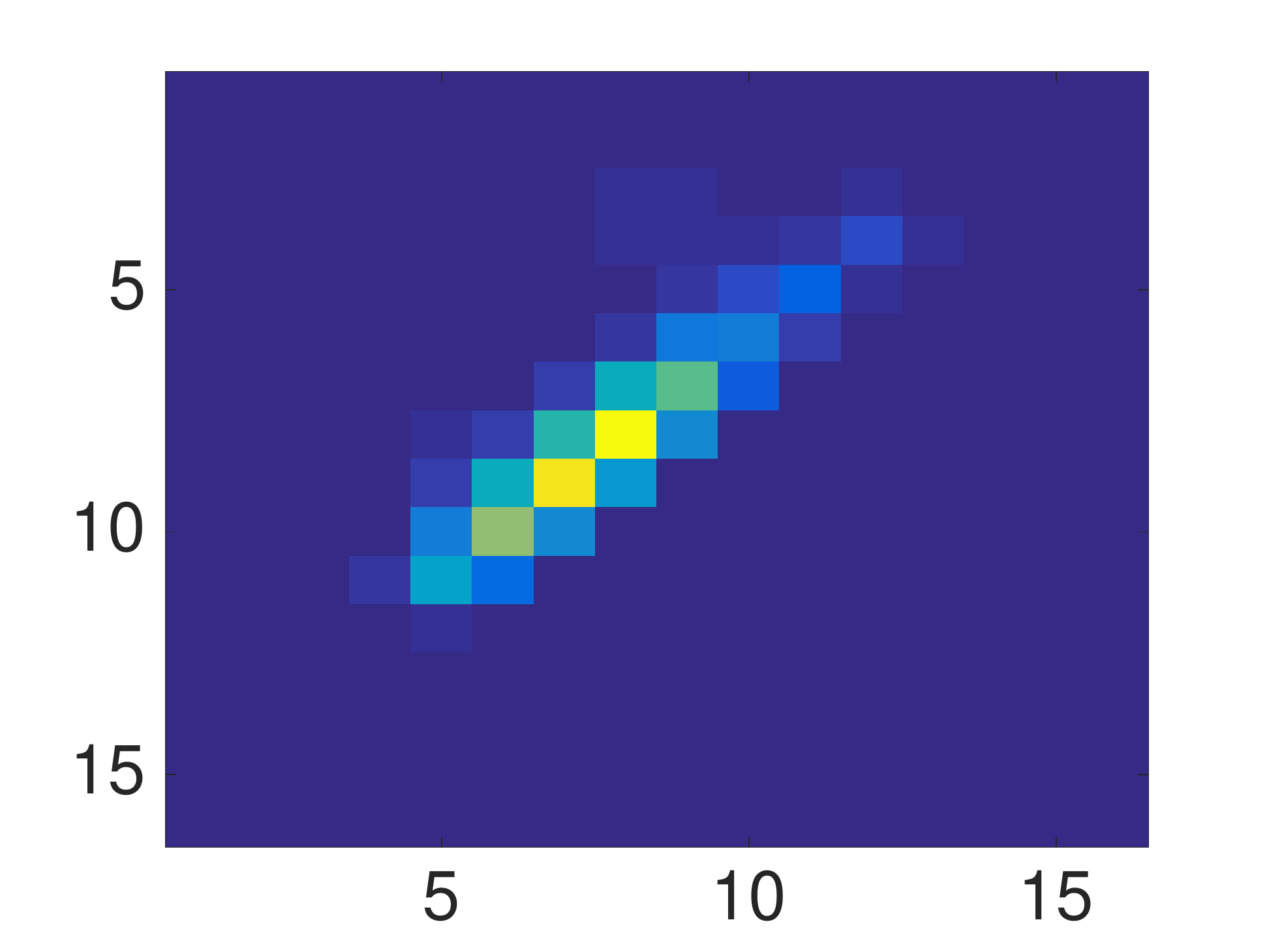}}
	\subfigure[]{\label{patch:fig2downNormPI}
	\includegraphics[width=0.4\linewidth]{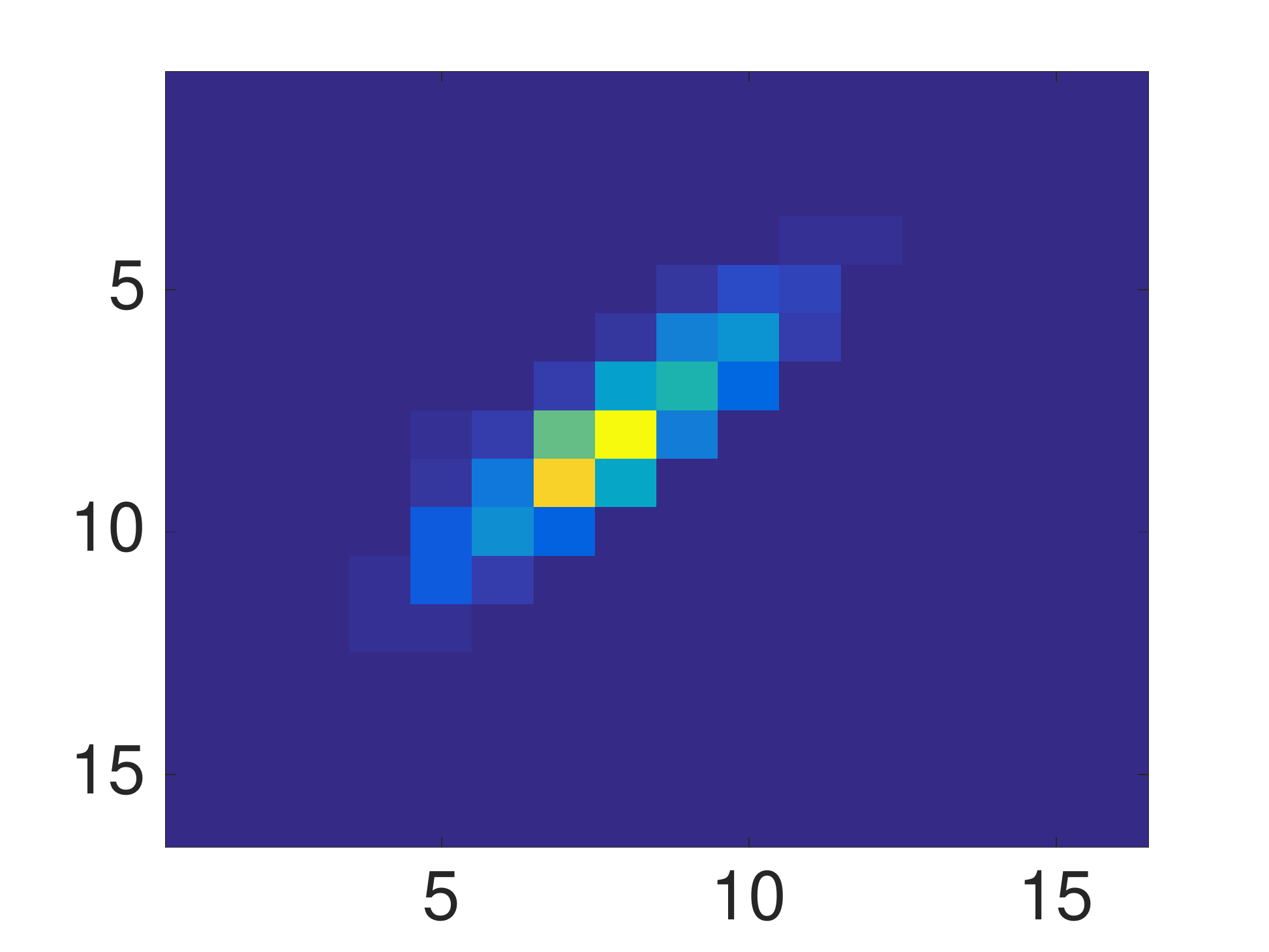}}\\
	\caption{The effect of local normalization: two patches (a) and (b) which are located at different depth levels of the surface together with their PIs (c) and (d). The two patches become more similar after applying local min-max normalization (e) and (f). A similar effect is observed for the corresponding PIs (g) and (h).}
	\label{fig:norm}
\end{figure}

\paragraph{Local normalization:} Aside from standardizing the value distribution of the depth maps additional normalization can be applied locally \emph{across} the individual patches of the surface. Different types of normalization add different invariants to the resulting feature representations which may result in more robust classification. By performing a patch-based normalization we remove the global reference between the patches, i.e. each patch's value distribution is normalized individually, independently from the neighboring patches. Thus, after normalization the absolute depth reference is removed. What remains is pure topological surface information. This means that two patches with similar topological structure at different depth levels of the surface become similar and are represented by similar PDs. Without patch-based normalization the absolute depth information is retained and the two patches yield different (shifted) PDs, see Fig.~\ref{fig:norm} for an example.

Given a patch P, different types of local normalization can be applied, such as:
\begin{itemize}
\setlength{\itemsep}{-0.1\baselineskip}
\item z-standardization: $\overline{P}=(P-mean(P))/std(P)$,
\item min-max normalization: $\overline{P}=(P-min(P))/(max(P)-min(P))$ and
\item positional standardization: $\overline{P}=(P-median(P))/mad(P)$,
\end{itemize}
where $std$ refers to the standard deviation and $mad$ to the median absolute deviation.

\subsection{Stability}
\label{sec:stability}

A crucial property of persistent homology is the stability of persistence diagrams. Briefly speaking, we expect that when a measurement function (in our case represented by a patch) changes sligthly, then its persistence diagram do not change too much as well. There are two major metrics in PDs space: bottleneck distance ($d_B$) and Wasserstein distance ($d_W$).
Cohen-Steiner et al. prove in~\cite{Cohen-Steiner2007,Cohen-Steiner2010} results about stability of PD under $d_B$ and $d_W$ metrics for classes of the measurement functions. Especially, by~\cite[Wasserstein Stability Theorem]{Cohen-Steiner2010} we know that the computation of persistence diagrams  for Lipschitz functions is stable under $d_W$. 

In the case of depth map it is easy to observe that the measurement function is Lipschitz (the size and pixel values are bounded).
In~\cite{adams2015persistent} it is proven that PIs are stable under $d_W$, as well. Hence, we consider the  process of obtaining PIs from surface patches being stable in theory. In our experiments we analyze this the stability of the resulting descriptors in practice.

\section{Datasets and evaluation protocol}
\label{sec:setup}
For our experiments, we employ a recently released dataset of high-resolution 3D reconstructions from the archaeological domain with a resolution of approximately 0.1~mm~\cite{poier2016petrosurf3d}. The dimensions of the scanned surfaces ranges from approx. $20 \times 30$ cm to $30 \times 50$ cm. The reconstructions represent natural rock surfaces that exhibit human-made engravings (so-called rock art). The engravings represent animals, symbols and figures that have been engraved by humans in ancient times. See Fig.~\ref{fig:depthMap} for an example surface. The engraved regions in the surface exhibit a different surface texture than the surrounding natural rock surface.   In our experiments we aim at automatically separating the engraved areas from the natural rock surface.

For each surface a precise ground truth has been generated by domain experts that labels all engravings on the surface. The dataset contains two classes of surface topographies: class 1 represents engraved areas and class 2 represents the natural rock surface. Class priors are imbalanced. Class 1 represents approximately 19\% of the data and is thus underrepresented. A corresponding ground truth labeling is depicted in Fig.~\ref{fig:depthMap}c. 

In total 26 high-resolution surface reconstructions have been acquired as described above. To accelerate experiments, we form two datasets of different sizes:

\paragraph{Small-scale dataset: }
The small scale dataset contains depth maps from 4 surface reconstructions with a total number of 12.3 millions of points. This dataset has also been employed for the experiments in~\cite{zeppelzauer2015efficient} and thus represents a baseline dataset. We employ this dataset for our initial experiments to reduce computation time and to enable a broader set of experiments in our evaluation. Furthermore, by using this dataset our results become comparable to previous work. In our evaluation the depth maps of the first two surface reconstructions are employed for training and the remaining two surfaces represent the independent evaluation set.

\paragraph{Large-scale dataset: }
This dataset contains depth maps of all the 26 surface reconstructions that have been acquired. This dataset is employed in our evaluation to demonstrate the generalization ability of the proposed approach and to verify if conclusions drawn from the small-scale dataset can also be transferred for the larger and more complex one. For classification experiments we randomly select 13 surfaces \review{as a training set} and employ the remaining ones for evaluation.

\review{We further perform experiments on synthetic data to provide an experimental proof-of-concept for our topological approach.}

\begin{figure}%
\centering
	\subfigure[ideal natural]{
	\includegraphics[width=0.31\linewidth]{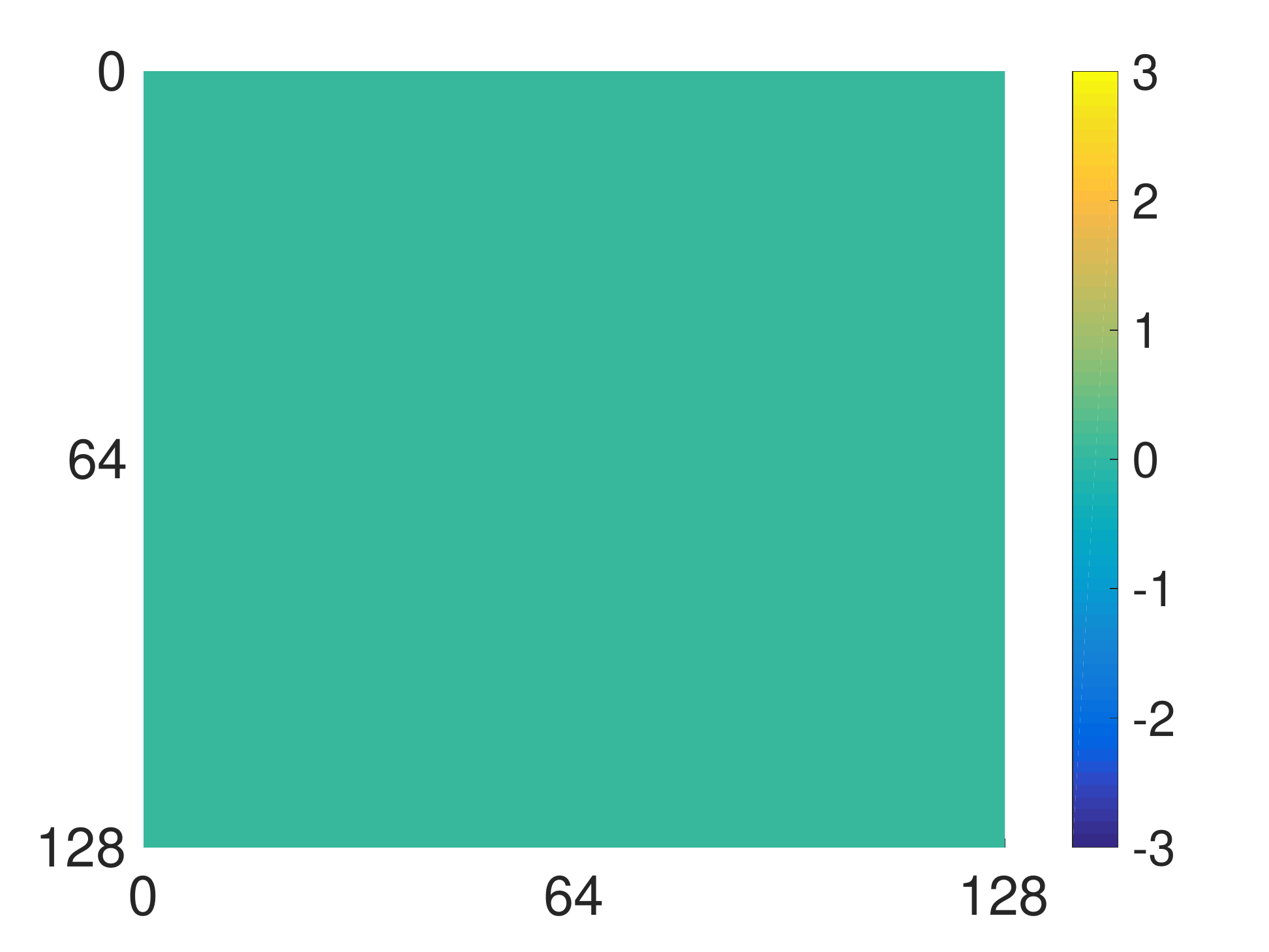}}
	\subfigure[ideal engraved I]{
	\includegraphics[width=0.31\linewidth]{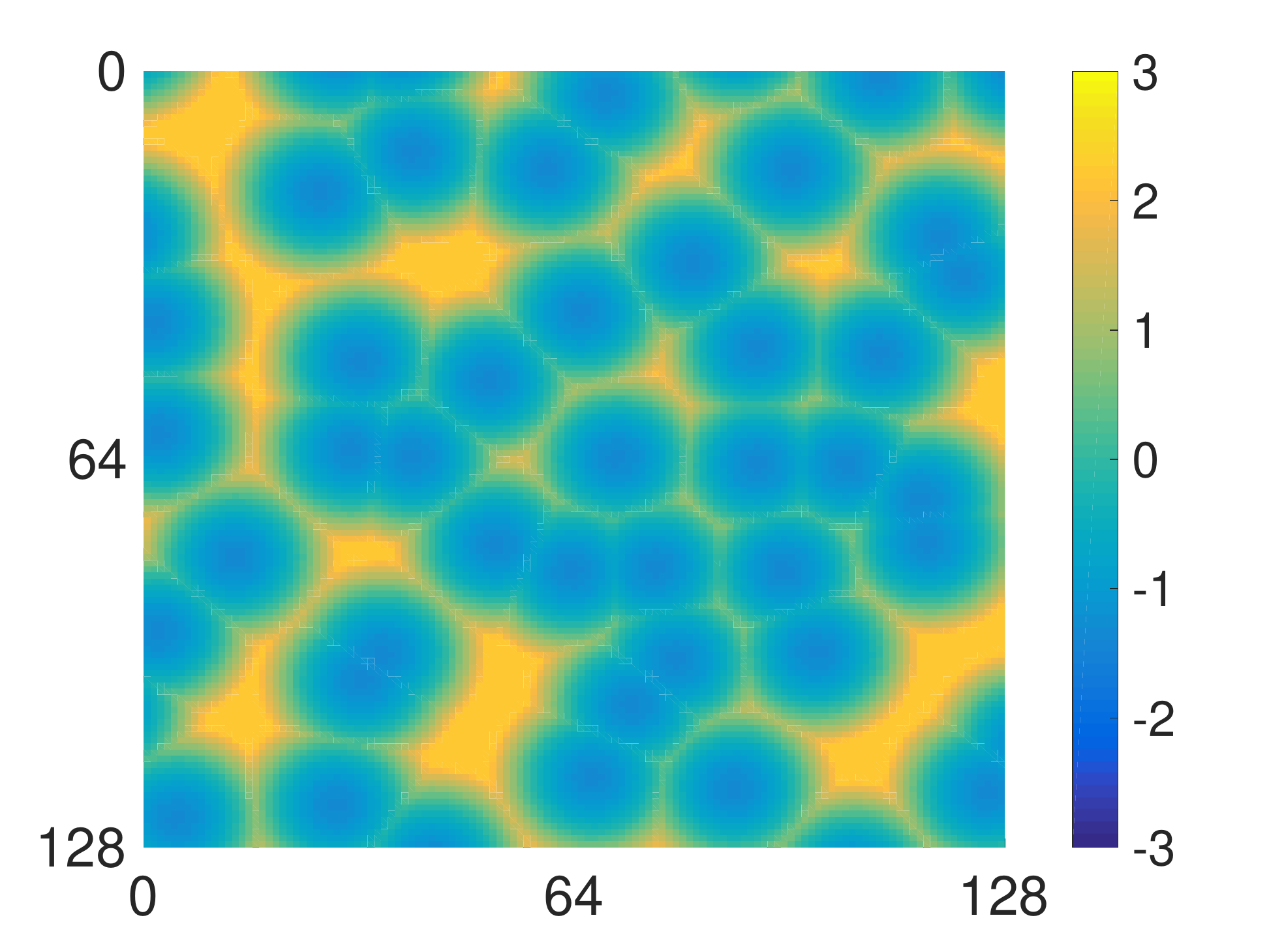}}
	\subfigure[ideal engraved II]{
	\includegraphics[width=0.31\linewidth]{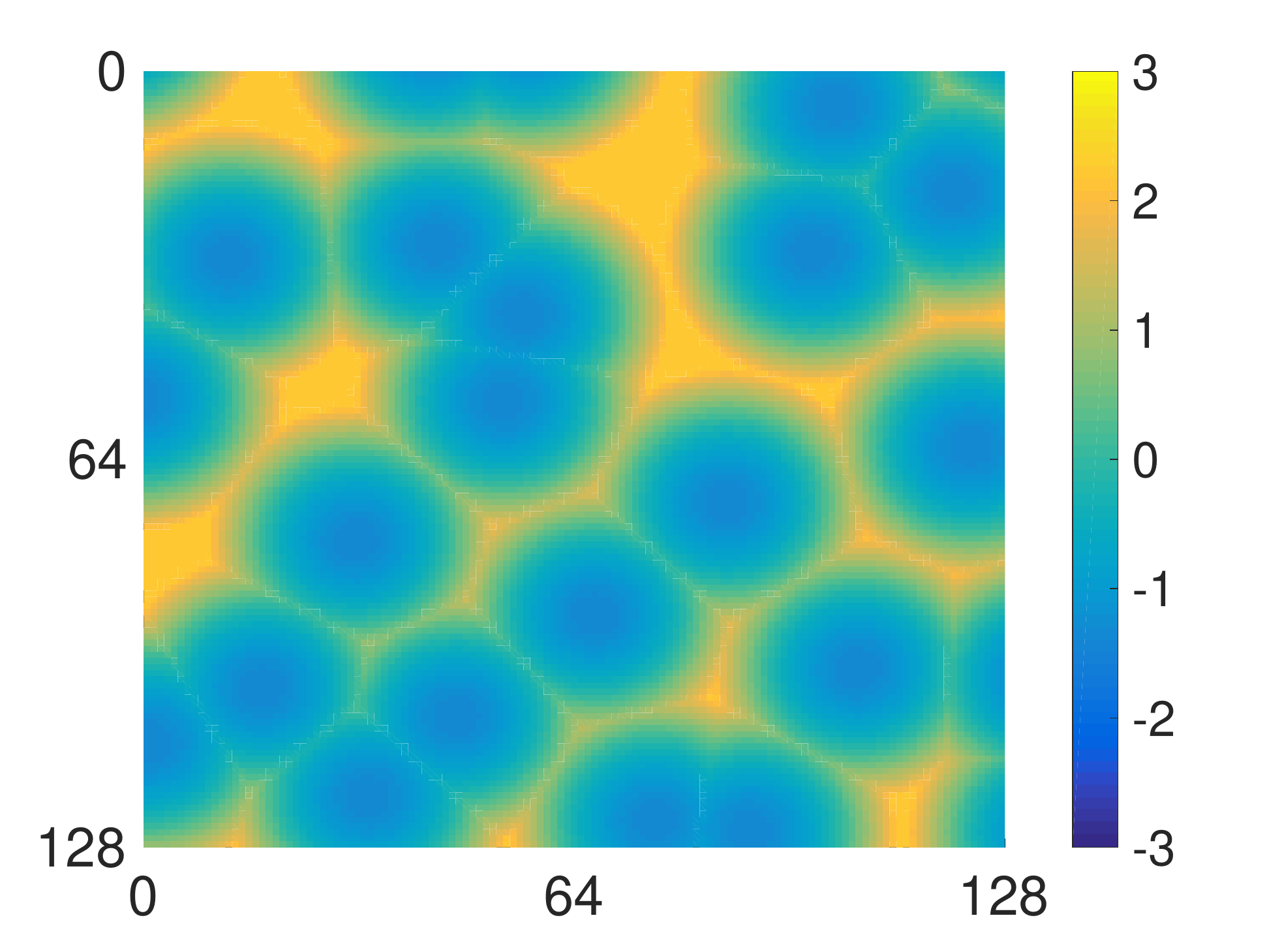}}\\
	\subfigure[natural+noise]{
	\includegraphics[width=0.31\linewidth]{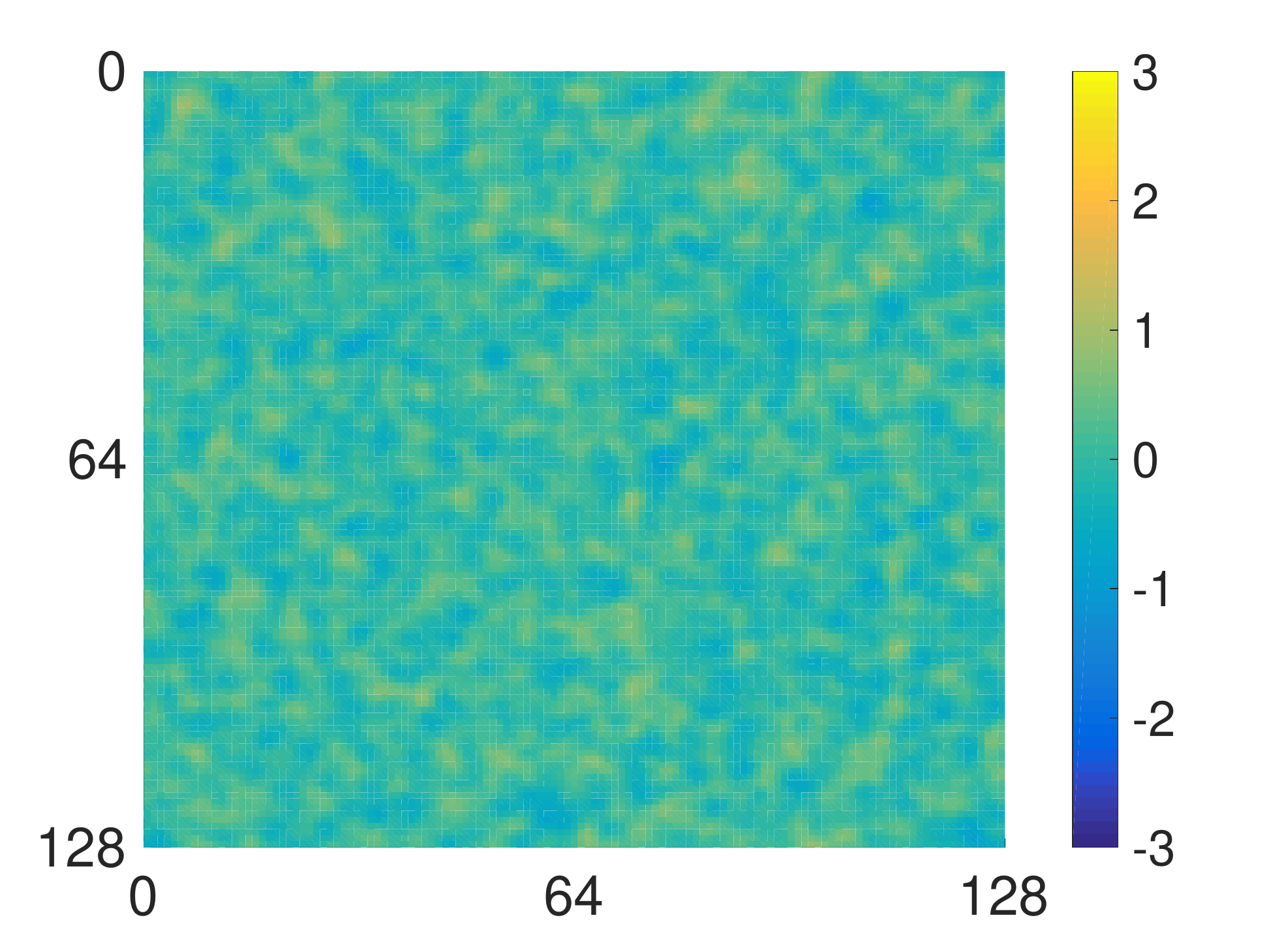}}
	\subfigure[engraved I+noise]{
	\includegraphics[width=0.31\linewidth]{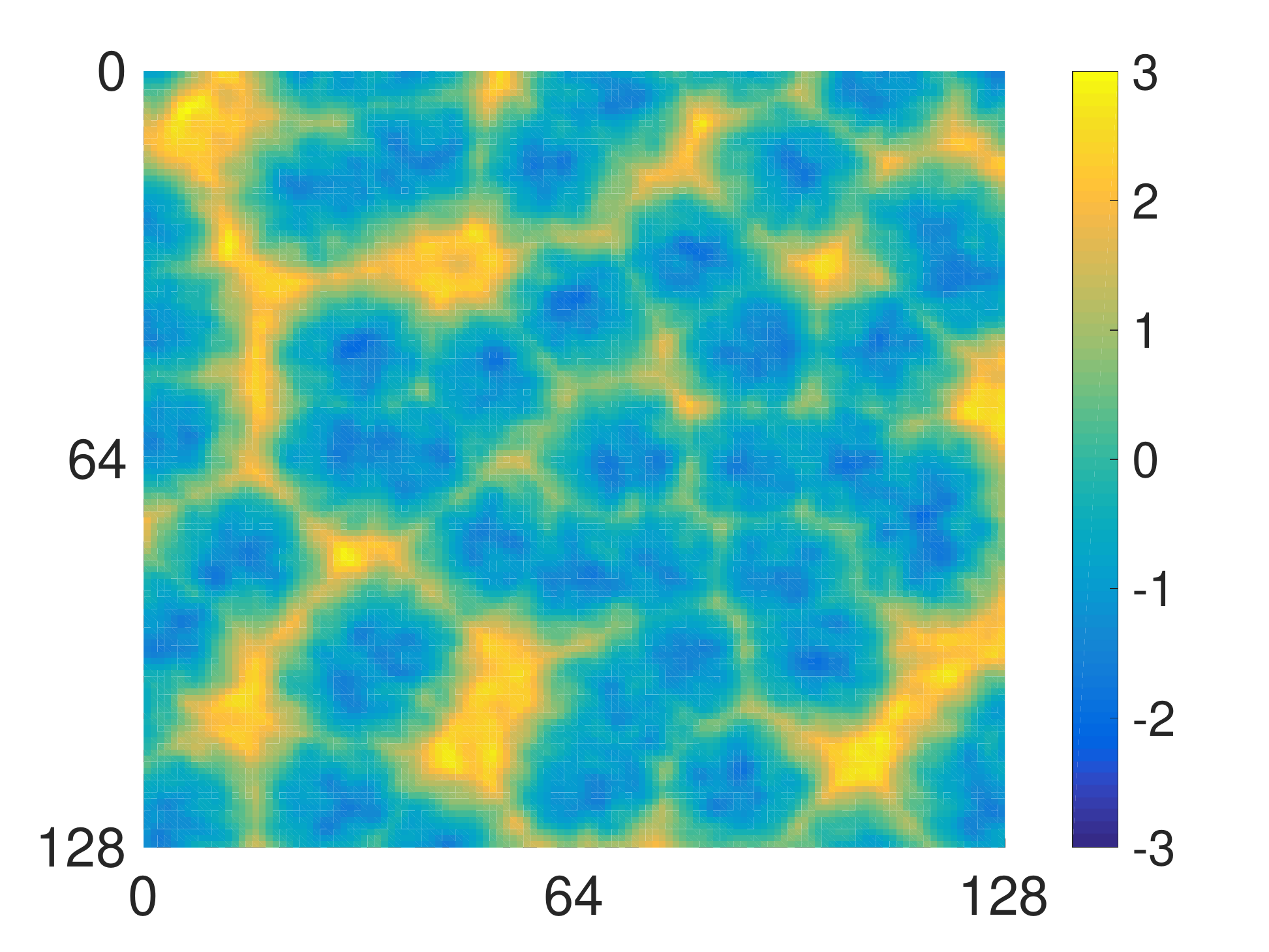}}
	\subfigure[engraved II+noise]{
	\includegraphics[width=0.31\linewidth]{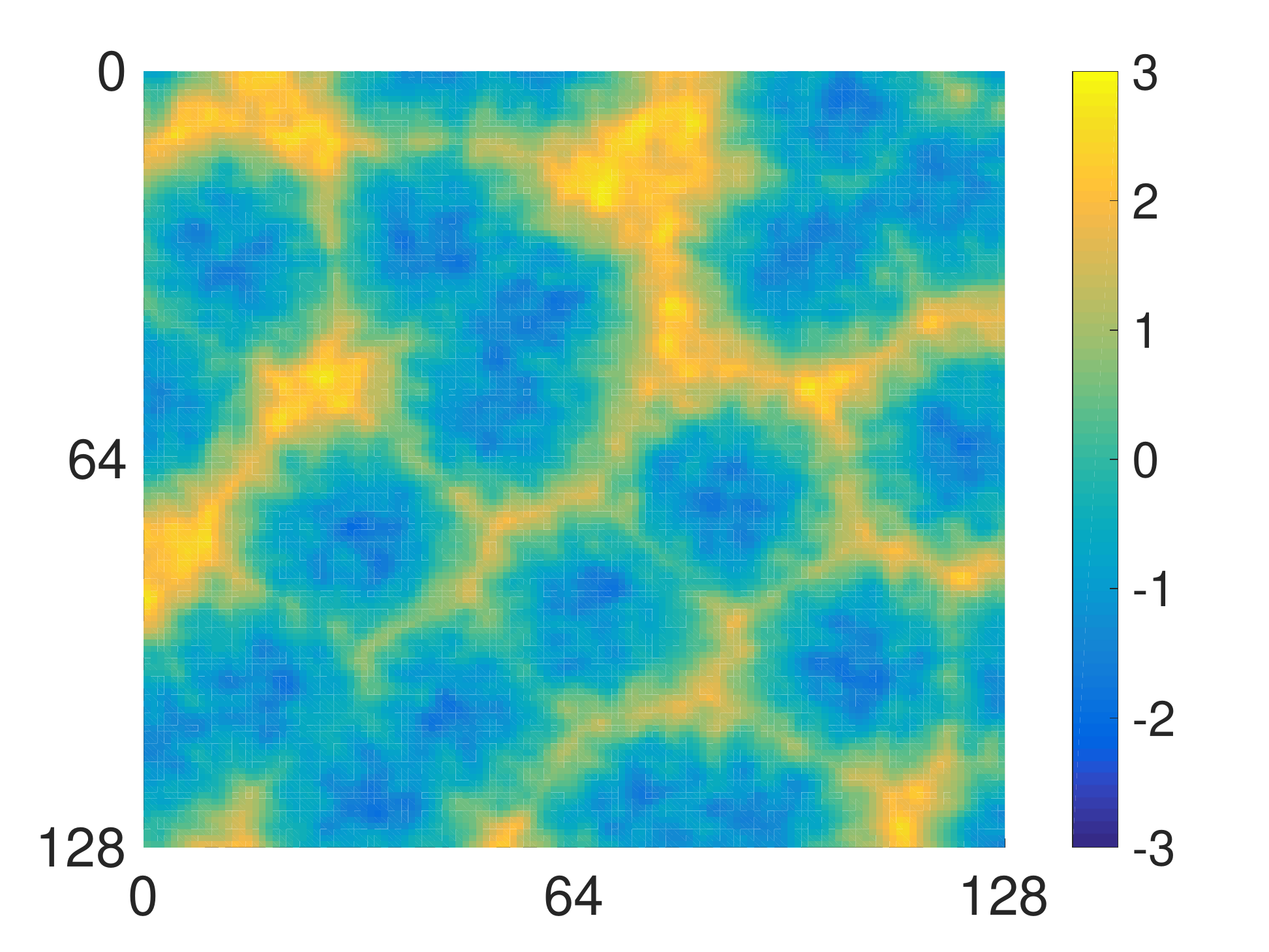}}\\
	\subfigure[PI natural]{
	\includegraphics[width=0.31\linewidth]{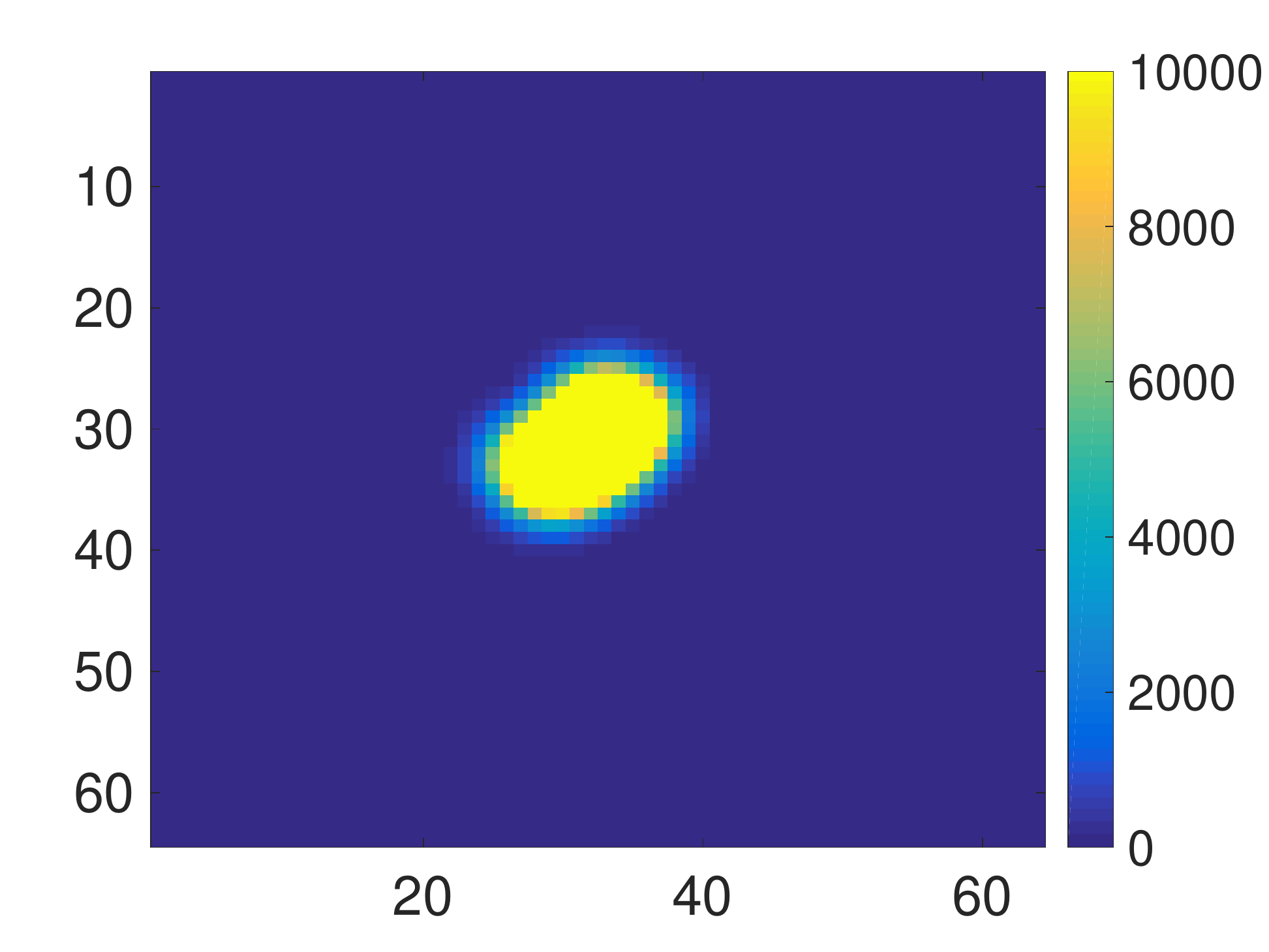}}
	\subfigure[PI engraved I]{
	\includegraphics[width=0.31\linewidth]{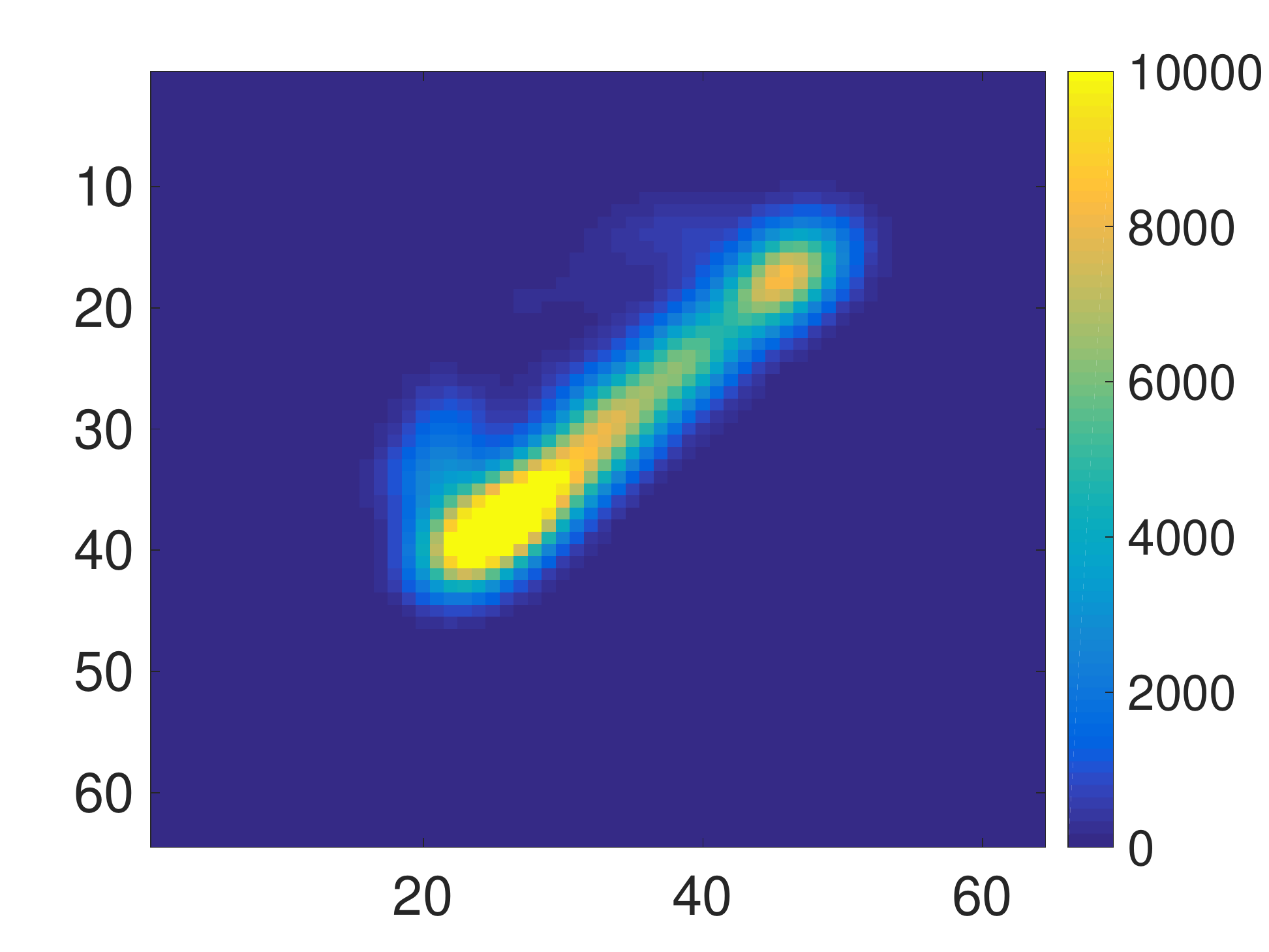}}
	\subfigure[PI engraved II]{
	\includegraphics[width=0.31\linewidth]{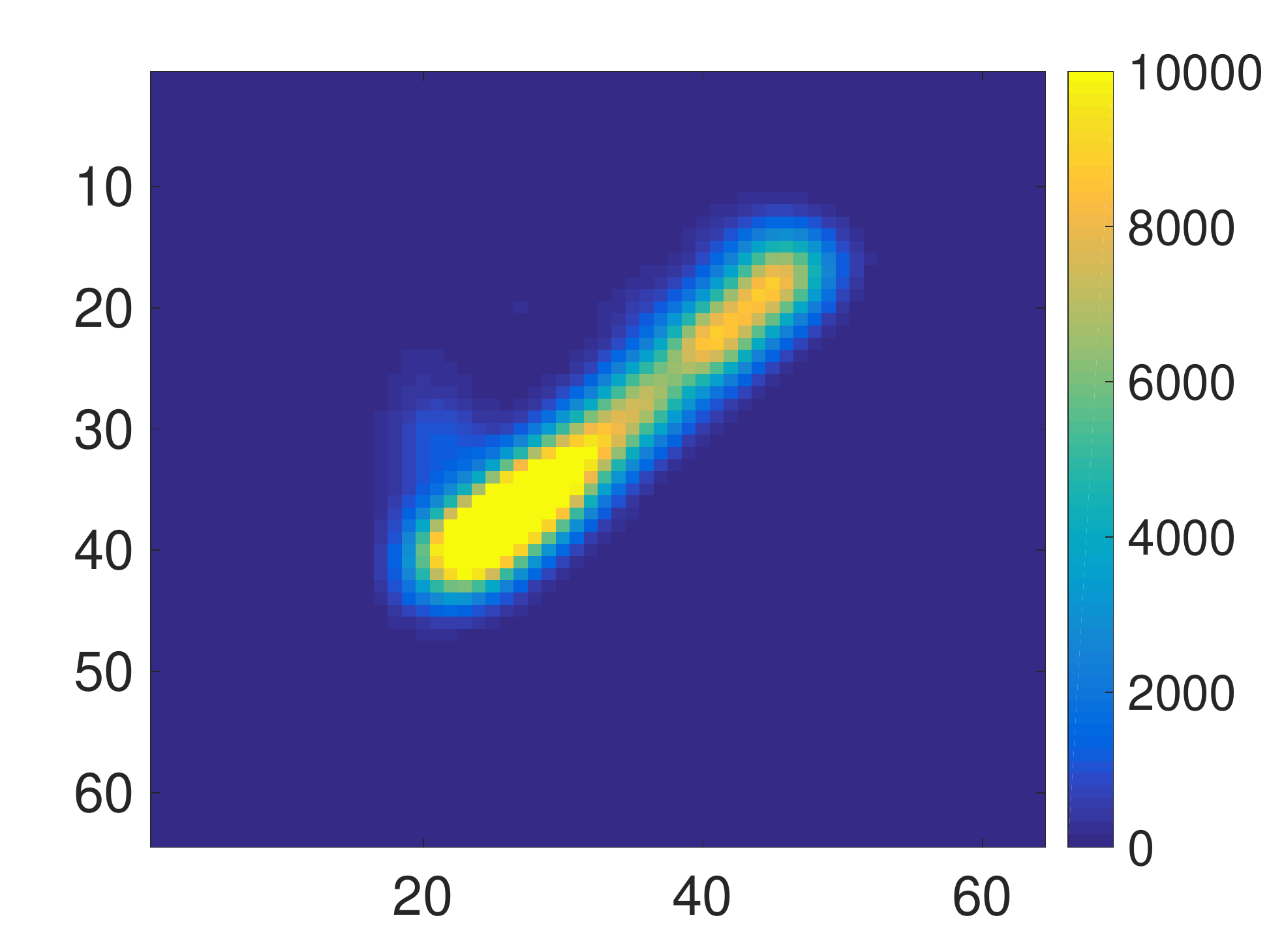}}\\
	\caption{\review{Synthetic dataset: ``natural" and ``engraved" surfaces without noise (a-c); the same surfaces with noise (d-f); and their PIs (g-i). PI generated for the ``natural" surface is unimodal, while PIs generated for ``engraved" surfaces are bimodal and less coherent.}}
	\label{fig:synthetic}
\end{figure}

\paragraph{Synthetic dataset: }
\review{The synthetic dataset serves as a proof-of-concept to simulate the process of engraving a natural rock surface and the  effect of the resulting surface topographies on the topological descriptors. We start with an ideally flat surface that represents a ``natural" rock surface without any irregularities, see Fig.~\ref{fig:synthetic}a. All values of this surface are constant and equal to zero. In Figs.~\ref{fig:synthetic}b and c we simulate differently dense and coarse engravings by adding negative Gaussian distributions to the flat surface. The spacing between Gaussians equals $20 \pm 5$ and $30 \pm 5$ pixels. In a next step, we add noise to all surfaces to simulate natural irregularities and effects of long-time exposition of the surfaces to weather. Noise is generated using the surface generator of \cite{garcia1984monte}\footnote{implementation available from: \url{http://www.mysimlabs.com/surface_generation.html}, last visited June 2017}. Fig.~\ref{fig:synthetic}d shows the resulting noise, i.e. the simulated flat natural surface. We further add the same noise to the simulated engraved surfaces from Fig.~\ref{fig:synthetic}b and c, see Figs.~\ref{fig:synthetic}e and f. The depth values of all surfaces are z-standardized. 
}

\paragraph{Evaluation Protocol}
Our evaluation protocol is defined as follows: from the input depth maps, we extract overlapping square image patches which are input to feature extraction (either for topological features or non-topological features). The resulting feature vectors for each patch are then input to classification. The classification performance is measured by the Dice Similarity Coefficient (DSC). DSC measures the matching between the ground truth and the predicted class labels, i.e. the mutual overlap between an automatic labeling $X$ of a depth map and a manual (ground truth) labeling $Y$:
\[
\text{DSC}(X,Y)=\frac{2 |X \cap Y|}{|X|+|Y|}.
\]
DSC is between $0$ and $1$, where $1$ means a perfect classification.

For each classification experiment we first select a subset $S_i$ of patches from the training set $T$. Next, we estimate suitable classifier parameters by 5-fold cross-validation on the subset $S_i$. With the optimized parameters we train a classifier on the entire subset $S_i$. This classifier is finally tested on the evaluation set. There is no parameter tuning on the evaluation set.

Each classification experiment is repeated 10 times for different randomly selected subsets $S_i$ from the training set $T$. In each repetition other patches from the training set $T$ are selected and thus the actual training data varies over the 10 repetitions. The result are 10 classification results (DSC values) for each experiment. We report the mean and the standard deviation over all 10 repetitions. By conducting each classification experiment with different subsets of the training set we reduce the dependency of our results on the training data.

As mentioned above, the class priors in our dataset are imbalanced. Skewed datasets pose problems to most classification techniques and often yield suboptimal models as one class dominates the other classes. A classifier especially designed for imbalanced datasets is Random Undersampling Boosting (RUSBoost)~\cite{seiffert2010rusboost}. RusBoost builds upon AdaBoost~\cite{freund1997decision} which is an ensemble method that combines the weighted decisions of weak classifiers to obtain a final decision for a given input patch. RUSBoost extends this concept by a data sampling strategy that enforces similar class priors. During each training iteration the majority class in the training set is undersampled in a random fashion to balance the resulting class priors. In this manner, the weak classifiers can be learned from balanced datasets without being biased from the skewed class distribution. In previous experiments RUSBoost has already achieved robust results and high classification performance and has  outperformed among others cost-sensitive SVMs~\cite{fumera2002cost} with linear and RBF kernels which are also suitable for imbalanced data.

The code used in our study was implemented in Matlab. Most of the descriptors were extracted with VLFeat library~\cite{vedaldi2010vlfeat}, except for PD\_AGG and PI which were computed with the CAPD:: RedHom library~\cite{RedHom,juda2014capd} with the PHAT~\cite{url:PHAT,PHAT} algorithm for persistence homology.

\section{Experiments and results}
\label{sec:experimentalResults}

\review{We first investigate the expressive power of topological descriptors for our synthetic data set and then perform a comprehensive study on the robustness and expressiveness of the topological descriptors for the analysis of real-world 3D surfaces. We investigate different variants of their computation and compare them with traditional non-topological descriptors. Finally, we investigate the effect of combining topological with non-topological descriptors. For each set of experiments we first describe the experimental setup and the investigated aspects in the context of the experiment and then report on the obtained results. Unless otherwise stated the experiments on the 3D reconstructions were performed on the small-scale dataset.}

\subsection{Experiments on synthetic data}
\label{sec:synthetic}

\review{We generated PI descriptors for the synthetic surfaces using the baseline configuration (see Tab.~\ref{tab:parameters}). The PI generated for the natural surface in Fig.~\ref{fig:synthetic}g shows a bright cluster in the center of PI. This is plausible, because most of the holes are born and die at the same time. The PIs of the two engraved surfaces in Figs.~\ref{fig:synthetic}h and i exhibit a completely different spatial distribution. The reason is that there are two types of holes on the surfaces. One group of holes corresponds to the natural irregularities (noise) and the other one to the engravings. Holes from both groups are born and die at different times. Thus the PIs become bimodal. This is strongly different from the natural surface for which PI is unimodal and more coherent.}

\review{These experiments demonstrate that the topological descriptors capture different surface topographies well. In the following sections we analyze real-world 3D surfaces and perform a detailed analysis of topological and non-topological descriptors.}

\subsection{Topological baseline descriptors}

In an initial experiment we extract topological descriptors which serve as a baselines for later comparison. From the depth maps in the training and evaluation sets we extract topological descriptors in a patch-based manner. The patch size is $128 \times 128$~pixels (i.e. $10.8 \times 10.8$~mm) and the step size between two patches is 16~pixels (1.35~mm). The baseline descriptors include: (i) aggregated descriptors from the persistence diagram (PD\_AGG), see Section~\ref{subsec:PDAGG} and (ii) descriptors obtained from the persistence image (PI), see Section~\ref{subsec:PI}.

For both descriptors we select those parameters for which they have yielded best results in our previous investigation~\cite{zeppelzauer2016topological}. See Tab.~\ref{tab:parameters} for the parameter selections for the baseline configurations and an overview of all possible parameter values.

PD\_AGG results in an 12-dimensional feature vector (see Section~\ref{subsec:PDAGG}), while the feature vector derived from PI has a dimension $D$ of 136 which originates from the fact that all values under the diagonal are zero and are skipped. Thus, $D=(R^2+R)/2$ where $R$ refers to the  resolution of the PI ($R=16$ for the baseline configuration). 

As already mentioned in Section~\ref{sec:normalization} the input depth maps have different depth value ranges and are z-standardized to make different surfaces comparable. After z-standardization the depth maps have zero mean and unit variance and thus cover a similar value range (except for outliers). Nevertheless we unify the PDs by defining a parameter ``diagram limits" as an interval $[min, max]$. 
Both values $[min, max]$ define the axis limits of the underlying PD. We set these values to $[-5, 5]$ which can safely be done as after z-standardization the depth values most likely lie in the interval $[-5, 5]$ except for individual outliers. Note that if patch-based normalization is used (see Section~\ref{sec:normalization}), the diagram limits need to be adapted according to the type of normalization which may change the depth value range, e.g. in the case of min-max normalization depth values are moved to the interval $[0,1]$.

\begin{table}
  \begin{center}
  \resizebox{\linewidth}{!}{
 	\begin{tabular}{lllc}
\textbf{Descriptor} & \multicolumn{1}{l}{\textbf{Parameter}} & \multicolumn{1}{l}{\textbf{Possible values}} & \multicolumn{1}{l}{\textbf{Baseline configuration}} \\
\hline
PD\_AGG      & Local normalization   & \{none, z-std, p-std, minmax\}   & none                   \\
           & Diagram limits  & [min, max]                                 & [-5, 5]                 \\
           & Outlier removal & \{yes, no\}                   & no                     \\
           & Mapping         & \{none, CLBP\}                & none                   \\
           & Pre-filtering   & \{none, Schmid, MR\}         & none                   \\
           \hline
PI         & Local normalization   & \{none, z-std, p-std, minmax\}   & none                   \\
           & Diagram limits  & [min, max]                                 & [-5, 5]                 \\
           & Sigma           & $\mathbb{R}_{+}$            & 0.001                  \\
           & Resolution      & $\mathbb{N}_{+}$              & 16                     \\
           & Weighting       & \{none, linear, exponential\} & none                   \\
           & Outlier removal & \{yes, no\}                   & no                     \\
           & Mapping         & \{none, CLBP\}                & none                   \\
           & Pre-filtering   & \{none, Schmid, MR\}         & none                  \\
           \hline

\end{tabular}
 } 
	\caption{Topological descriptors, their computation parameters and the specification of the baseline configuration. The abbreviations ``z-std'' stands for z-standardization, ``p-std'' for positional standardization, and ``minmax" for min-max normalization.}
    \label{tab:parameters}
  \end{center}
\end{table}

The baseline configuration of PD\_AGG yields a DSC of $0.655\pm0.012$. The baseline for PI outperforms PD\_AGG with a DSC of $0.733\pm0.003$. A reason for the better performance of PI is that PI preserves spatial information from the PD as already mentioned in Section~\ref{subsec:PI} while PD\_AGG captures only overall statistics. We repeat each experiment 10 times with different training data (see Section~\ref{sec:setup} for details). The standard deviation across all the 10 experiments is low (0.012 and 0.003) which shows that the dependency on the training data is low as well for both descriptors. 

\begin{table}
  \begin{center}
  \resizebox{\linewidth}{!}{
 	\begin{tabular}{llrrrr}
\textbf{Descriptor} & & \multicolumn{4}{c}{\textbf{DSC}} \\
\hline
PD\_AGG &  &  &  &  & $\mathbf{0.655\pm0.012}$ \\
\hline
PI &  & \multicolumn{1}{c}{\textbf{$R=8$}} & \multicolumn{1}{c}{\textbf{$R=16$}} & \multicolumn{1}{c}{\textbf{$R=32$}} & \multicolumn{1}{c}{\textbf{$R=64$}} \\
& $\sigma=0.00025$ & $0.701\pm0.003$ & $0.729\pm0.003$ & $0.732\pm0.002$ & $0.725\pm0.007$ \\

& $\sigma=0.0005$ & $0.712\pm0.005$ & $0.730\pm0.002$ & $0.731\pm0.004$ & $0.730\pm0.003$ \\

& $\sigma=0.001$ & $0.722\pm0.006$ & $\mathbf{0.733\pm0.003}$ & $0.734\pm0.004$ & $0.735\pm0.004$ \\

& $\sigma=0.002$ & $0.723\pm0.002$ & $0.728\pm0.003$ & $0.730\pm0.003$ & $0.730\pm0.003$ \\
\hline
\end{tabular}
 } 
	\caption{DSC obtained for PD\_AGG and PI with various resolutions and sigmas. Numbers in bold refer to the baseline descriptors.}
    \label{tab:pdpi}
  \end{center}
\end{table}

\subsection{PI with different resolutions and sigmas}

To investigate the sensitivity of resolution and sigma for the computation of PI, we extract PIs for different resolutions (8, 16, 32, 64~pixels) and standard deviations (0.00025, 0.0005, 0.001, 0.002). The remaining parameters of the baseline configuration remain constant.

The results are presented in Tab.~\ref{tab:pdpi}. The performance of the baseline descriptors is highlighted bold. Results show that the sensitivity of PIs to both parameters is low. With increasing resolution the performance slightly improves (especially by going from 8x8 to 16x16~pixels). Higher resolutions do not provide significant improvements. Since the computation time grows quadratically with resolution it must be stressed that it is desirable to keep the resolution as small as possible. 

Experiments for different sigmas show that the best results are obtained with $\sigma = 0.001$ independently from the resolution. Thus both parameters can be optimized independently from each other which saves computation time in parameter tuning and model selection in practice.

\subsection{Different normalization variants}

We perform  experiments with different normalization variants with PI and PD\_AGG and compare them to the baseline feature configurations. Variants include:
\begin{enumerate}
\item \textit{No normalization at all}: we do not provide diagram limits for the computation of the PD. Thus each PD adapts to the value range of the underlying image patch and thus PD\_AGG and PI adapt completely to the value range. 
\item \textit{Globally defined diagram limits}: for each input patch the same diagram limits for PD are applied. This corresponds to the baseline features.
\item \textit{Patch-based normalization with z-standardization}: each input patch is z-standardized individually before feature extraction.
\item \textit{Patch-based normalization with positional standardization}: same as above with positional standardization.
\item \textit{Patch-based normalization with min-max normalization}: same as above with minmax normalization.
\end{enumerate}

The first variant demonstrates what happens if we apply the topological features directly on the data without any adaptions and is expected to represent a lower bound of performance. In variant 2 the depth relations (i.e. the absolute depth differences) between patches are retained. In variants 3-5 depth relations between patches are removed and the only the pure topological information inside each patch is retained.

\begin{figure}%
\centering
	\includegraphics[width=0.95\linewidth]{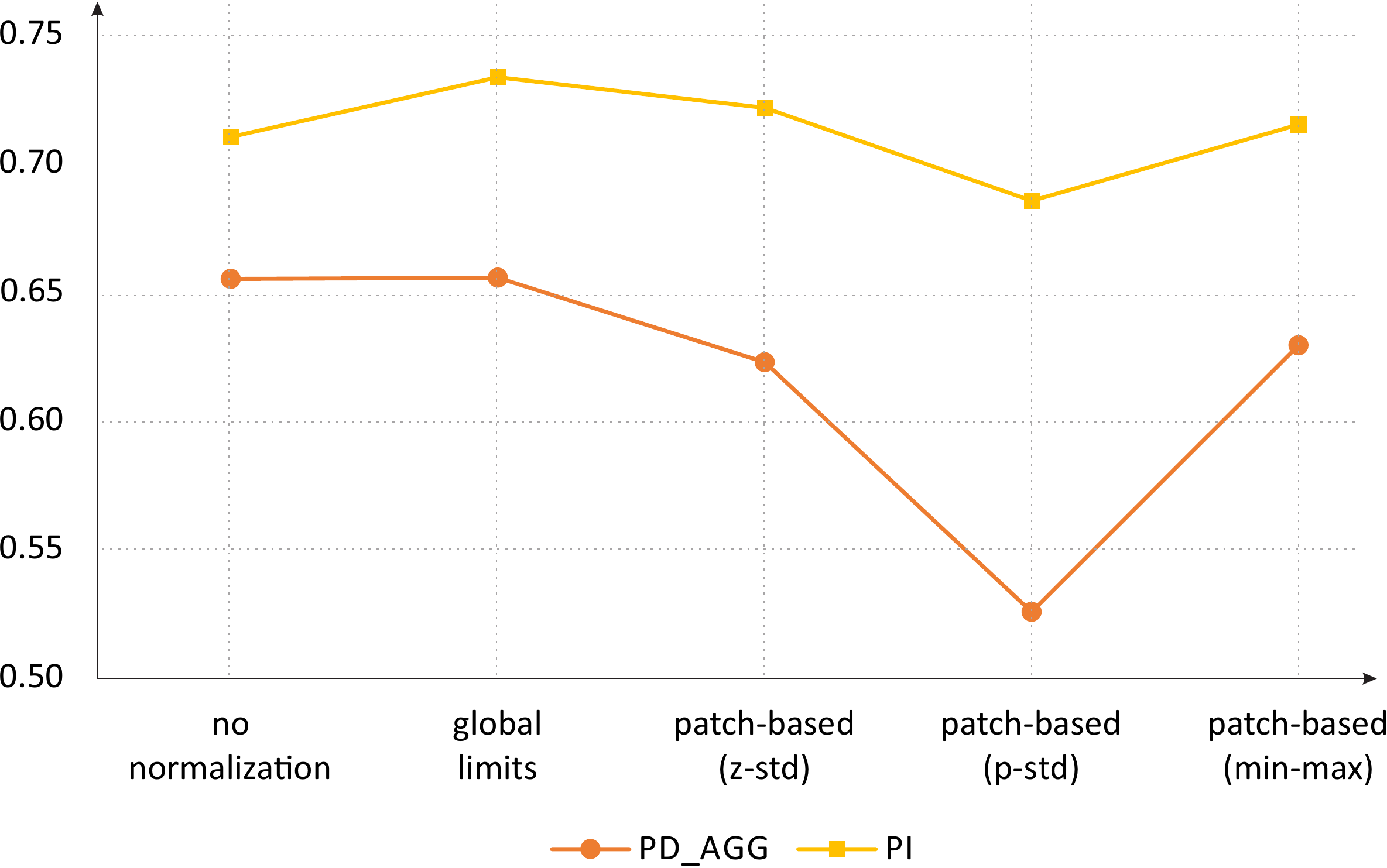}
	\caption{The effect of different normalization methods for PI (yellow line, square markers) and PD\_AGG (orange line with circular markers). Patch-based normalizations (z-std, p-std, and min-max) are outperformed by global normalization. Applying no normalization at all (first experiment) is not beneficial for 3D surface analysis.}
	\label{fig:normalization}
\end{figure}

Results for the different normalization variants are presented in Fig.~\ref{fig:normalization}. The best accuracy is obtained with globally defined diagram limits, which preserve the depth relations between the patches. Applying no normalization at all is suboptimal for PI. The three patch-based normalization variants all yield lower results. This shows that the depth relations between patches are important for the classification task. 

Interestingly, positional standardization which is robust to outliers in the depth distribution leads to the weakest results. This indicates that the outliers in the distribution are of high  importance for the task.

Due to the different normalizations the above descriptors (especially variant 2 and the patch-based normalization variants) may represent complementary information that may be beneficial in combination. To investigate this further, we evaluate different combinations of PD\_AGG and PI with different normalization types. We select patch-based z-standardization in the following experiments. Finally, we investigate the effect of combining PI with PD\_AGG (with similar and different normalizations). For the combination of features we concatenate the respective feature vectors before feeding them into the classifier (early fusion). Results are summarized in Tab.~\ref{tab:variants}.

\begin{table}
  \begin{center}
  \resizebox{1.0\linewidth}{!}{
 	\begin{tabular}{lll}
\textbf{Descriptor(s)} & \textbf{Normalization} & {\textbf{DSC}} \\
\hline
 PD\_AGG & Global limits (baseline) & $0.655\pm0.012$ \\
 PD\_AGG  & Global limits + z-std & $0.671\pm0.008$* \\
\hline
 PI & Global limits (baseline) & $0.733\pm0.002$ \\
 PI & Global limits + z-std & $0.735\pm0.004$ \\
 \hline
 PI+PD\_AGG & Global limits (baselines) & $0.730\pm0.003$ \\
 PI+PD\_AGG & z-std & $0.723\pm0.005$ \\
 PI+PD\_AGG & Global limits + z-std & $0.735\pm0.002$* \\
\hline
\end{tabular}
 } 
	\caption{DSC for combining PI with PD\_AGG (with the same and different normalization types). Asterisks (*) correspond to results significantly better than corresponding baseline topological descriptors with $p<0.05$.}
    \label{tab:variants}
  \end{center}
\end{table}

The only result in Tab.~\ref{tab:variants} where the combination of different normalized descriptors improves results is for PD\_AGG (DSC increases to $0.671\pm0.008$ with $p<0.05$\footnote{Statistical significance is computed with the Wilcox signed rank test, as most of the samples do not pass the Shapiro-Wilk normality test.}). In case of PI only a small improvement by 0.002 is observed which is not significant. 
Interestingly, the combination of PI with PD\_AGG does not improve results over the baseline of PI alone. A reason for this may be that PI implicitly captures the information that is represented in PD\_AGG.

\subsection{Outlier removal}

Outliers are usually considered as unwanted artifacts. In persistence homology, however the role of outliers is different. Outliers in the PD, i.e. points with a large distance to the diagonal, are usually considered as the most meaningful and strong topological components of the underlying data. 

\begin{figure}%
\centering
	\subfigure[]{
	\includegraphics[width=0.31\linewidth]{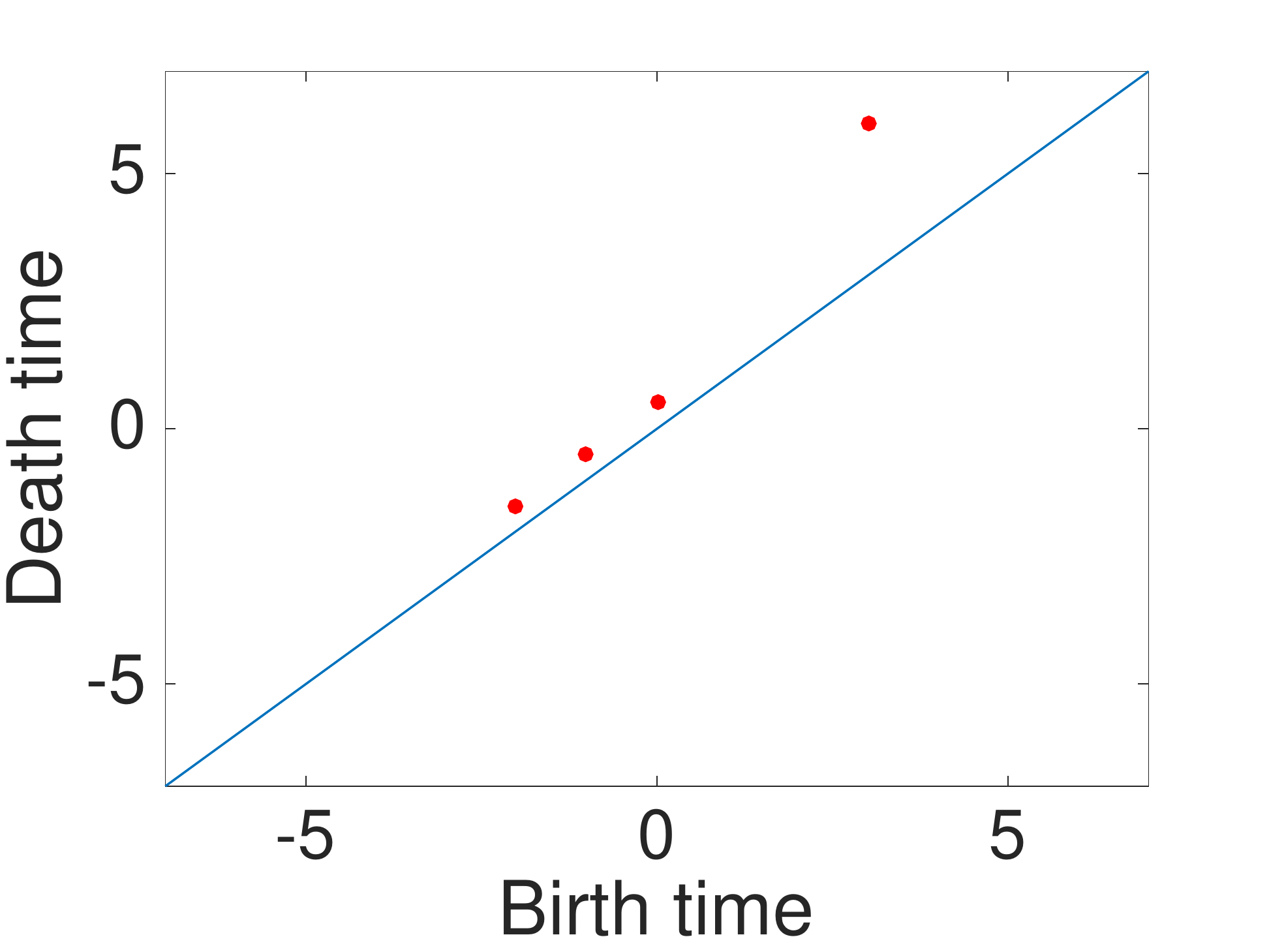}}
	\subfigure[]{
	\includegraphics[width=0.31\linewidth]{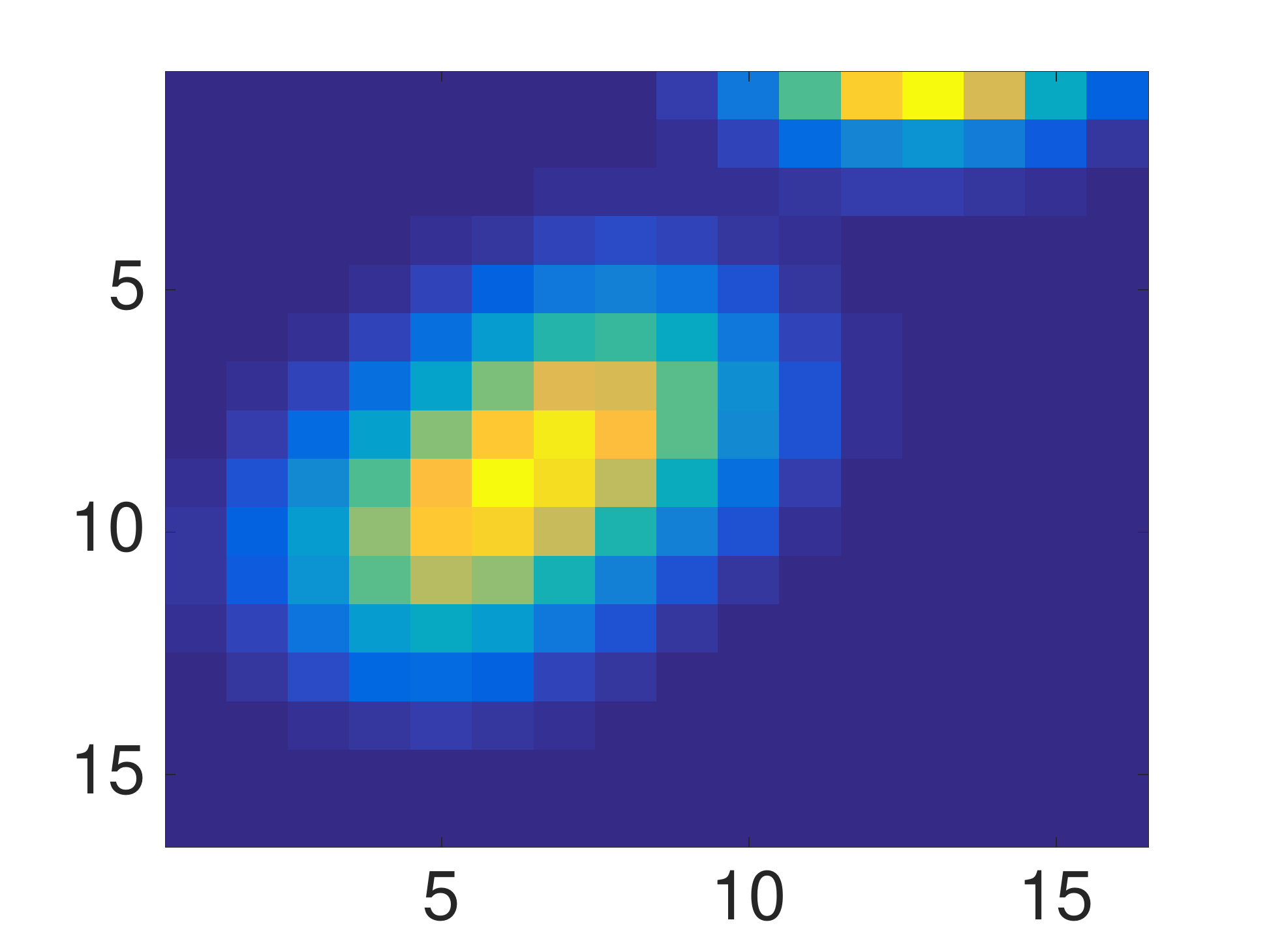}}
	\subfigure[]{
	\includegraphics[width=0.31\linewidth]{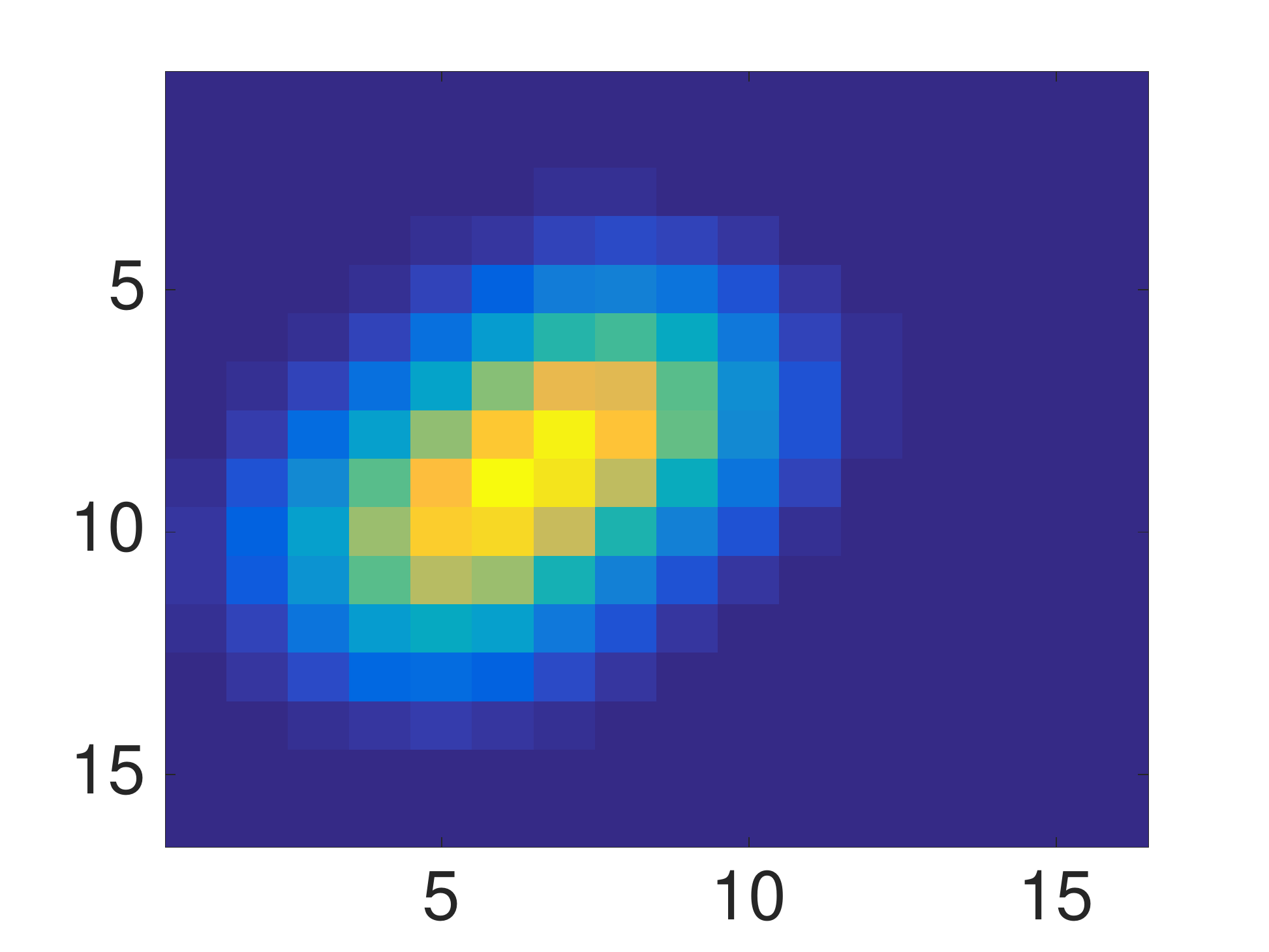}}
	\caption{The effect of outliers on the PI. A PD containing 4 sample points $[-2, -1.5]$, $[-1, -0.5]$, $[0, 0.5]$ and $[3, 6]$ (a) and the corresponding PIs generated without (b) and with (c) outlier removal, with diagram limits $[-5, 5]$. Although the point $[3, 6]$ lies outside of the domain of the PI (and is thus considered an outlier), the Gaussian centered around it still extends into the domain of the PI when outliers are not explicitly removed.}
	\label{fig:outliers}
\end{figure}

In our approach outliers are those points in the PD which extend the pre-defined diagram limits. There are two alternatives: to keep them or to remove these outliers completely from the further computation. For the two investigated topological descriptors this has different consequences. In the case of PD\_AGG the outliers are simply included in the derived statistics, which is the same as extending the diagram limits to the farthest outliers. For PI this means that although the outliers lie outside of the domain of the PI the Gaussian centered around an outlier during PI construction may still extend into the domain of the PI (see Fig.~\ref{fig:outliers}c).

We compare the effect of outlier removal by computing respective variants of our baseline features. The results show that removing outliers decreases results for PI from $0.733\pm0.003$ to $0.731\pm0.003$ and for PD\_AGG from $0.655\pm0.012$ to $0.639\pm0.011$. In case of PD\_AGG the decrease is significant with $p<0.05$). This indicates that outliers play an important role for the topological descriptors.

\subsection{Weighting in PI computation}
\label{subsec:weighting}

The original formulation of PI includes a weighting function for the individual Gaussian components which grows exponentially with distance to the diagonal of the PD (function $g$ in Equation~\ref{eq:PI} in Section~\ref{subsec:PI}). The intuition behind this is to emphasize stronger topological components in the PD (further away from the diagonal) and to attenuate the influence of noise which is often located close to the diagonal. Previous experiments have shown that the weighting is in practice not always beneficial~\cite{zeppelzauer2016topological}. To further investigate the role of weighting in the PI computation, we compare the performance of PI with different weighting functions namely: constant weighting, linear weighting and exponential weighting.

Experiments reveal that weighting significantly decreases performance with $p<0.005$: from $0.733\pm0.003$ to $0.714\pm0.008$ in case of linear weighting and to $0.715\pm0.007$ in case of exponential weighting. Best results are obtained when all Gaussians are equally weighted. This indicates that short intervals close to the diagonal play an important role in the context of surface texture description.
This contradicts with the original assumptions made in the design of PI. We investigate this aspect further in Section~\ref{subsec:discriminativity}.

\subsection{Pre-filtering}
\label{subsec:preFilteringExp}
We investigate different ways to pre-filter the surface patches before extracting topological features. This has recently been proposed to obtain more robust and expressive representations~\cite{li2014persistence,reininghaus2014stable}. We employ three different methods for pre-filtering (as described in Section~\ref{sec:preprocessing}): (i) Schmid filter bank~\cite{schmid2001constructing}, (ii) MR filter bank~\cite{geusebroek2003fast} and (iii) CLBP encoding~\cite{guo2010completed}.

For MR and Schmid filter banks we employ the default parameters proposed by the respective authors. The only exception is that in case of Schmid filter banks we use two times smaller $\tau$, as in preliminary experiments there is hardly any difference in the filter responses for $\tau \geqslant 2$. Each patch is filtered with all filters from the respective filter bank. For each filtered patch topological features are extracted separately and are then concatenated into one high-dimensional feature vector.

Additionally, we extract CLBP features~\cite{guo2010completed} from the raw patches as proposed in~\cite{li2014persistence} and~\cite{reininghaus2014stable} and then extract topological descriptors from the CLBP\_S and CLBP\_M maps. For CLBP computation we employ \review{the implementation of \cite{CLBPcode}} with different encodings: rotation invariant LBP (ri) and rotation invariant uniform LBP (riu2) as well as different numbers of neighborhood samples $n=\left\{8,16\right\}$ and different radii $r=\left\{3,5\right\}$.

\begin{table*}
  \begin{center}
  \resizebox{0.6\textwidth}{!}{
 	\begin{tabular}{llllrr}
\textbf{Descriptor} & \textbf{Pre-filtering} & & & \multicolumn{2}{c}{\textbf{DSC}} \\
\hline
PD\_AGG & (baseline) & & & & $0.655\pm0.012$ \\
PD\_AGG & Schmid & & & & $0.643\pm0.023$ \\
PD\_AGG & MR & & & & $0.607\pm0.044$ \\
&  & & & \multicolumn{1}{c}{$n=8$} & \multicolumn{1}{c}{$n=16$} \\
PD\_AGG & CLBP & riu2 & $r=3$ & $0.671\pm0.004$** & $0.674\pm0.009$** \\
& CLBP & & $r=5$ & $\mathbf{0.708\pm0.005}$** & $0.678\pm0.008$** \\
& CLBP & ri & $r=3$ & $0.697\pm0.006$** & $0.670\pm0.019$* \\
& CLBP & & $r=5$ & $0.700\pm0.006$** & $0.687\pm0.006$** \\
\hline
PI & (baseline) & & & & $\mathbf{0.733\pm0.002}$ \\
PI & Schmid & & & & $0.702\pm0.007$ \\
PI & MR & & & & $0.698\pm0.004$ \\
  &  & & & \multicolumn{1}{c}{$n=8$} & \multicolumn{1}{c}{$n=16$} \\
PI & CLBP & riu2 & $r=3$ & $0.712\pm0.014$ & $0.725\pm0.004$ \\
& CLBP & & $r=5$ & $0.707\pm0.014$ & $0.719\pm0.006$ \\
& CLBP & ri & $r=3$ & $0.709\pm0.009$ & $0.722\pm0.007$ \\
& CLBP & & $r=5$ & $0.706\pm0.006$ & $0.706\pm0.005$ \\
\hline
\end{tabular}
 } 
	\caption{Results for pre-filtering depth maps before extraction of topological features. Asterisks (*) and (**) correspond to results significantly better than corresponding baseline topological descriptor with $p<0.05$ and $p<0.01$, respectively. Bold represent the best results for PD\_AGG and PI.}
    \label{tab:prefiltering}
  \end{center}
\end{table*}

The results of experiments are presented in Tab.~\ref{tab:prefiltering}. It can be observed that prefiltering with Schmid and MR filter banks does not improve performance, both for PD\_AGG and PI. CLBP, however, improves DSC significantly in case of PD\_AGG (from $0.655\pm0.012$ to $0.708\pm0.005$, $p<0.01$ and CLBP with ``riu2" encoding, $r=5$ and $n=8$). The effect of pre-filtering is further analyzed in Section~\ref{subsec:robustness}. From the experiments so far it seems that for PI the best strategy is to apply it on the raw data directly, while for PD\_AGG which are much simpler and coarser descriptors the additional pre-filtering can be beneficial.

\begin{figure}%
\centering
	\includegraphics[width=0.95\linewidth]{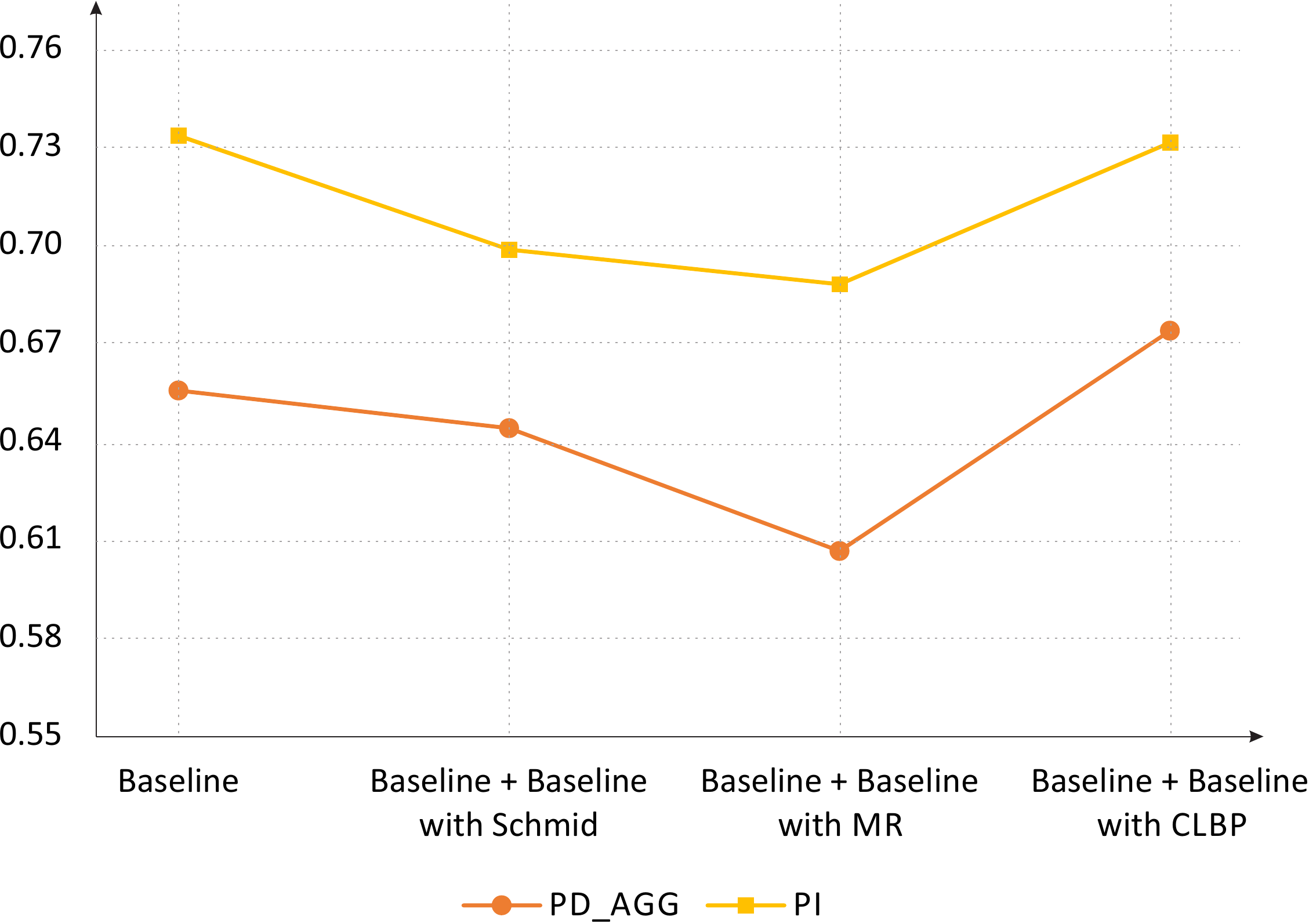}
	\caption{DSC for combining baseline topological descriptors with the respective Schmid, MR and CLBP filtered variants. For PI the combination is not able to further improve results. PD\_AGG benefits from adding the CLBP-filtered variant but not significantly.}
	\label{fig:preFilterCombi}
\end{figure}

Finally, we combine our baseline features with the respective Schmid, MR and CLBP filtered variants by early fusion (concatenation) to investigate if the descriptors can benefit from each other. Experiments show that most of the combinations decrease performance (see Fig.~\ref{fig:preFilterCombi}), which is consistent with the pre-filtering results presented above. The only positive effect in performance is achieved by the combination of PD\_AGG with the respective CLBP filter variant. This is also consistent with the previous results where CLBP improved PD\_AGG. The result of the combined descriptors is, however, weaker than that of PD\_AGG based on CLBP alone (0.674 vs 0.708). This shows that there is no synergy attained by the combination.

\subsection{Feature selection}
\label{subsec:featureSel}
By applying pre-filtering (as in the previous Section~\ref{subsec:preFilteringExp}) or by using PIs with high resolutions the dimension of the resulting feature vectors grows linearly and quadratically, respectively. Thus, strategies to reduce the dimension  may be useful in practice. We apply different feature selection strategies to the PI descriptors to investigate to which degree dimensionality reduction affects classification performance and how much redundancy the descriptors exhibit. We apply the following three strategies:
\begin{enumerate}
\item \textit{Gini importance}: \review{we select Gini importance~\cite{breiman2001random} as a strategy because it is a by-product of the employed classifier (RUSBOOST, see Section \ref{sec:setup}) and it can be obtained without additional cost. It directly reflects the importance of each descriptor entry for distinguishing between the classes.} \newline Computation: we compute the Gini importance for each entry of the PI descriptor during training of the classifier and keep only the N most important (highest values) entries. 
\item \textit{Fisher criterion}: \review{While Gini takes dependencies between descriptor entries into account (through the classifier tree hierarchy), the Fisher criterion~\cite{duda1973pattern} is completely independent from the classifier and evaluates the importance of every entry independently. For this reason we decided to use is as a complementary strategy.} \newline 
Computation: we compute the Fisher criterion for each descriptor entry during training of the classifier and keep only the N most discriminative entries (highest Fisher criterion)
\item \textit{Combined Gini and Fisher criterion}: \review{we combine both strategies to see if there are synergy effects between them.} \newline Computation: for each descriptor entry we compute Gini importance and Fisher criterion and average the rankings obtained by both measures. The N entries with the highest combined ranking are retained.
\end{enumerate}

Additionally, to provide a baseline, we select features randomly to investigate the actual benefit of feature selection strategies.

\begin{figure}%
\centering
	\includegraphics[width=0.95\linewidth]{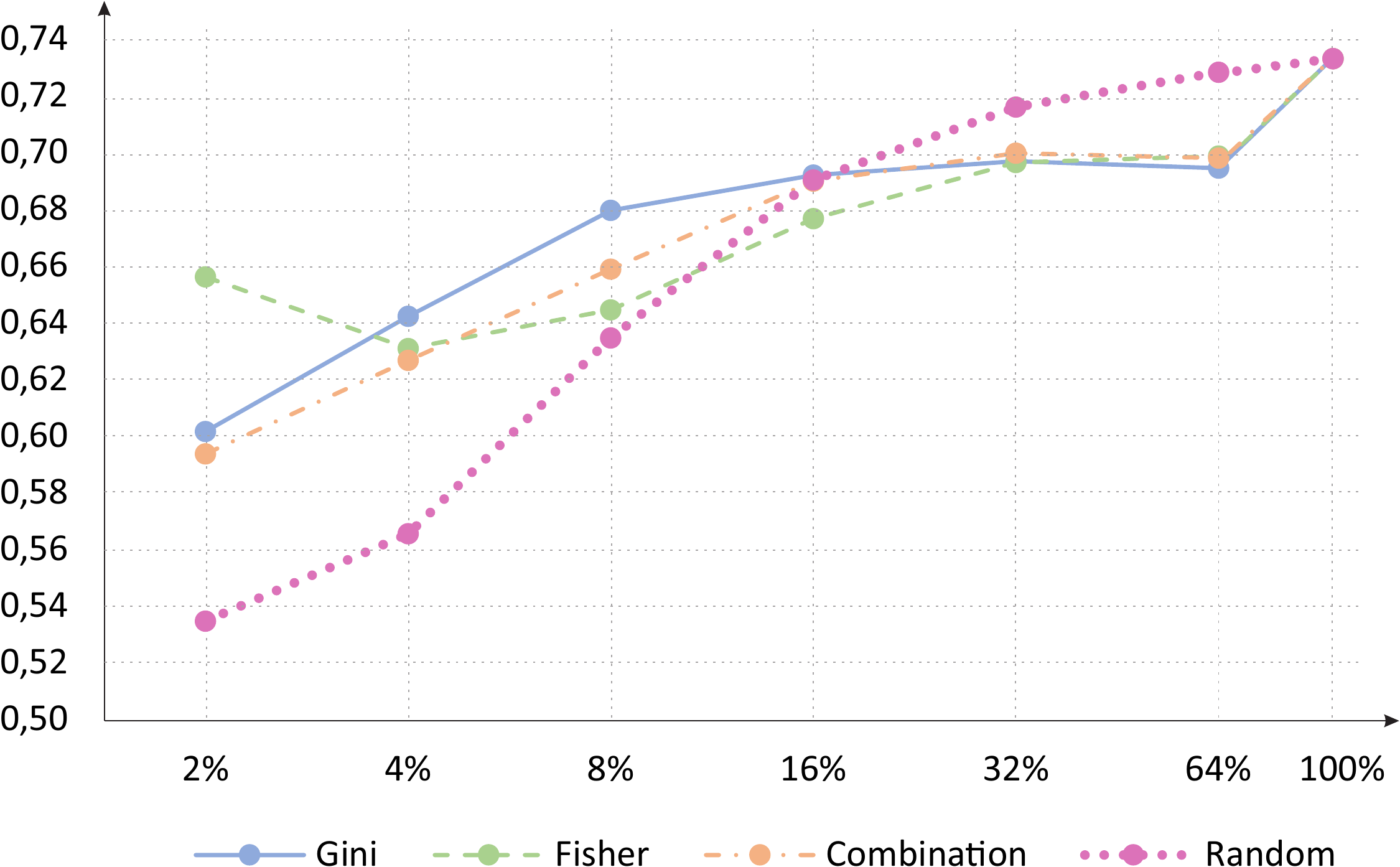}
	\caption{Classification performance (DSC) with different feature selection strategies for differently large subsets of the baseline PI descriptor.}
	\label{fig:featureSelection}
\end{figure}

For all three feature selection approaches we vary N from 2, 4, 8, 16, 32, 64 to 100\% of the PI descriptor (136 dimensions). Results are illustrated in Fig.~\ref{fig:featureSelection}. Note, that the x-axis in Fig.~\ref{fig:featureSelection} is not linear. By selecting only 2\% of the components, Gini yields a DSC of $0.6015\pm0.0197$. With the Fisher criterion even a DSC of $0.6560\pm0.0213$ is achieved. Both values strongly outperform the random baseline ($0.535\pm0.06$). With increasing number of features the performance increases continuously for Gini. The Fisher criterion is less stable and is outperformed in most cases by Gini as can be seen in Fig.~\ref{fig:featureSelection}. A reason for the better performance of Gini is that Gini is computed during classifier training and expresses the importance of a given entry for the classifier training. Fisher, in contrast, is computed independently from the classifier from the training data directly. The combined scheme is located approximately between the two approaches which may stem from averaging the respective rankings of the individual measures. Interestingly, feature selection is only reasonable up to a certain number of features. By selection of more than 16\% of features the random baseline outperforms the selection strategies. The reason for this behavior is that the feature selection strategies do not take the inter-correlations between the entries into account and thus redundant components are easily selected when the number of selected features increases. We further investigate the importance and discriminativity of individual entries in Section~\ref{subsec:discriminativity}.

\subsection{Comparison with non-topological features}
\label{subsec:nonTopoFeatures}

From the patches of the depth map we extract a number of non-topological image descriptors in the same patch-wise manner as the topological descriptors. Non-topological descriptors include: MPEG-7 Edge Histogram (EH)~\cite{mpeg7standard}, Dense SIFT (DSIFT)~\cite{lowe2004distinctive}, Local Binary Patterns (LBP)~\cite{ojala1996comparative}, Histogram of Oriented Gradients (HOG)~\cite{dalal2005histograms}, Gray-Level Co-occurrence Matrix (GLCM)~\cite{haralick1973textural}, Global Histogram Shape (GHS)~\cite{zeppelzauer2015efficient}, Spatial Depth Distribution (SDD)~\cite{zeppelzauer2015efficient}, as well as enhanced versions of GHS and SDD (short EGHS and ESDD) that apply additional enhancements to the depth map described in~\cite{zeppelzauer2015efficient}.

\review{For the extraction of non-topological features, we employ publicly available implementations if available, such as VLFEAT for SIFT and HOG\cite{vedaldi08vlfeat}, the LBP implementation of \cite{LBPcode}, and the GLCM implementation of \cite{GLCMcode}. GHS, SDD, EGHS, and ESDD are implemented as in~\cite{zeppelzauer2015efficient}.}

Results for the individual descriptors are shown in Fig.~\ref{fig:nonTopoFeatures} (blue line). The best individual non-topological descriptor is ESDD (DSC of $0.743\pm0.002$) followed by the combination of EGHS+ESDD (DSC of $0,728\pm0.005$). The performance of our two baseline descriptors are shown with a dotted line (PD\_AGG) and a dashed line (PI) in black for comparison. The PD\_AGG descriptor achieves similar results to LBP, EH, GLCM, and EGHS. Remarkably, PI alone already outperforms all non-topological features except ESDD. This attests a strong explanatory power of the PI for the description of surface texture while PD\_AGG rather lacks in expressiveness.

\subsection{Combination with non-topological features}
\label{subsec:combiTopoNonTopo}

\begin{figure}%
\centering
	\includegraphics[width=0.95\linewidth]{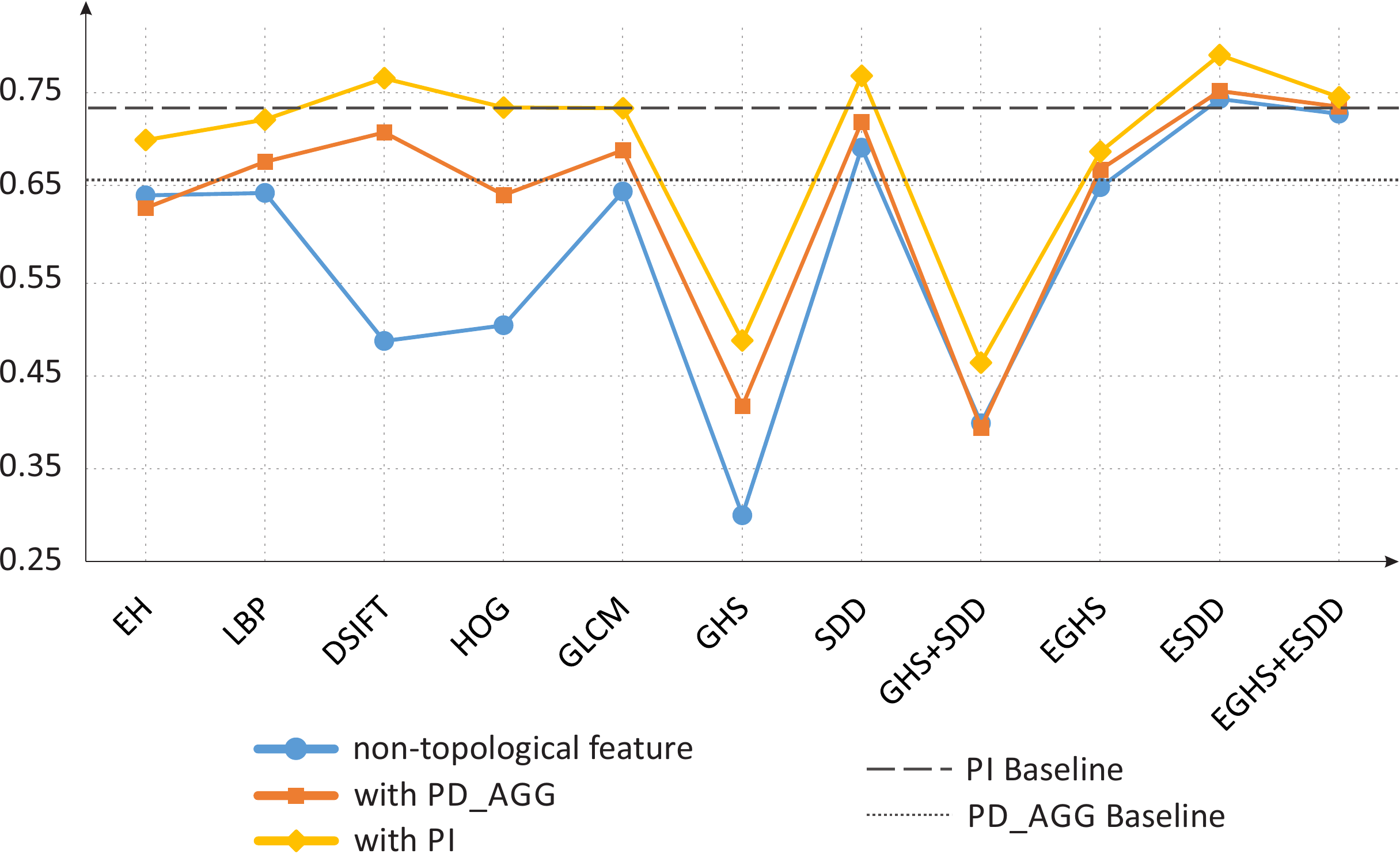}
	\caption{Classification results (DSC) for different non-topological descriptors (blue line with circular markers). The orange (square marker) and yellow (diamond marker) lines represent the performance when the features are combined with PD\_AGG and PI, respectively. The baselines for PD\_AGG and PI are plotted as dotted and dashed lines, respectively.}
	\label{fig:nonTopoFeatures}
\end{figure}

\begin{figure*}%
\centering
	\subfigure[Surface]{
	\includegraphics[width=0.31\linewidth]{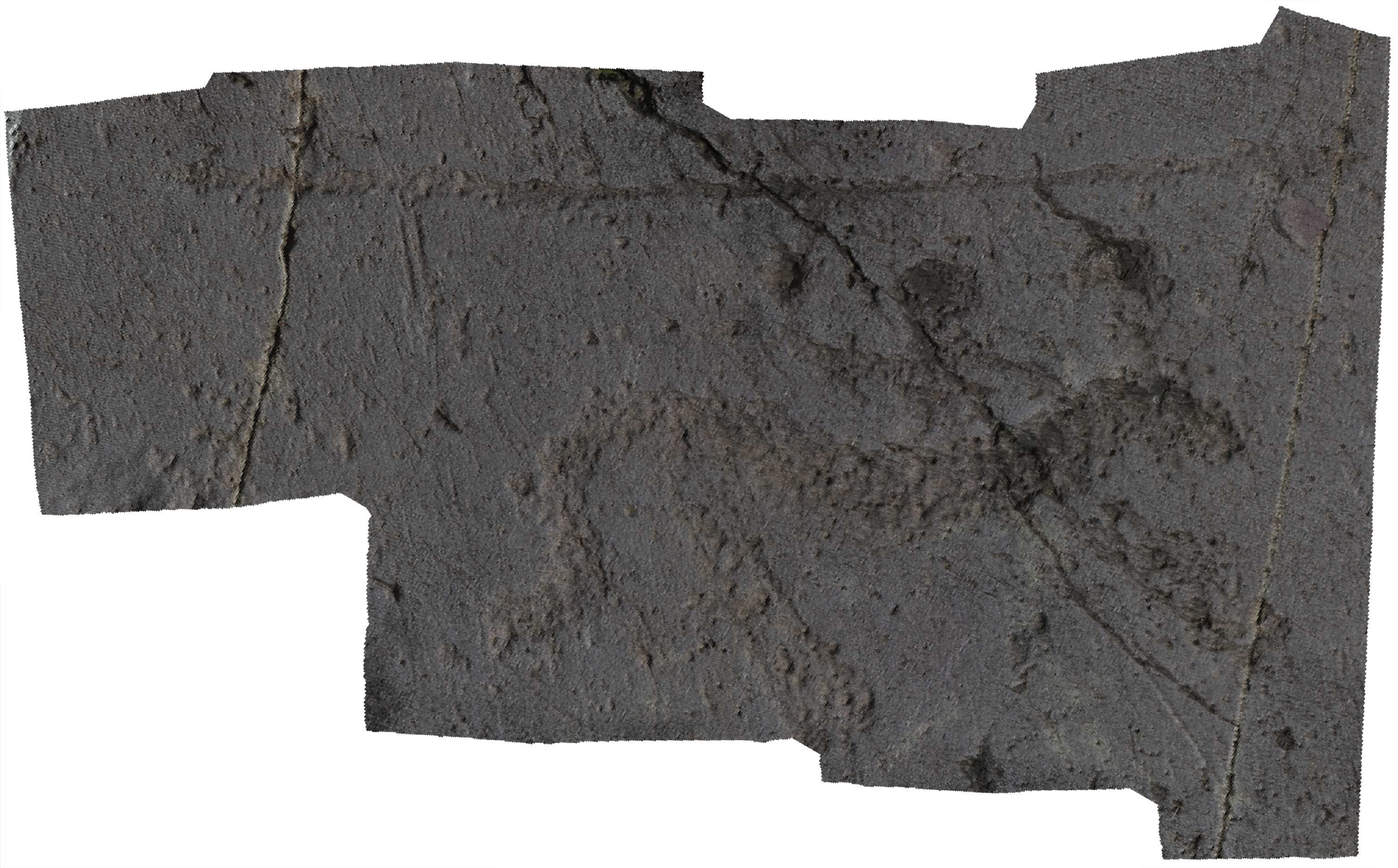}}
	\subfigure[Depth map]{
	\includegraphics[width=0.31\linewidth]{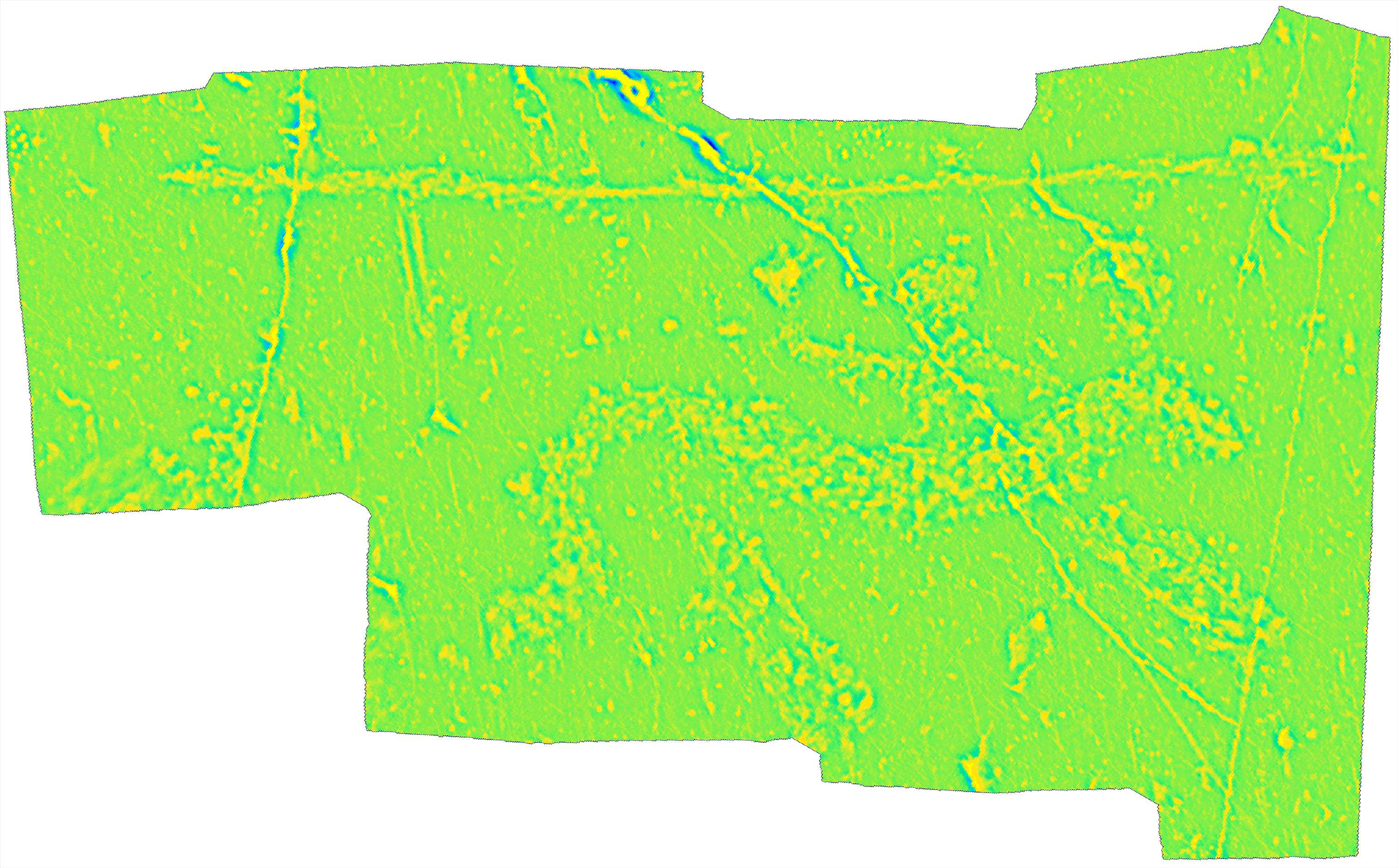}}
	\subfigure[Expert ground truth]{
	\includegraphics[width=0.31\linewidth]{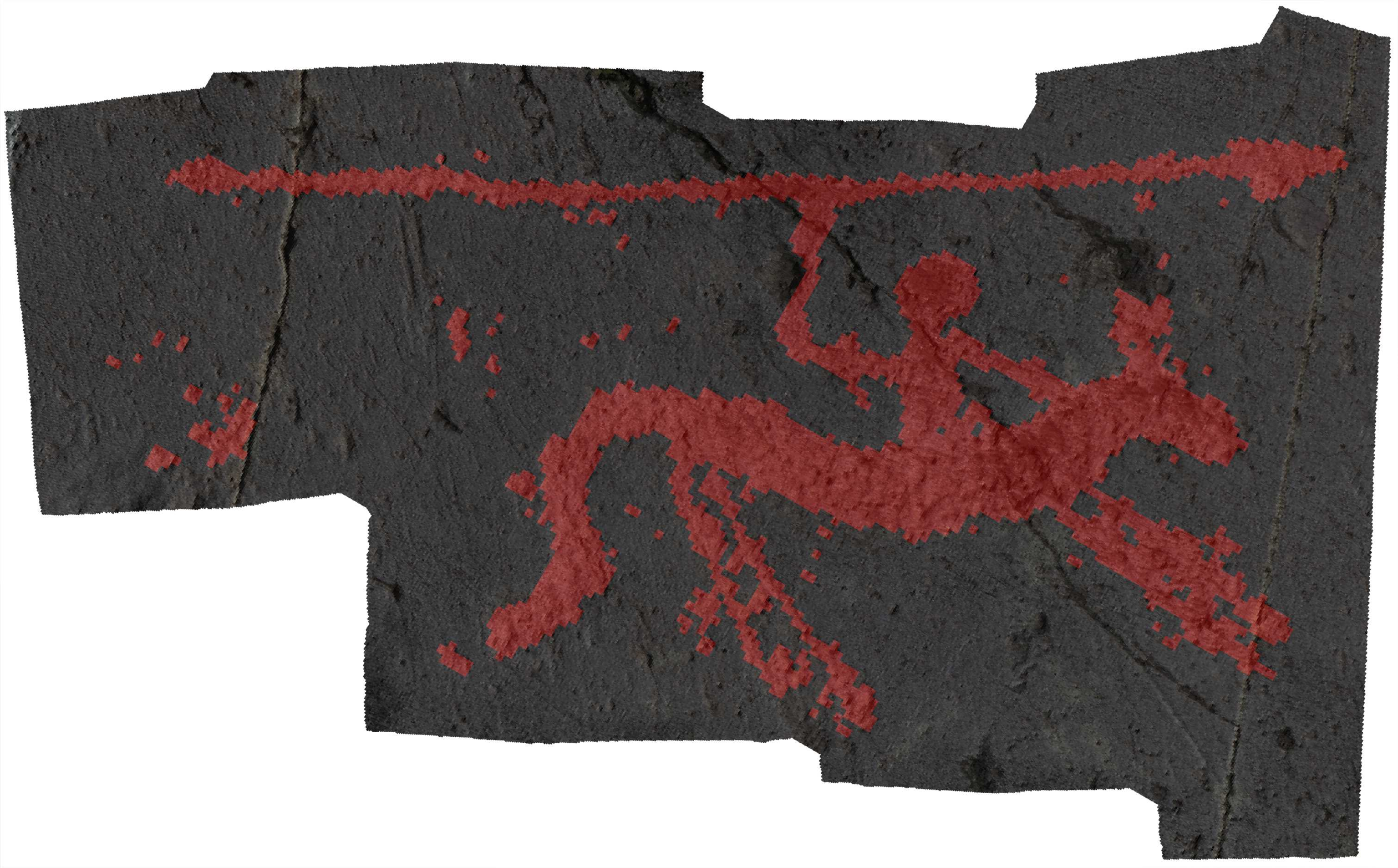}}
	\subfigure[Result ESDD]{
	\includegraphics[width=0.31\linewidth]{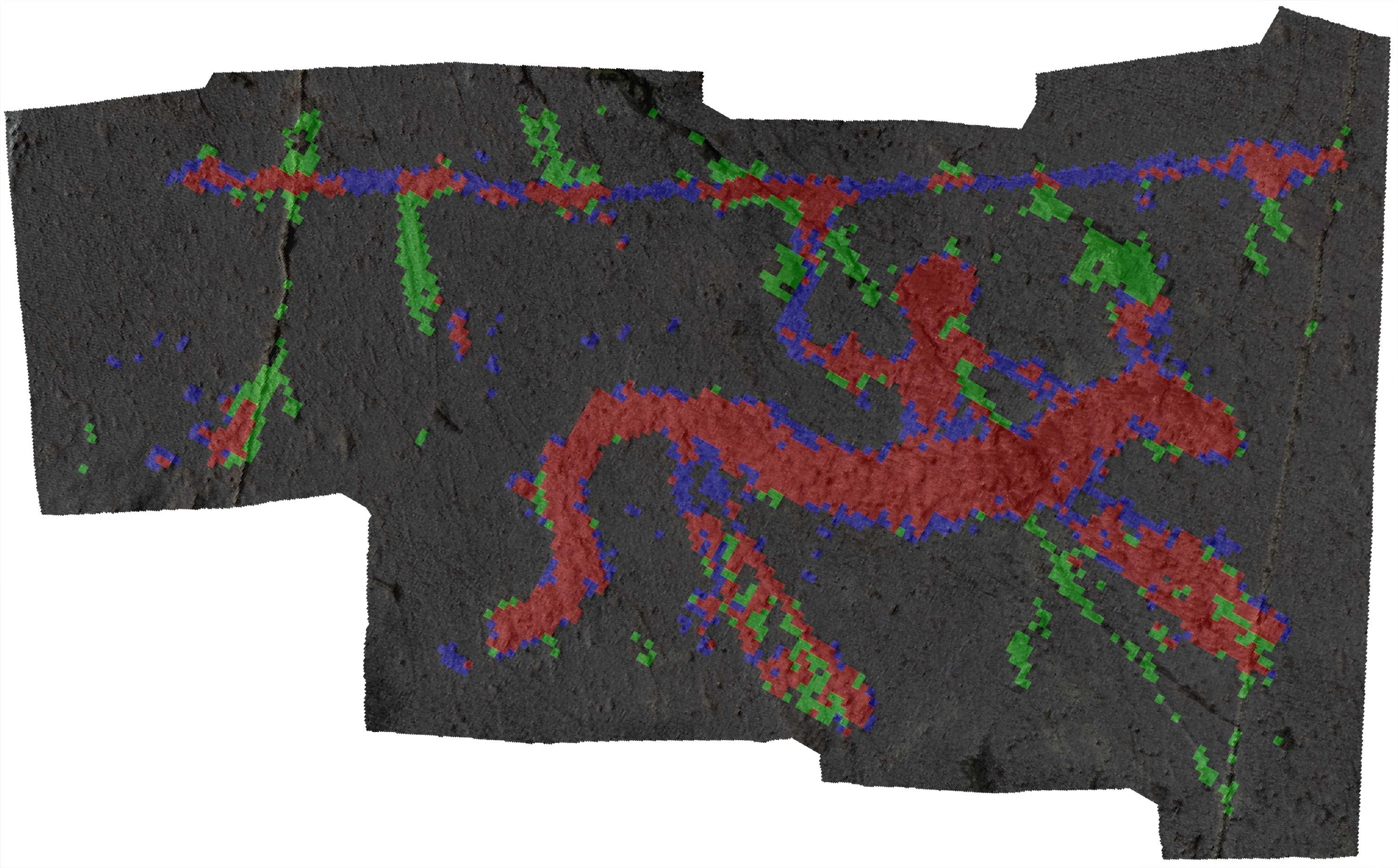}}
	\subfigure[Result PI]{
	\includegraphics[width=0.31\linewidth]{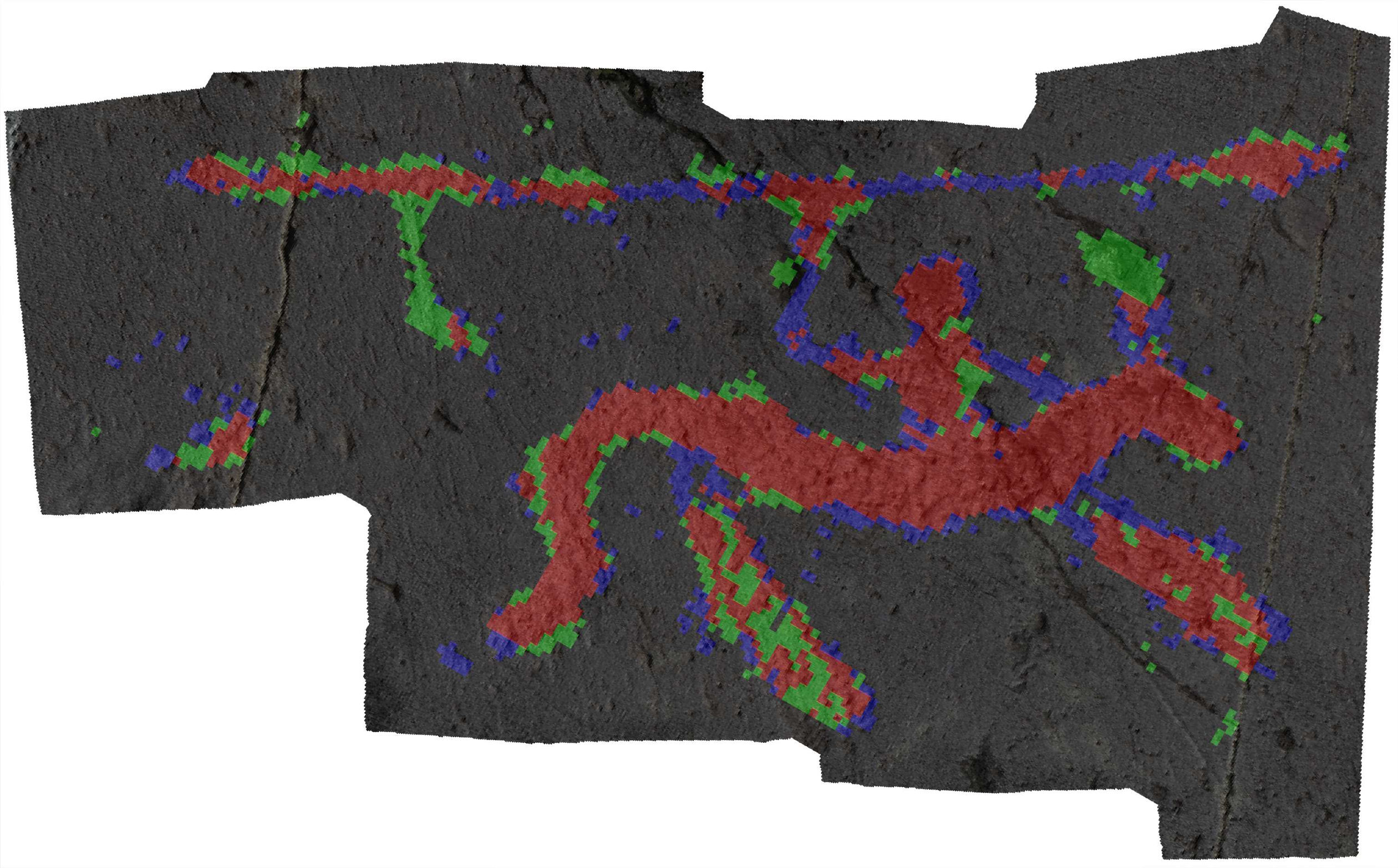}}
	\subfigure[Result ESDD+PI]{
	\includegraphics[width=0.31\linewidth]{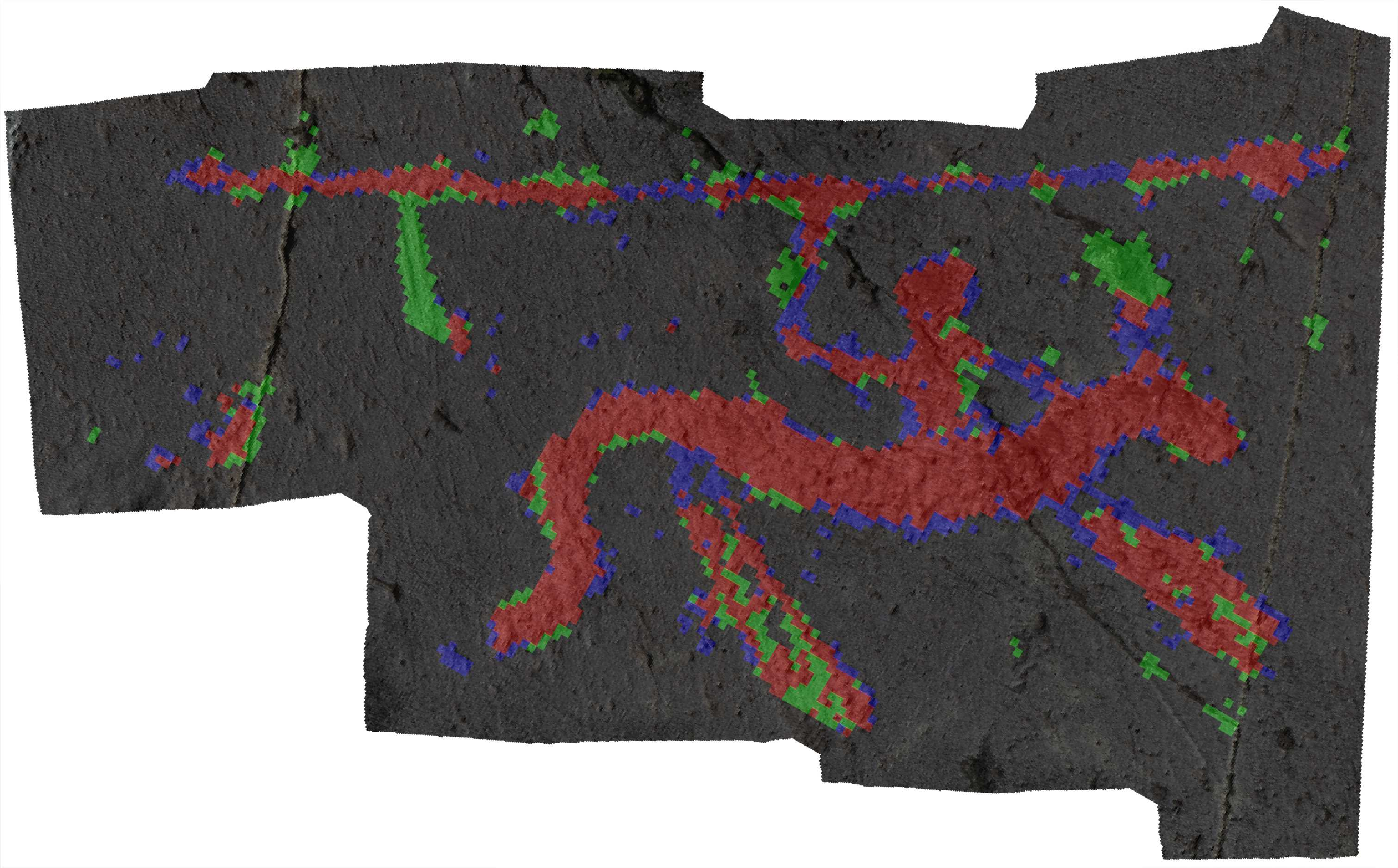}}
	\caption{\review{Classification results for a surface from the test set: (a) the original rock surface; (b) the depth map of the surface (in false color and contrast enhanced); (c) the ground truth labeling obtained by domain experts (engraved areas are highlighted red); (d,e,f) show the classification results of ESDD, PI and the combination of both features. Red areas correspond to correctly classified engraved areas (class 1), green areas are false positives detections of class 1 and blue refers to false negatives (non-detected engraved areas). All remaining (uncolored) pixels are true positives for class 2.}}
	\label{fig:qualitativeResult}
\end{figure*}

In previous investigations the addition of topological descriptors to other (non-topological) descriptors has improved classification performance~\cite{zeppelzauer2016topological}. We investigate the effect of combining both types of descriptors by adding the best configurations of PI and PD\_AGG to the evaluated non-topological descriptors from Section~\ref{subsec:nonTopoFeatures}. The results are shown as additional curves in Fig.~\ref{fig:nonTopoFeatures}. The orange curve (square markers) shows the performance when non-topological descriptors are combined with PD\_AGG and the yellow curve (diamond-shaped markers) shows the performance in combination with PI. The combination with PD\_AGG improves all descriptors except for EH. PI improves all descriptors and some in fact by a large margin. Even the performance of the best non-topological descriptor (ESDD) can be further improved when combined with PI (from $0.743\pm0.002$ to $0.788\pm0.005$ with $p<0.01$). \review{A corresponding example result is shown in Figure \ref{fig:qualitativeResult}. The surface shows a horse with a rider that holds a long spear in its hand. The first row shows the original rock-surface, the depth map, and the ground-truth labeling (engraved areas in red, remaining areas uncolored). The second row shows the results of ESDD, PI and ESDD+PI. Classification results are color coded: red refers to true positive detections of class 1 (engraved area), green to false positive detections of class 1 and blue to false negative detections of class 1. Uncolored areas are true positive detections of class 2. The results for ESDD and PI are comparable in overall quality but show different strengths and weaknesses of the descriptors. ESDD better captures fine details than PI (e.g. the rear legs) but produces more false detections (green) at natural surface irregularities like the vertical and diagonal cracks in the rock surface. PI is more robust to such irregularities and captures the shape of the engraved ares well but generates more false detections along the boundary of the engraved areas (e.g. along the spear and the horse's tail). In combination the mutual weaknesses compensate each other (e.g. the shaft of the spear and the arm of the rider are better detected) and yield a more accurate overall segmentation.} 
These results illustrate well that topological descriptors represent complementary information about the surface texture compared to traditional texture descriptors and that their combination is beneficial.

\subsection{Large-scale experiments}

Experiments so far were performed on the small-scale dataset which was chosen to save computation time and in turn gave us the opportunity to broaden our evaluation to many different aspects, parameter values, configurations etc. A central question in our evaluation is how topological descriptors compete with non-topological descriptors which currently represent the state-of-the-art for 3D surface and texture analysis \cite{poier2016petrosurf3d}. To answer this question more profoundly, we run the experiments from Sections~\ref{subsec:nonTopoFeatures} and~\ref{subsec:combiTopoNonTopo} on the large-scale dataset from Section~\ref{sec:setup}. We evaluate all descriptors without further parameter tuning to obtain objective and unbiased results, i.e. no fine tuning is performed on the large-scale dataset at all. The topological descriptors are then compared to a selection of the most promising non-topological descriptors. Furthermore, we evaluate the performance of the best descriptor combinations evaluated so far. 

\begin{figure}%
\centering
	\includegraphics[width=0.96\linewidth]{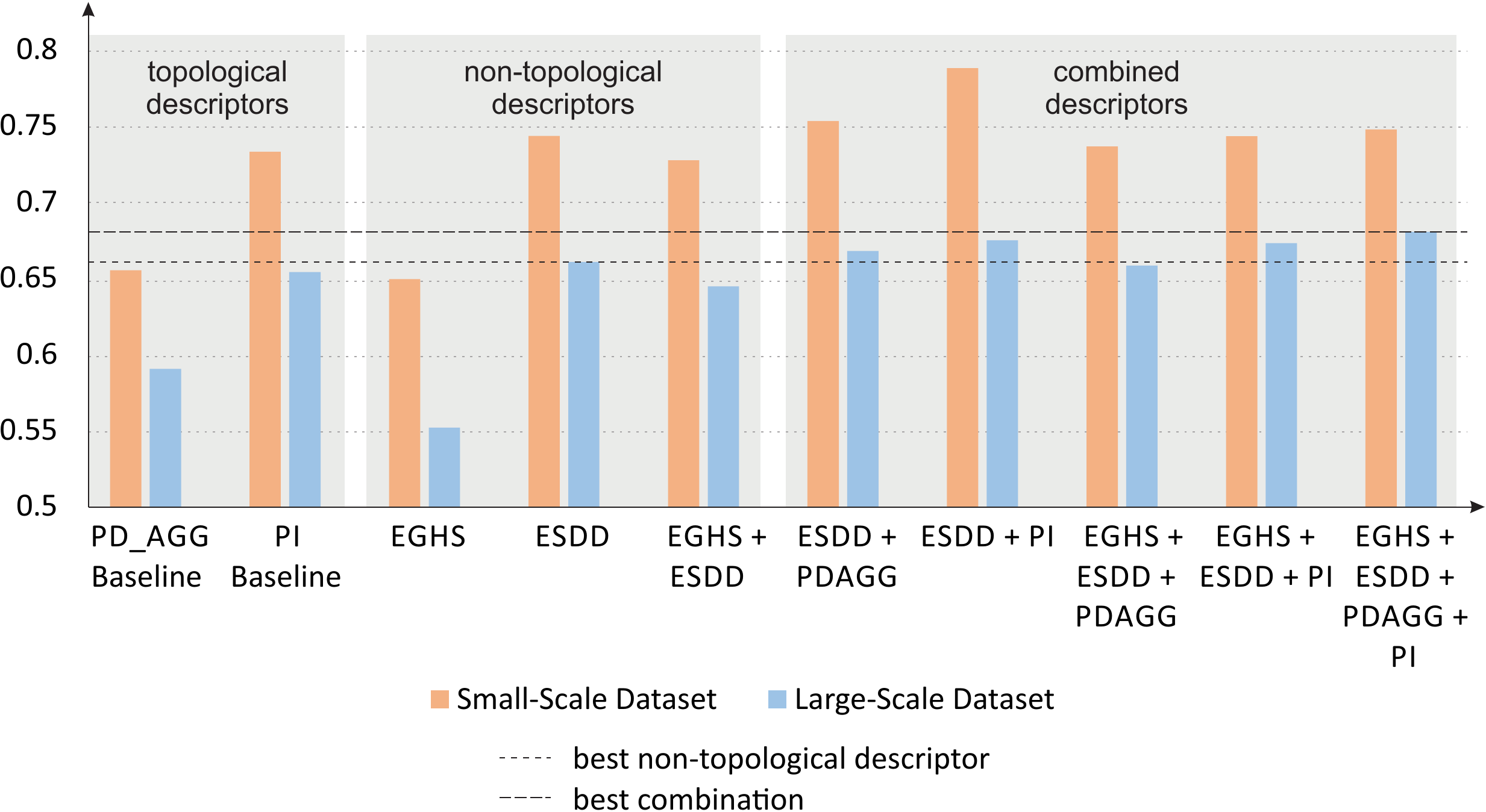}
	\caption{Performance comparison between small-scale and large-scale dataset in terms of DSC. Also for the large-scale dataset the topological descriptors add additional information and are necessary to obtain peak performance.}
	\label{fig:smallVsLargeScale}
\end{figure}

Results for the large-scale dataset (together with the corresponding results for the small-scale dataset) are presented in Fig.~\ref{fig:smallVsLargeScale}. The overall performance of all descriptors drops for the large-scale dataset. This is expected and results from the higher complexity of the dataset (more heterogeneous data). The absolute performance difference between the two datasets is, however, quite consistent between all descriptors ($\Delta=0.082$ with a quite low standard deviation of $\pm0.014$), which shows that the classifier can generalize well from all descriptors. The best topological descriptor is again PI with $0.654\pm0.002$ which is similar to the best non-topological descriptor (ESDD) with $0.661\pm0.001$ (dashed line).  
The smaller difference in peak performance between topological and non-topological descriptors compared to the small-scale dataset shows that the topological descriptors better generalize to the more complex data in the large dataset.

The combination of topological and non-topological descriptors yields a performance gain in most cases. Peak performance is obtained when all evaluated descriptors are combined: $0.681\pm0.002$. This improvement is significant with $p<0.01$ compared to the best combination of non-topological descriptors. The two horizontal lines in Fig.~\ref{fig:smallVsLargeScale} represent the peak performance without and with topological information. The vertical spacing between the lines represents the gain obtained by adding topological information in classification ($+0.02$). Note that the gain for the small-scale dataset is even larger ($+0.045$).

We further compare our results with those obtained in \cite{poier2016petrosurf3d} on the same dataset. In contrast to the investigation in \cite{poier2016petrosurf3d} which uses 4-fold cross-validation on the entire dataset, we employ 5-fold cross-validation on the training set and additionally use half of the dataset as a completely independent test set. The use of an independent test set makes our evaluation protocol is more demanding that that of  \cite{poier2016petrosurf3d}. Reference \cite{poier2016petrosurf3d} reports results for two approaches: Random Forests (RF) and Convolutional Neural Networks (CNNs). \review{For the CNN the fully convolutional net by \cite{Long2015cvpr} has been fine-tuned on randomly selected patches from the training data, see \cite{poier2016petrosurf3d} for details}. RF yields a DSC of $0.568$. CNN outperforms RF with a DSC of $0.667$. With a DSC of $0.654$ our best topological feature (PI) yields a similar performance level as CNNs in \cite{poier2016petrosurf3d}. PI combined with non-topological descriptors even outperforms the results of the CNN with a peak performance of $0.681$. Thus, the performance of the investigated descriptors can be considered state-of-the-art. 

\subsection{Robustness and sensitivity}
\label{subsec:robustness}

\begin{figure}%
\centering
	\subfigure[]{\label{patch:snr5}
	\includegraphics[width=0.31\linewidth]{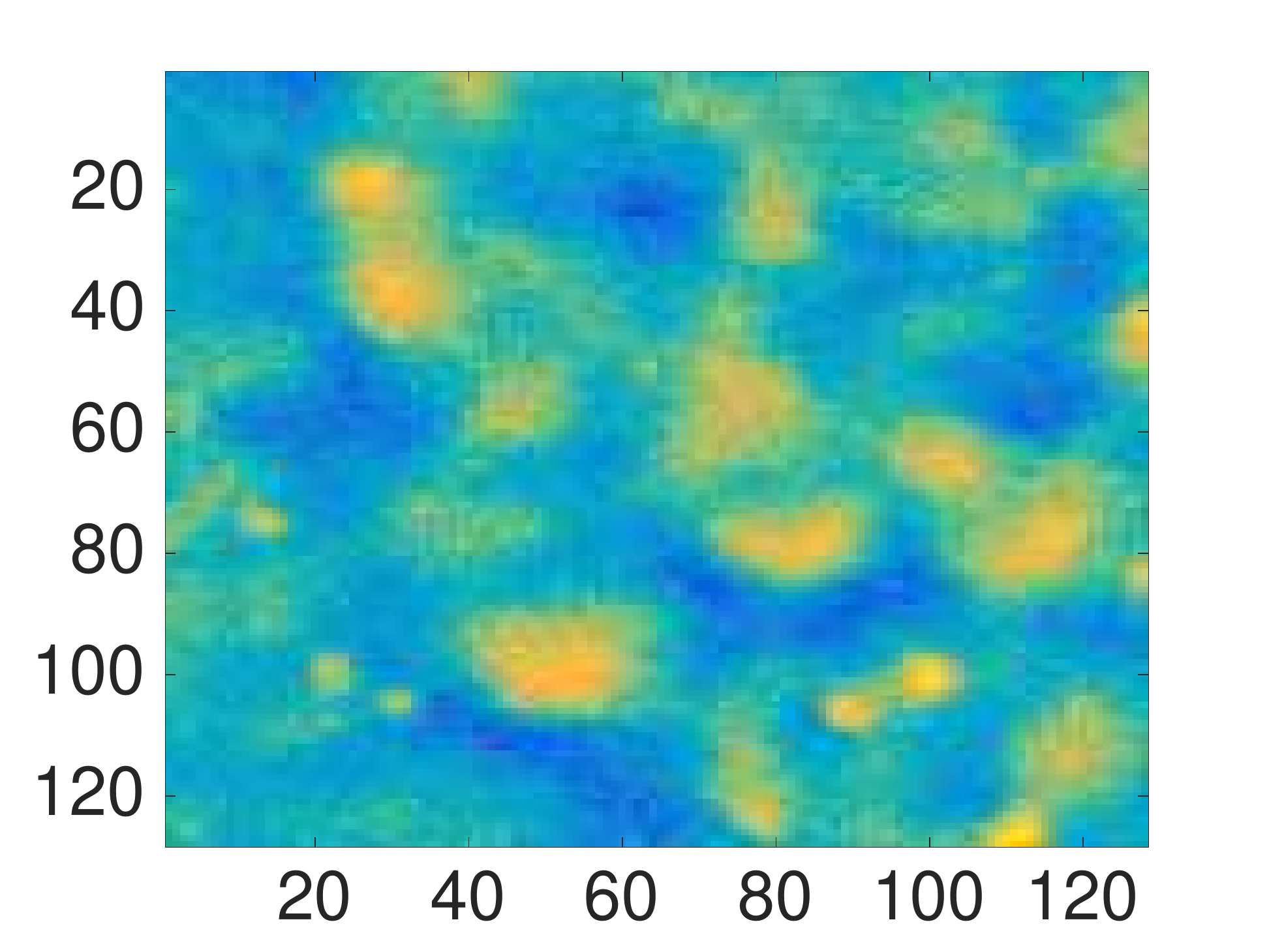}}
	\subfigure[]{\label{patch:snr10}
	\includegraphics[width=0.31\linewidth]{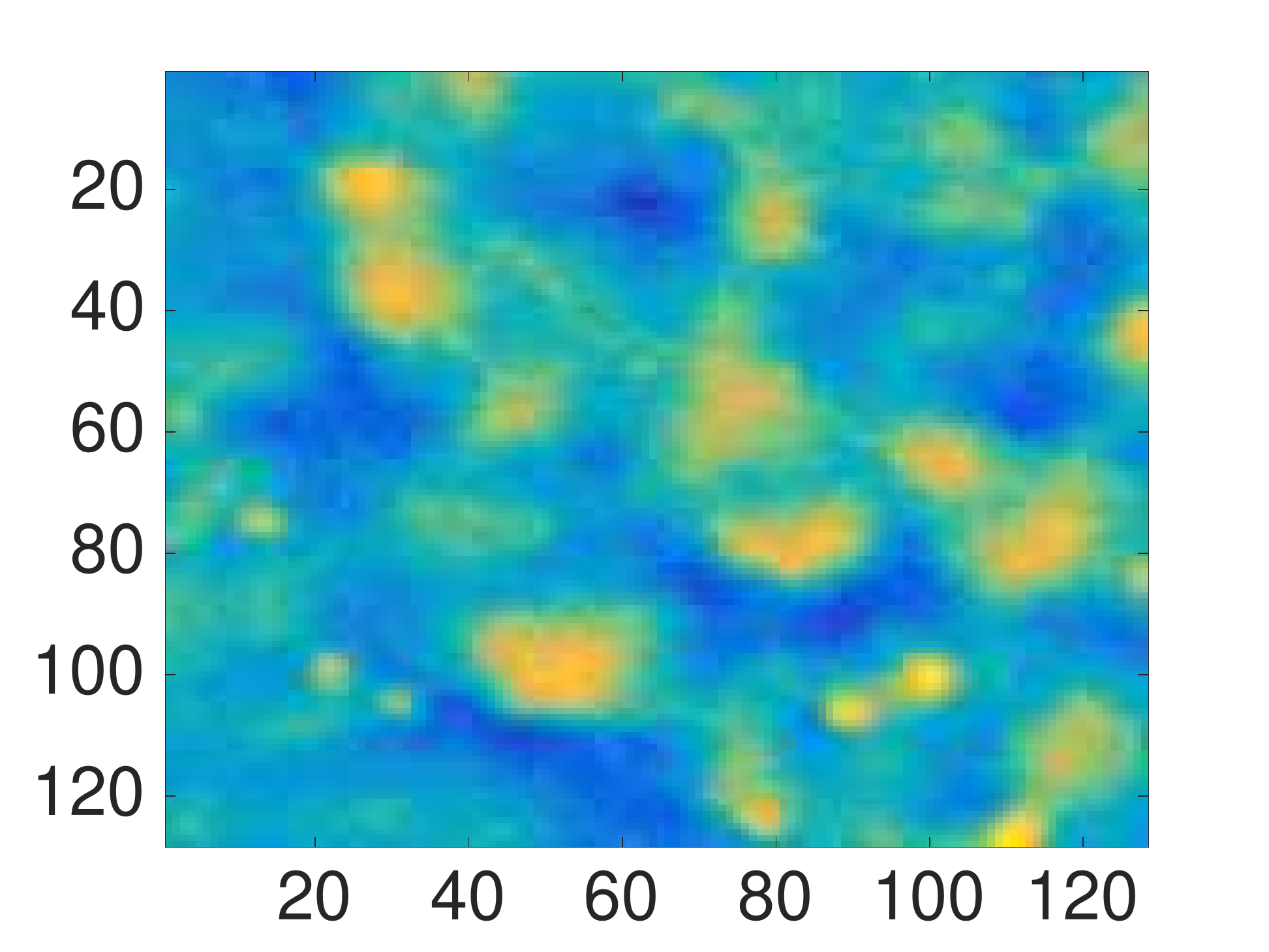}}
	\subfigure[]{\label{patch:snr15}
	\includegraphics[width=0.31\linewidth]{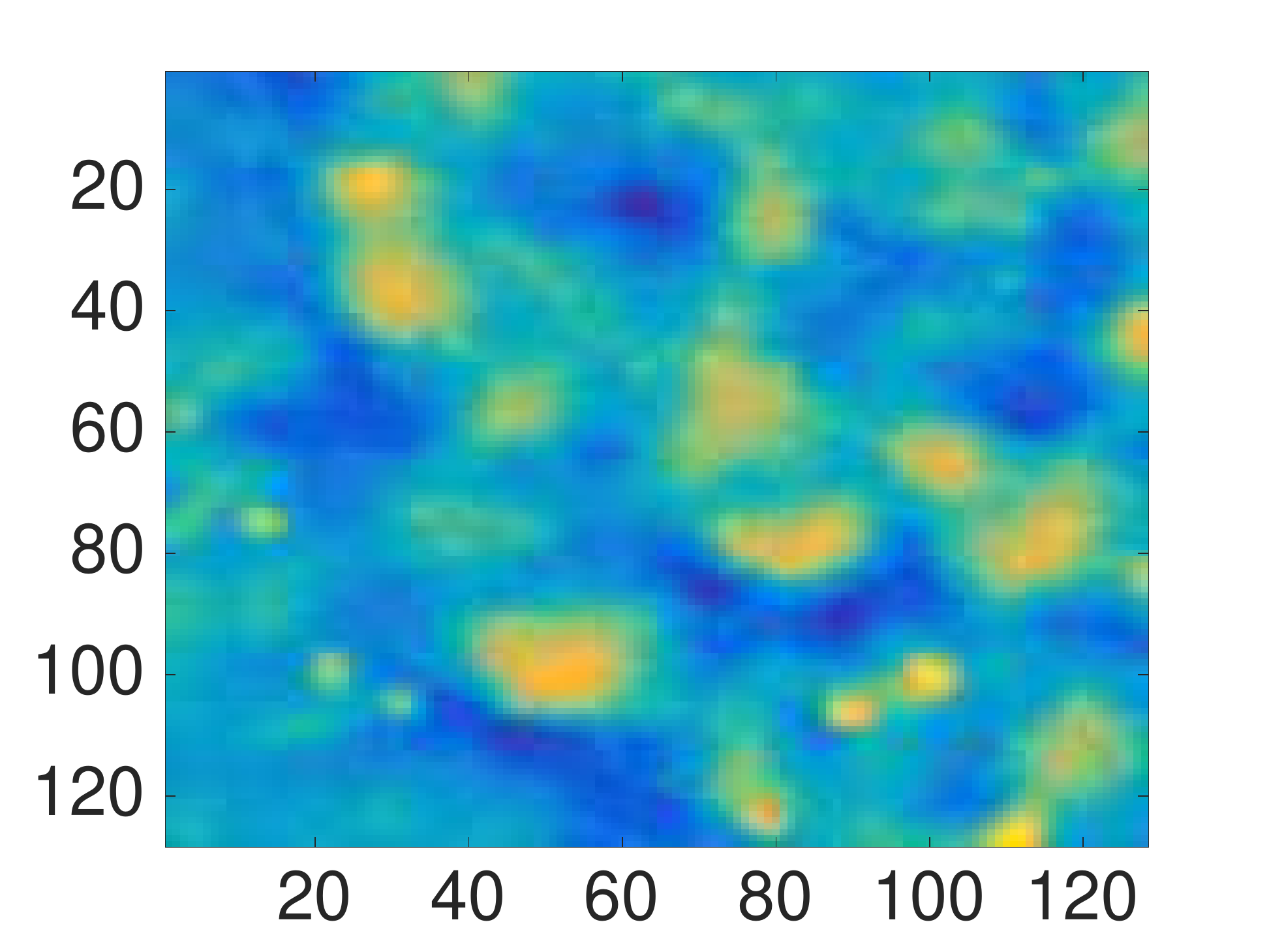}}\\
	\caption{The patch from Fig.~\ref{fig:patch}a with random Gaussian noise. SNR equals: (a) 5; (b) 10; (c) and 15.}
	\label{fig:noiseExample}
\end{figure}

\begin{figure}%
\centering
	\subfigure[]{\label{patch:noise}
	\includegraphics[width=0.47\linewidth]{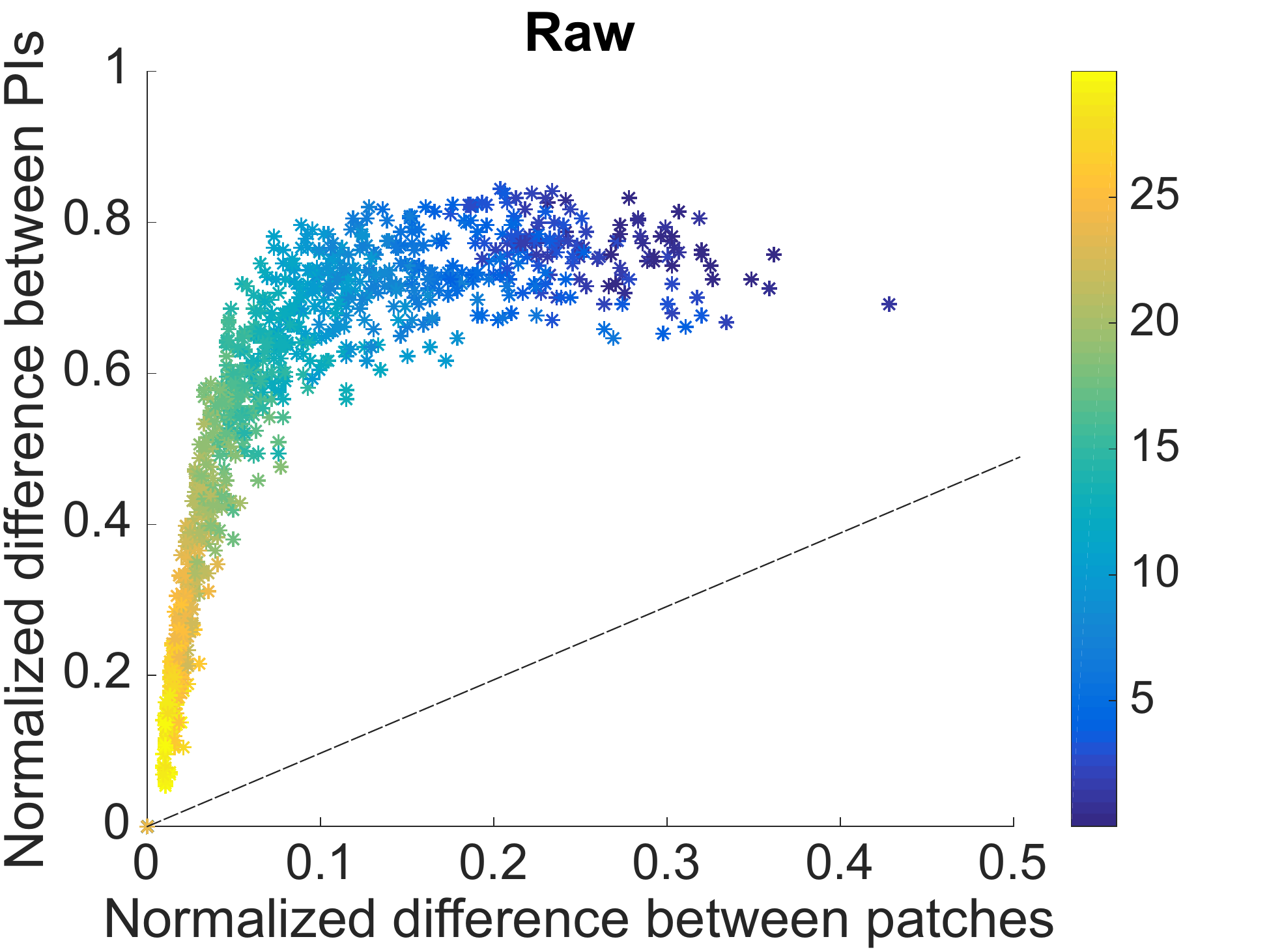}}
	\subfigure[]{\label{patch:noiseSchmid}
	\includegraphics[width=0.47\linewidth]{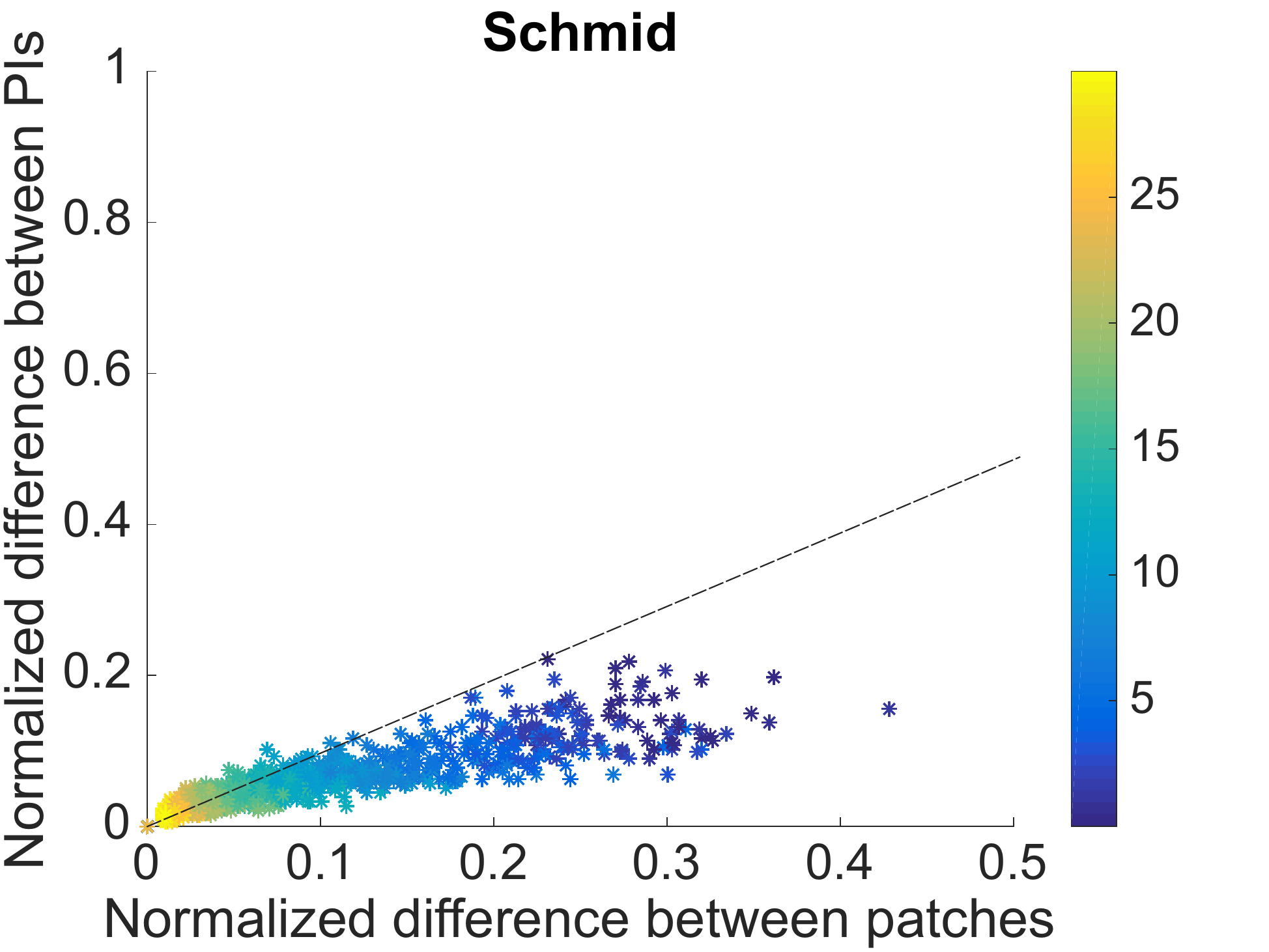}}
	\subfigure[]{\label{patch:noiseRSF1}
	\includegraphics[width=0.47\linewidth]{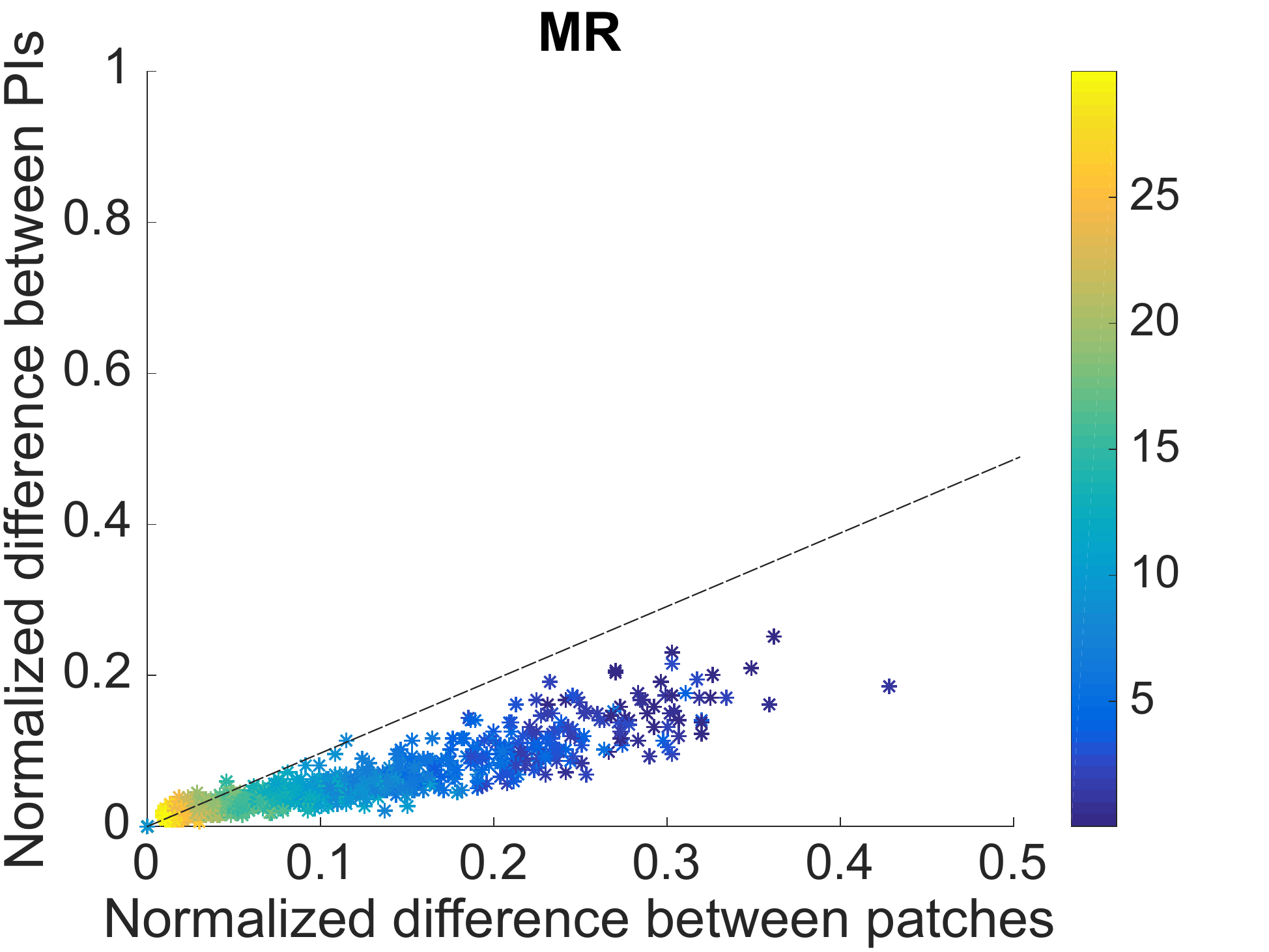}}
	\subfigure[]{\label{patch:noiseCLBPS}
	\includegraphics[width=0.47\linewidth]{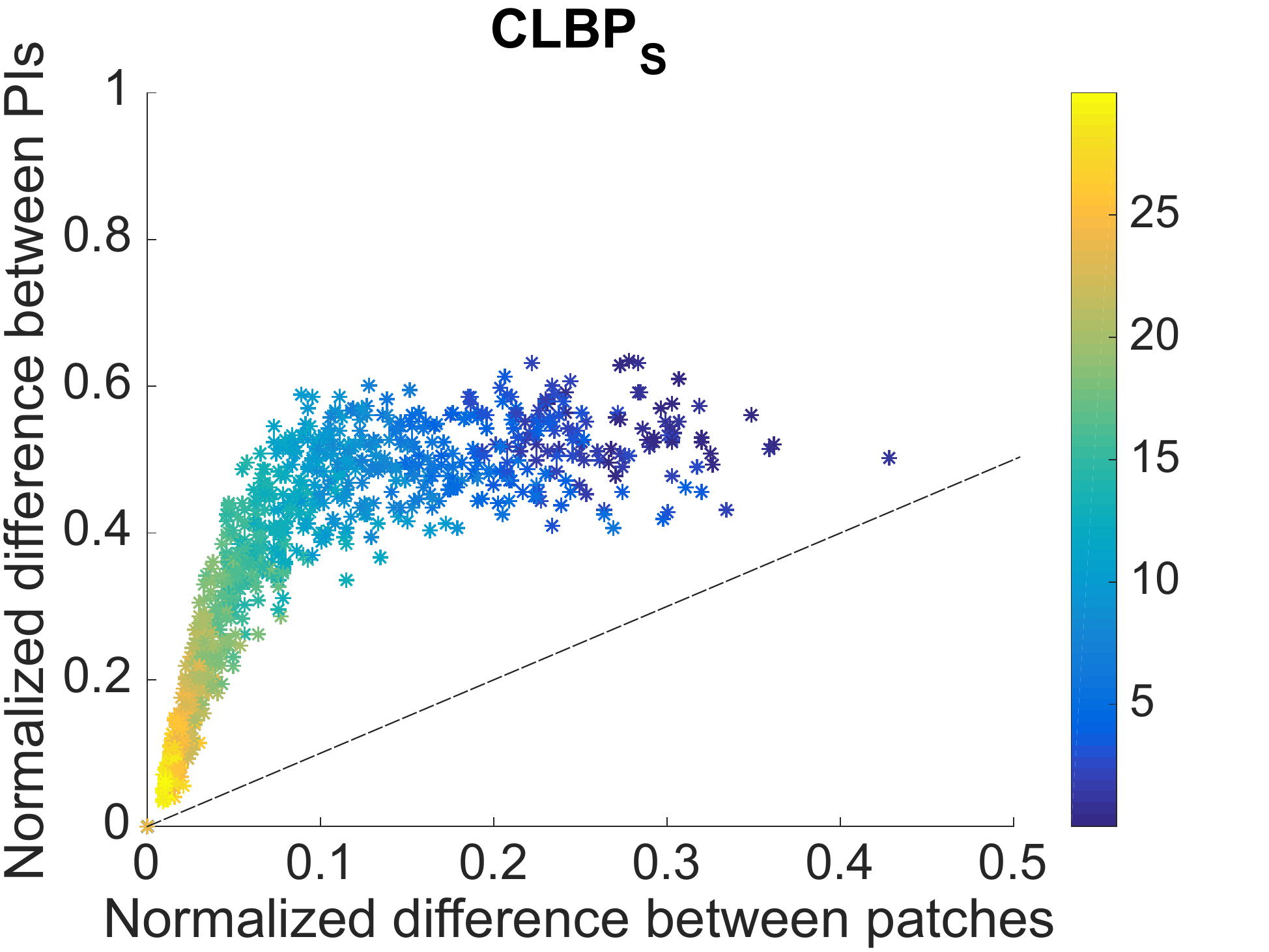}}
	\caption{The noise analysis: abscissa represents difference between the original patch and the same patch with random Gaussian noise; ordinate represents normalized difference between their PIs; color corresponds to value of SNR. The dashed line is the diagonal ($f(x) = x$) of the plot. Plots correspond to: (a) original patches; as well as patches filtered with (b) Schmid filter; (c) MR filter; (d) and CLBP.}
	\label{fig:noise}
\end{figure}

A crucial property of persistent homology is stability. In the following experiments we evaluate if the theoretical conclusions about stability from Section~\ref{sec:stability} hold in practice. For this purpose, we randomly select number of patches and add different levels of random Gaussian noise (see Fig.~\ref{fig:noiseExample}a-c for an example). Next, we compute the differences between the raw patches and their noisy variants as well as the differences between the PIs computed from the raw and noisy patches. The normalized difference $d$ between two patches is computed as: 
$$d(P,N)=\dfrac{\sum_{p \in P}{|P(p) - N(p)|}}{\sum_{p \in P}{|P(p)|+|N(p)|}},$$
where $N$ and $P$ is the original patch with and without random Gaussian noise. The normalized difference between two PIs is computed analogously to that of patches by replacing $N$ and $P$ with the corresponding PIs. 
In Fig.~\ref{fig:noise} we plot the differences between the patches (x-axis) vs. the differences between the PIs (y-axis). Color encodes the signal-to-noise ratio (SNR).

With decreasing SNR (increasing noise level) the differences between patches and PIs increases, see Fig.~\ref{fig:noise}a. Interestingly, the differences between the PIs exceeds the differences between the raw patches (i.e. points are above the diagonal) which shows that there is a strong sensitivity to the introduced noise. 

Different types of pre-filtering (described in Section~\ref{sec:preprocessing}) can also influence the stability of the resulting representations. To evaluate this further, we perform a similar analysis for patches filtered with Schmid filter, MR filter and CLBP (see Fig.~\ref{fig:noise}b-d). Experiments reveal that the difference between PIs grows much slower in case of Schmid filter and MR filter (nearly all points are below the diagonal). This is due to the fact, that both filters are defined based on Gaussian functions and have a low-pass characteristic. Thus, noise is removed to a wide extent in both cases. For CLBP the behavior is similar to the raw patches. However, the absolute value of difference $d$ is much lower ($d<0.6$ instead of $d<0.9$) than for the raw patches which shows that CLBP contributes to the robustness of the resulting PI.

\begin{figure}%
\centering
	\subfigure[]{\label{patch:overlap}
	\includegraphics[width=0.47\linewidth]{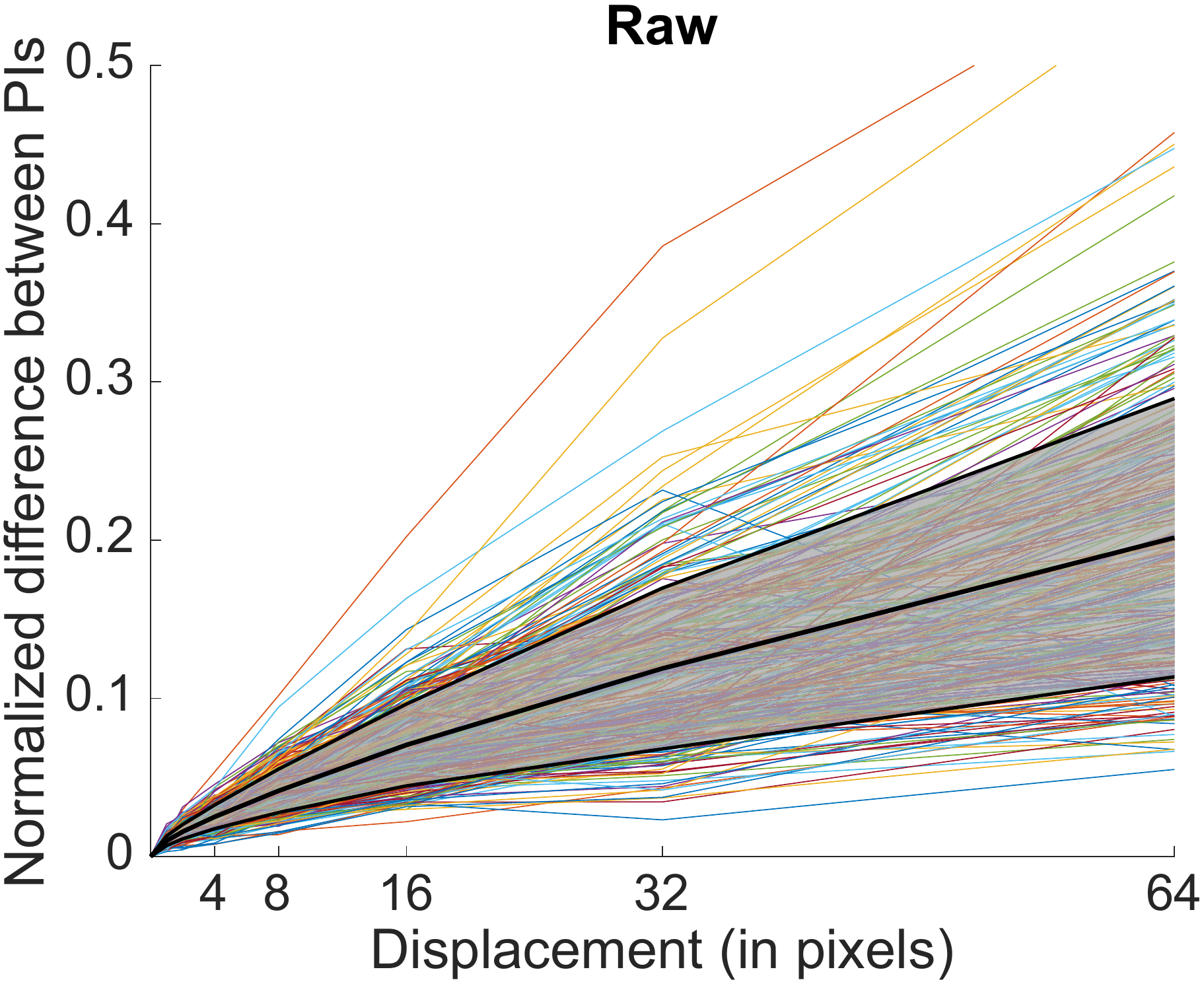}}
	\subfigure[]{\label{patch:overlapSchmid}
	\includegraphics[width=0.47\linewidth]{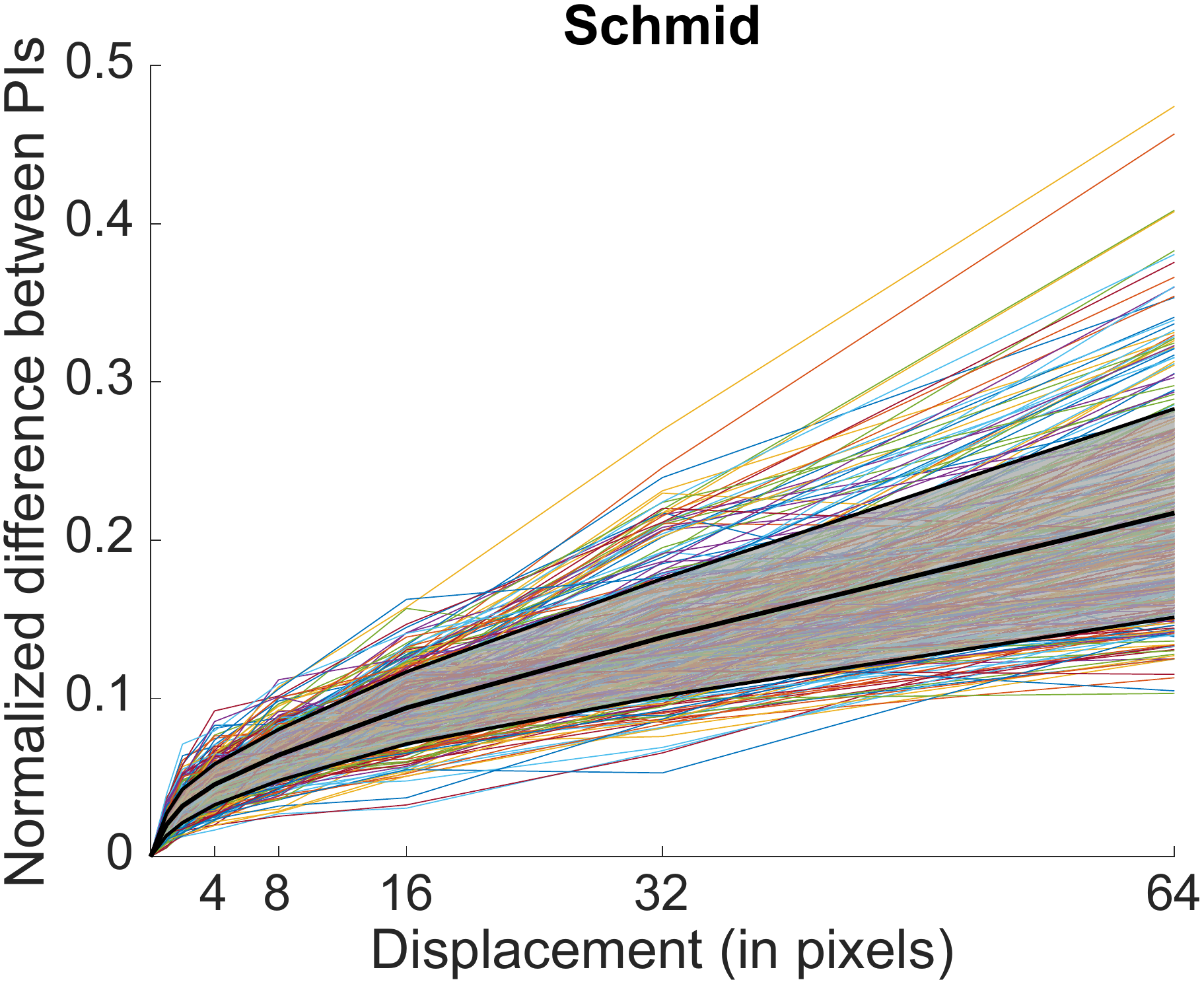}}
	\subfigure[]{\label{patch:overlapRSF1}
	\includegraphics[width=0.47\linewidth]{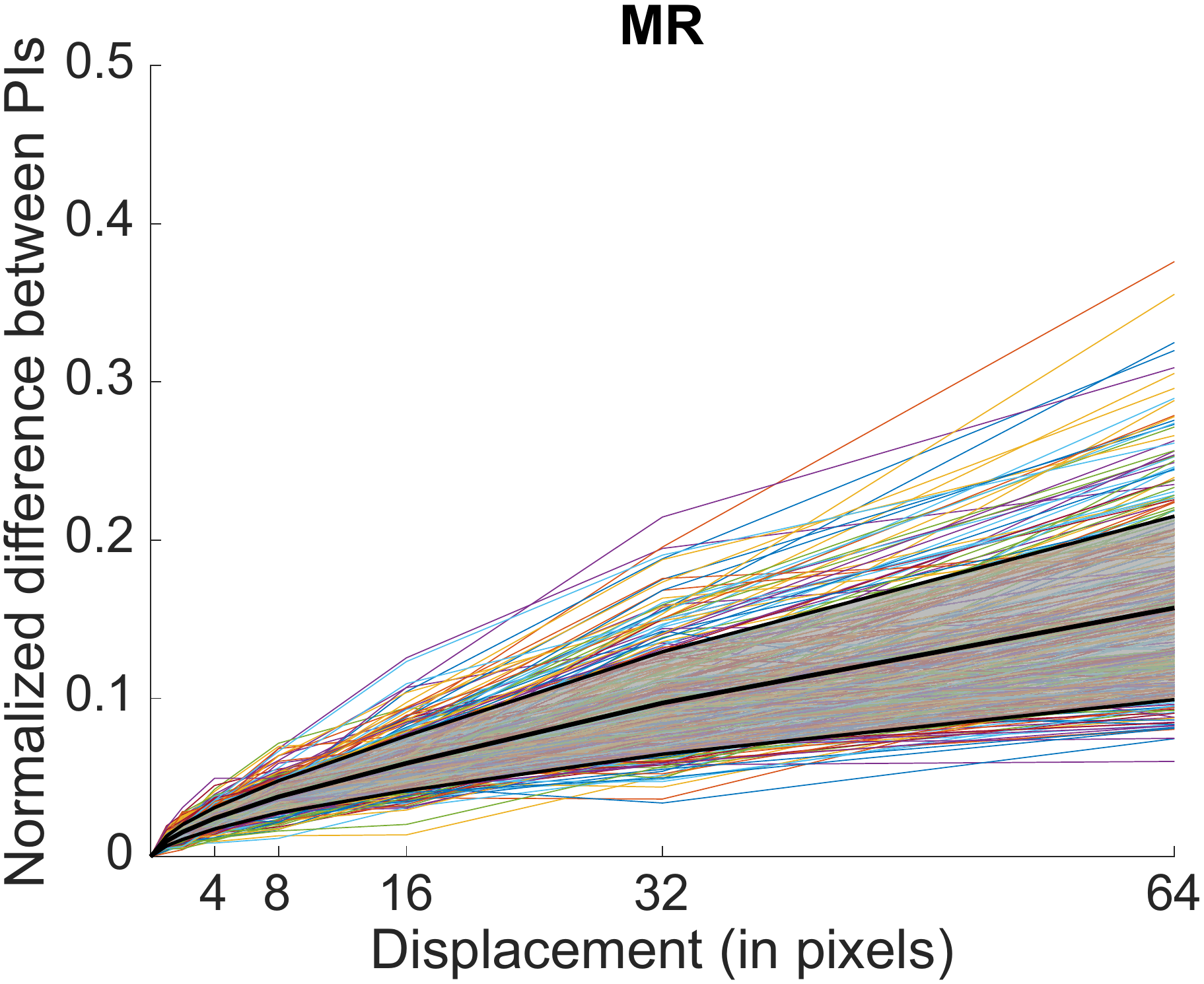}}
	\subfigure[]{\label{patch:overlapCLBPS}
	\includegraphics[width=0.47\linewidth]{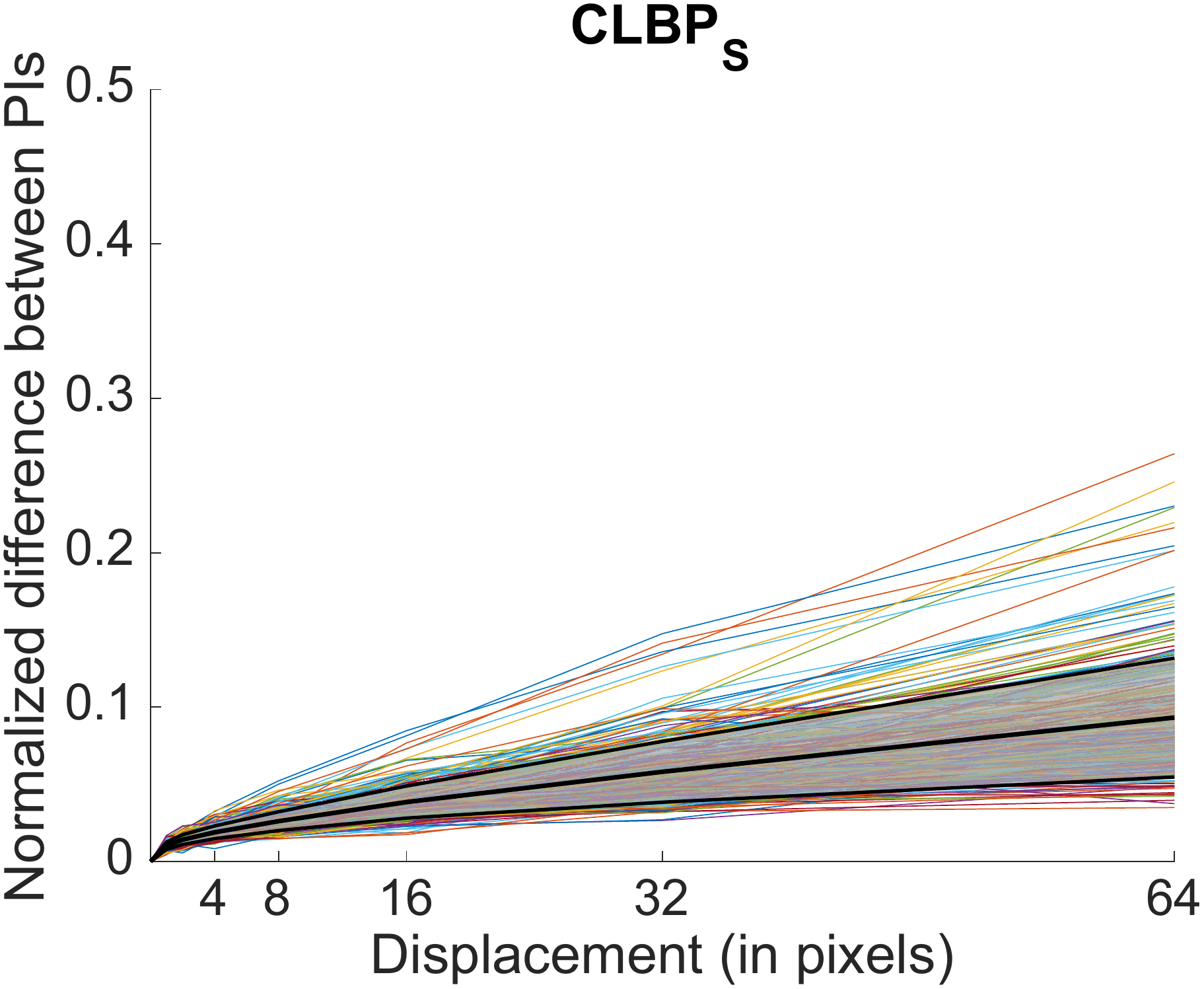}}
	\caption{Displacement analysis: abscissa represents displacement between the original patch and its neighbor; ordinate represents normalized difference between their PIs. Each curve corresponds to a single patch. Middle black curve represents mean value, while the upper and lower black lines presents the standard deviation, i.e. the grayish area represents the interval of mean $\pm$ standard deviation. Plots correspond to: (a) original patches; as well as patches filtered with (b) Schmid filter; (c) MR filter; (d) and CLBP.}
	\label{fig:overlap}
\end{figure}

Furthermore, we investigate how sensitive PIs are towards displacement of the underlying patches. For this purpose we compute the differences in the PIs stemming from a pair of neighboring and overlapping patches. Again, we randomly select number of patches. Next we shift the patches by 4, 8, 16, 32, and 64~pixels. From the resulting 5 pairs of patches we compute PIs and compute their normalized difference as above. In Fig.~\ref{fig:overlap}a-d we plot the differences in PIs for all displacements as separate lines for each patch. The trend of the curves shows that the difference between the PIs grows with the amount of displacement in all four cases. Interestingly, the differences are noticeably smaller in the case of CLBP, which indicates that it introduces an additional robustness to displacements to the PI which is not the case for Schmid and MR filters.

We observe, that our experiments confirm theoretical results from Section~\ref{sec:stability}. Furthermore, the computational results show that the constants in the stability theorems have only little influence on the results.

\subsection{Discriminativity}
\label{subsec:discriminativity}

\begin{figure}%
\centering
	\subfigure[]{\label{patch:meanFisherImpLog}
	\includegraphics[width=0.47\linewidth]{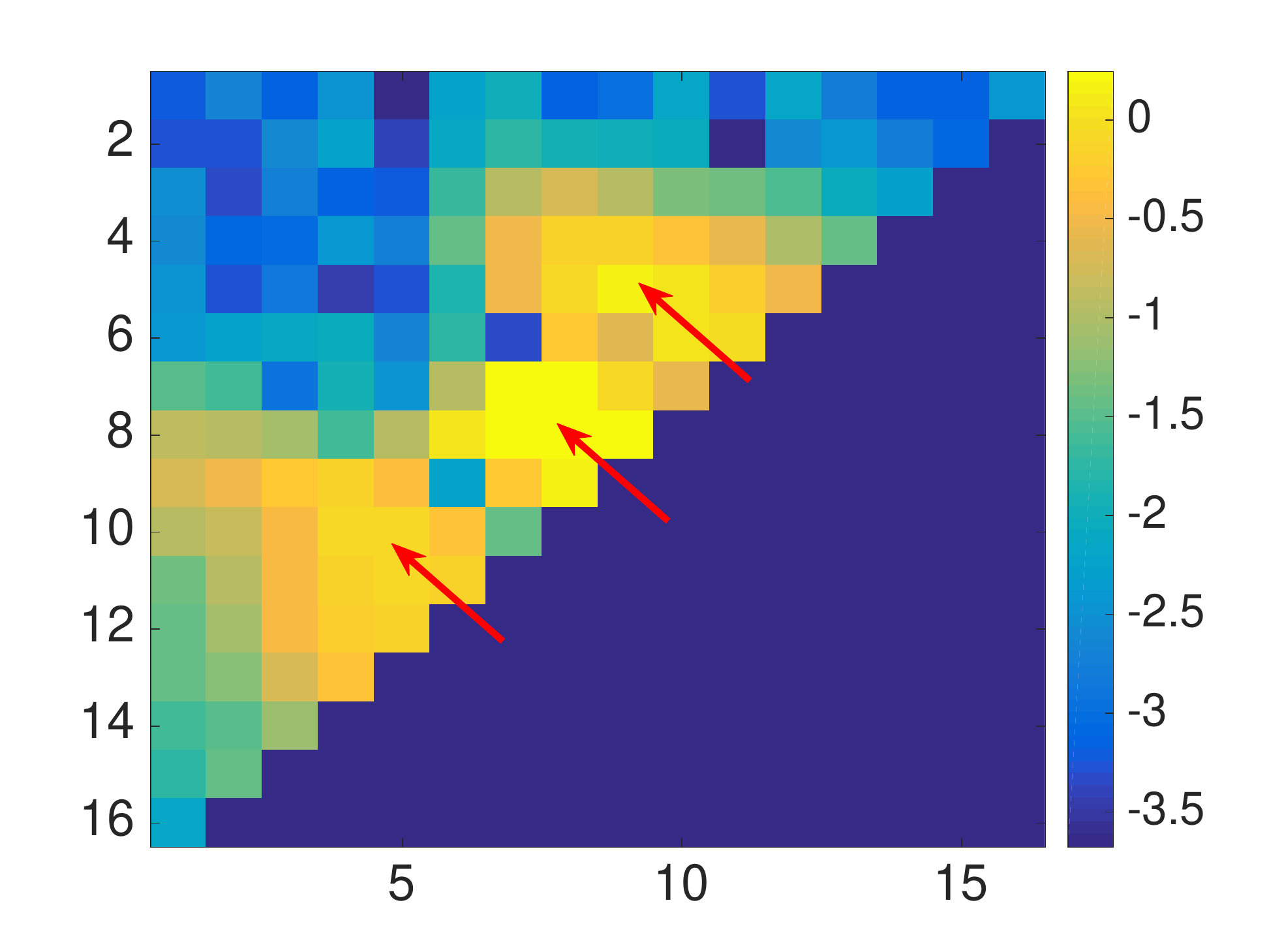}}
	\subfigure[]{\label{patch:meanGiniImpLog}
	\includegraphics[width=0.47\linewidth]{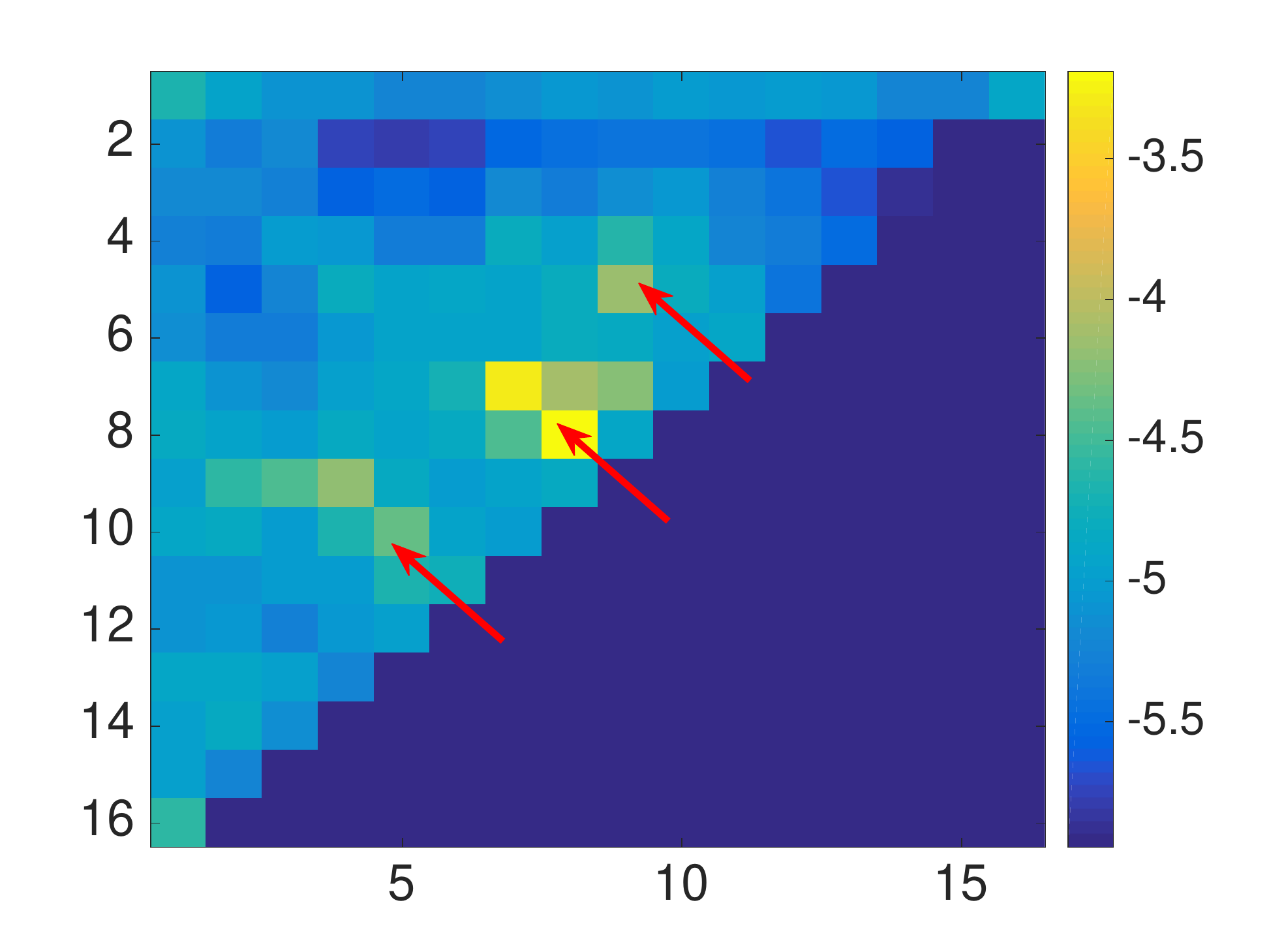}}
	\caption{Importance of PI pixels in the classification process, measured with: (a) Fisher discriminant; (b) and Gini importance. The scale in both cases is logarithmic (base 10). Brighter pixels are more important (discriminative) in classification. Red arrows correspond to three groups of important pixels.}
	\label{fig:importance}
\end{figure}

An important question in the context of PI is if all entries in the PI (pixels) are of equal importance or if some pixels are more important than others. Experiments of feature selection in Section~\ref{subsec:featureSel} have indicated that a small set of pixels exist that are more important than the other pixels. To determine the importance of the individual pixels we compute for each pixel (a) the Fisher discriminant and (b) the Gini importance for the entire training set. Fig~\ref{fig:importance} shows the resulting images where each pixel in the PI is colored according to its importance.

It turns out that the most important pixels are those located in the center of the PI quite near to the diagonal. According to Fisher criterion there are also two smaller groups of important pixels, above and below the center. This is less visible but also present in case of Gini, see red arrows in Fig.~\ref{fig:importance}b. 

\begin{figure}%
\centering
	\subfigure[]{\label{patch:fgAvgPI}
	\includegraphics[width=0.47\linewidth]{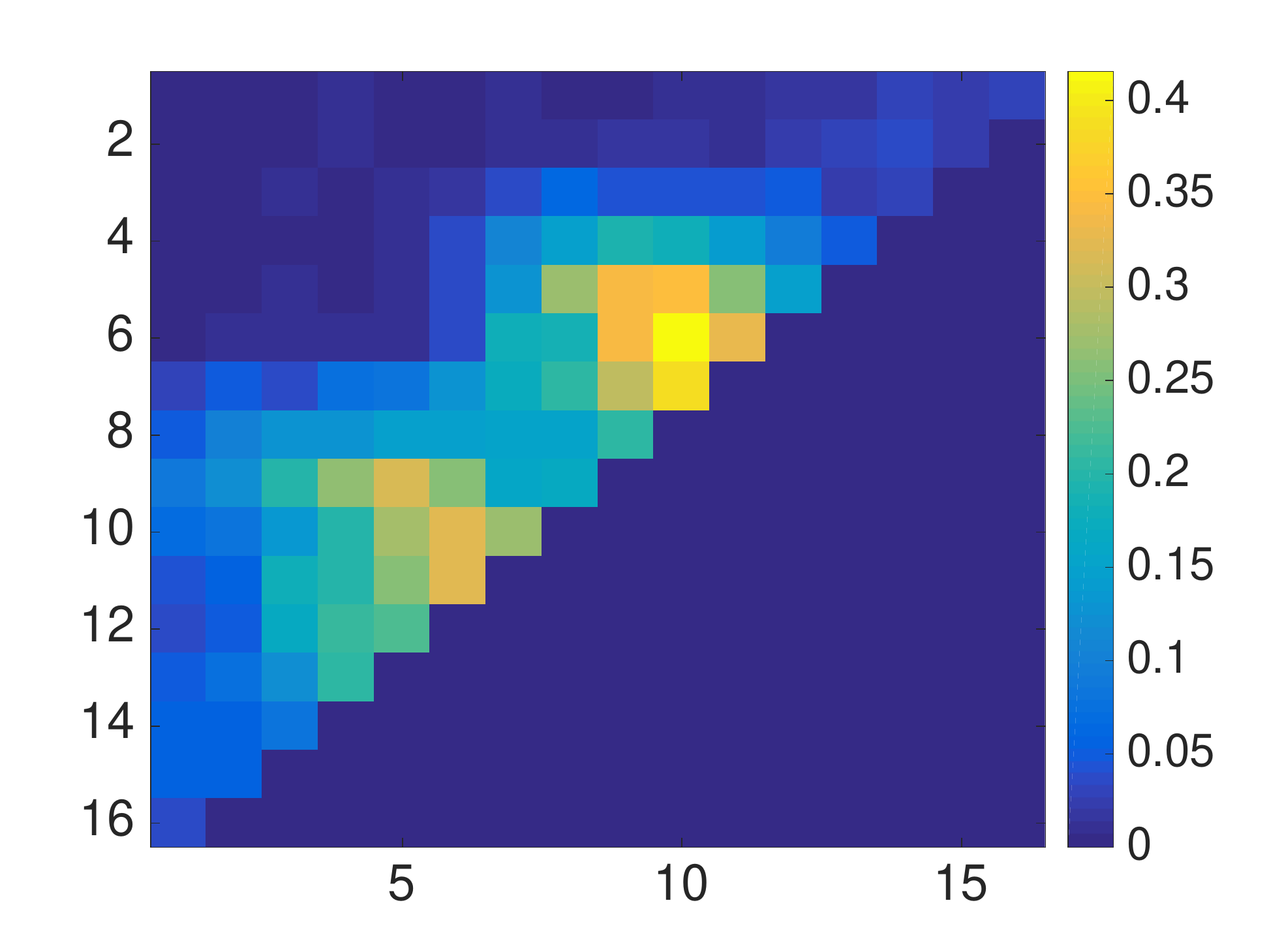}}
	\subfigure[]{\label{patch:bgAvgPI}
	\includegraphics[width=0.47\linewidth]{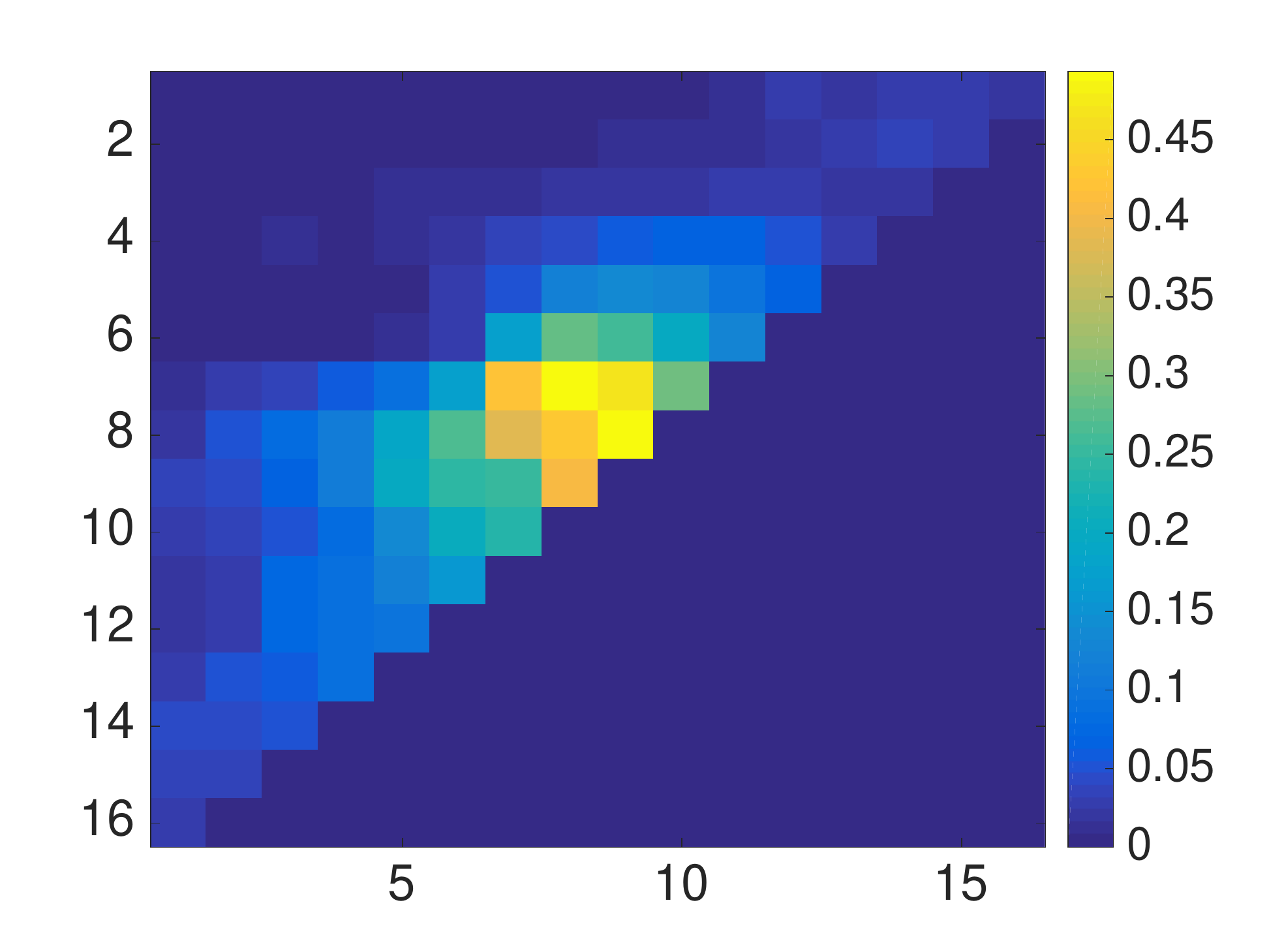}}
	\caption{Averaged PIs of the training patches from (a) class 1 and (b) class 2.}
	\label{fig:avgPI}
\end{figure}

In order to further understand those results, we generated average PIs for the investigated classes of surface texture separately, see Fig.~\ref{fig:avgPI}. The averaged PIs clearly show that the two classes exhibit highly different spatial patterns in the PIs which strongly correlate with the three groups of pixels identified with Fisher discriminant and Gini importance. This shows that PI captures the different classes of surface textures well.

\review{The results are further consistent with those obtained for the synthetic dataset (see Section \ref{sec:synthetic}). Class 1 representing engraved areas has bimodal distribution, while class 2 representing the natural rock surface is unimodal.}

\review{The analysis on discriminativity further provides insights into the question of weighting in PI computation (see Section \ref{subsec:weighting}). In general, PI requires a proper weighting function for stability reasons (see \cite{adams2015persistent}). Our analysis on discriminativity shows, however, that the components near the diagonal are most important and thus most probably do not represent noise. The reason for this is that the 3D surfaces have been smoothed and filtered for outliers prior to our processing. Thus noise (which usually appears along the diagonal of PD) has already been removed and all the remaining points actually represent important topological information. In our case, points near to the diagonal represent small surface deviations that correspond to the fine-structure of the surface and that are of high importance for surface texture classification (see Fig. \ref{fig:importance}).}  
\review{Thus, using weighting would be counterproductive in our case. In other situations, it is possible to avoid instability issues with PI by using another approach for constructing PI, such as the one described by Eqn. (9) in \cite{reininghaus2014stable}, which reduces the influence of the low-persistence points close to the diagonal when the scale $\sigma$ increases.}

\section{Conclusions}
\label{sec:concl}

\review{In this article we have presented an extensive study on topological descriptors for 3D surface texture analysis. From patches of the surfaces' depth maps we extract different topological descriptors which build upon the persistence diagram. The descriptors have fixed dimensions and can thus easily be combined with conventional classifiers as well as with traditional non-topological descriptors. We perform experiments on surface texture classification and investigate different aspects of the topological descriptors, such as sensitivity to parameters, robustness to noise, their discriminative power and the information density. Moreover, we propose and evaluate different normalization strategies, pre-filtering, and alternative weighting functions and investigate the complementary nature of topological and non-topological descriptors. The major conclusions drawn from our study are:}

\begin{itemize}

\item \review{PD\_AGG and PI capture relevant and discriminative information on surface textures, whereby our study clearly shows that the spatial information from the PD preserved in PI leads to a more powerful descriptor.}
\item \review{PIs exhibit a low sensitivity to changes in resolution and sigma whereas a certain minimum resolution must be ensured to obtain an expressive descriptor. Sigma shows to be independent from resolution to a large extent.
}
\item \review{For 3D surfaces with varying depth ranges, depth normalization is essential to obtain comparable topological descriptors. In our experiments global normalization leads to the best results.}
\item \review{While related works propose LBP-based representations our study shows that such a pre-filtering does not necessarily improve results in practice.} 
\item \review{Contrary to our expectations, it turned out that the most important information in PI is located near to the diagonal of the PD. A deeper analysis reveals that the most expressive information is spatially clustered in the PI and that these clusters are characteristic for the investigated classes.}
\end{itemize}

\review{
A number of open questions and future research directions were identified in our study. Major questions include: (i) how can weighting in PI construction be improved in a way that characteristics of the underlying data are taken into account? (ii) how can the resolution of PI be improved in areas where the most important information is concentrated to increase the expressiveness of PI? (iii) do more powerful pre-filter functions that build upon topological attributes such as~\cite{Nguyen2015} improve the descriptors and can we learn them from the data?}

Overall our investigation shows that topological descriptors can compete with non-topological descriptors for surface texture classification. A comparison with \cite{poier2016petrosurf3d} shows that our approach can compete with (and even slightly outperform) the performance of a CNN on the same dataset. Topological and non-topological descriptors achieve a significant improvement of performance when they are combined with each other. This confirms that topological descriptors capture relevant information that is not captured by traditional descriptors. Thus, they are a promising approach for further research.

\section{Acknowledgments}
Parts of the work for this article have been carried out in the project 3D-Pitoti which is funded from the European Community's Seventh Framework Programme (FP7/2007-2013) under grant agreement no 600545; 2013-2016. Other parts of the work from this article were supported by the National Science Centre, Poland under grant agreement no 2015/19/D/ST6/01215; 2016-2019.

\section*{References}
\bibliographystyle{model1-num-names}
\bibliography{mbmm2}

\begin{thebibliography}{66}
\expandafter\ifx\csname natexlab\endcsname\relax\def\natexlab#1{#1}\fi
\providecommand{\bibinfo}[2]{#2}
\ifx\xfnm\relax \def\xfnm[#1]{\unskip,\space#1}\fi
\bibitem[{Crandall et~al.(2011)Crandall, Owens, Snavely, and
  Huttenlocher}]{crandall2011}
\bibinfo{author}{D.~Crandall}, \bibinfo{author}{A.~Owens},
  \bibinfo{author}{N.~Snavely}, \bibinfo{author}{D.~Huttenlocher},
\newblock \bibinfo{title}{Discrete-continuous optimization for large-scale
  structure from motion},
\newblock in: \bibinfo{booktitle}{Computer Vision and Pattern Recognition
  (CVPR), 2011 IEEE Conference on}, \bibinfo{organization}{IEEE}, pp.
  \bibinfo{pages}{3001--3008}.
\bibitem[{Wu(2013)}]{wu2013}
\bibinfo{author}{C.~Wu},
\newblock \bibinfo{title}{Towards linear-time incremental structure from
  motion},
\newblock in: \bibinfo{booktitle}{3DTV-Conference, 2013 International
  Conference on}, \bibinfo{organization}{IEEE}, pp. \bibinfo{pages}{127--134}.
\bibitem[{Wohlfeil et~al.(2013)Wohlfeil, Strackenbrock, and
  Kossyk}]{wohlfeil2013}
\bibinfo{author}{J.~Wohlfeil}, \bibinfo{author}{B.~Strackenbrock},
  \bibinfo{author}{I.~Kossyk},
\newblock \bibinfo{title}{Automated high resolution 3d reconstruction of
  cultural heritage using multi-scale sensor systems and semi-global matching},
\newblock \bibinfo{journal}{International Archives of the Photogrammetry,
  Remote Sensing and Spatial Information Sciences, XL-4 W} \bibinfo{volume}{4}
  (\bibinfo{year}{2013}) \bibinfo{pages}{37--43}.
\bibitem[{Tuceryan and Jain(1998)}]{tuceryan1998}
\bibinfo{author}{M.~Tuceryan}, \bibinfo{author}{A.~Jain},
\newblock \bibinfo{title}{Texture analysis},
\newblock \bibinfo{journal}{The handbook of pattern recognition and computer
  vision} \bibinfo{volume}{2} (\bibinfo{year}{1998}) \bibinfo{pages}{207--248}.
\bibitem[{ASME(1996)}]{ansiSurfaceTexture1996}
\bibinfo{author}{ASME}, \bibinfo{title}{ASME B46.1-1995 Surface texture
  (surface roughness, waviness and lay): An American National Standard},
  \bibinfo{publisher}{The Anmerican Society of Mechanical Engineers (ASME), New
  York, USA}, \bibinfo{year}{1996}.
\bibitem[{Blunt and Jiang(2003)}]{blunt2003}
\bibinfo{author}{L.~Blunt}, \bibinfo{author}{X.~Jiang},
  \bibinfo{title}{Advanced techniques for assessment surface topography:
  development of a basis for 3D surface texture standards ``surfstand"},
  \bibinfo{publisher}{Elsevier}, \bibinfo{year}{2003}.
\bibitem[{Barcelo and Moitinho~de Almeida(2012)}]{barcelo2012}
\bibinfo{author}{J.~Barcelo}, \bibinfo{author}{V.~Moitinho~de Almeida},
\newblock \bibinfo{title}{Functional analysis from visual and non-visual data.
  an artificial intelligence approach},
\newblock \bibinfo{journal}{Mediterranean Archaeology and Archaeometry}
  \bibinfo{volume}{12} (\bibinfo{year}{2012}) \bibinfo{pages}{273--321}.
\bibitem[{Zeppelzauer and Seidl(2015)}]{zeppelzauer2015efficient}
\bibinfo{author}{M.~Zeppelzauer}, \bibinfo{author}{M.~Seidl},
\newblock \bibinfo{title}{Efficient image-space extraction and representation
  of 3d surface topography},
\newblock in: \bibinfo{booktitle}{Proceedings of the IEEE International
  Conference on Image Processing (ICIP)}, \bibinfo{publisher}{IEEE},
  \bibinfo{address}{Quebec, Canada}, \bibinfo{year}{2015}.
\bibitem[{Zeppelzauer et~al.(2016)Zeppelzauer, Zieli{\'n}ski, Juda, and
  Seidl}]{zeppelzauer2016topological}
\bibinfo{author}{M.~Zeppelzauer}, \bibinfo{author}{B.~Zieli{\'n}ski},
  \bibinfo{author}{M.~Juda}, \bibinfo{author}{M.~Seidl},
\newblock \bibinfo{title}{Topological descriptors for 3d surface analysis},
\newblock in: \bibinfo{booktitle}{International Workshop on Computational
  Topology in Image Context}, \bibinfo{publisher}{Springer},
  \bibinfo{year}{2016}, pp. \bibinfo{pages}{77--87}.
\bibitem[{Fasy et~al.(2014)Fasy, Lecci, Rinaldo, Wasserman, Balakrishnan, Singh
  et~al.}]{fasy2014confidence}
\bibinfo{author}{B.~T. Fasy}, \bibinfo{author}{F.~Lecci},
  \bibinfo{author}{A.~Rinaldo}, \bibinfo{author}{L.~Wasserman},
  \bibinfo{author}{S.~Balakrishnan}, \bibinfo{author}{A.~Singh}, et~al.,
\newblock \bibinfo{title}{Confidence sets for persistence diagrams},
\newblock \bibinfo{journal}{The Annals of Statistics} \bibinfo{volume}{42}
  (\bibinfo{year}{2014}) \bibinfo{pages}{2301--2339}.
\bibitem[{Poier et~al.(2017)Poier, Seidl, Zeppelzauer, Reinbacher, Schaich,
  Bellando, Marretta, and Bischof}]{poier2016petrosurf3d}
\bibinfo{author}{G.~Poier}, \bibinfo{author}{M.~Seidl},
  \bibinfo{author}{M.~Zeppelzauer}, \bibinfo{author}{C.~Reinbacher},
  \bibinfo{author}{M.~Schaich}, \bibinfo{author}{G.~Bellando},
  \bibinfo{author}{A.~Marretta}, \bibinfo{author}{H.~Bischof},
\newblock \bibinfo{title}{The 3d-pitoti dataset: A dataset for high-resolution
  3d surface segmentation},
\newblock in: \bibinfo{booktitle}{Proceedings of the 15th International
  Workshop on Content-Based Multimedia Indexing}, \bibinfo{publisher}{ACM},
  \bibinfo{address}{Firenze, Italy}, \bibinfo{year}{2017}.
\bibitem[{Johnson and Hebert(1999)}]{johnson1999using}
\bibinfo{author}{A.~E. Johnson}, \bibinfo{author}{M.~Hebert},
\newblock \bibinfo{title}{Using spin images for efficient object recognition in
  cluttered 3d scenes},
\newblock \bibinfo{journal}{Pattern Analysis and Machine Intelligence, IEEE
  Transactions on} \bibinfo{volume}{21} (\bibinfo{year}{1999})
  \bibinfo{pages}{433--449}.
\bibitem[{Darom and Keller(2012)}]{darom2012scale}
\bibinfo{author}{T.~Darom}, \bibinfo{author}{Y.~Keller},
\newblock \bibinfo{title}{Scale-invariant features for 3-d mesh models},
\newblock \bibinfo{journal}{Image Processing, IEEE Transactions on}
  \bibinfo{volume}{21} (\bibinfo{year}{2012}) \bibinfo{pages}{2758--2769}.
\bibitem[{Zaharescu et~al.(2009)Zaharescu, Boyer, Varanasi, and
  Horaud}]{zaharescu2009surface}
\bibinfo{author}{A.~Zaharescu}, \bibinfo{author}{E.~Boyer},
  \bibinfo{author}{K.~Varanasi}, \bibinfo{author}{R.~Horaud},
\newblock \bibinfo{title}{Surface feature detection and description with
  applications to mesh matching},
\newblock in: \bibinfo{booktitle}{Computer Vision and Pattern Recognition,
  2009. CVPR 2009. IEEE Conference on}, \bibinfo{organization}{IEEE}, pp.
  \bibinfo{pages}{373--380}.
\bibitem[{Steder et~al.(2011)Steder, Rusu, Konolige, and
  Burgard}]{steder2011point}
\bibinfo{author}{B.~Steder}, \bibinfo{author}{R.~B. Rusu},
  \bibinfo{author}{K.~Konolige}, \bibinfo{author}{W.~Burgard},
\newblock \bibinfo{title}{Point feature extraction on 3d range scans taking
  into account object boundaries},
\newblock in: \bibinfo{booktitle}{IEEE International Conference on Robotics and
  Automation (ICRA)}, \bibinfo{publisher}{IEEE}, \bibinfo{year}{2011}, pp.
  \bibinfo{pages}{2601--2608}.
\bibitem[{Belongie et~al.(2002)Belongie, Malik, and
  Puzicha}]{belongie2002shape}
\bibinfo{author}{S.~Belongie}, \bibinfo{author}{J.~Malik},
  \bibinfo{author}{J.~Puzicha},
\newblock \bibinfo{title}{Shape matching and object recognition using shape
  contexts},
\newblock \bibinfo{journal}{Pattern Analysis and Machine Intelligence, IEEE
  Transactions on} \bibinfo{volume}{24} (\bibinfo{year}{2002})
  \bibinfo{pages}{509--522}.
\bibitem[{Frome et~al.(2004)Frome, Huber, Kolluri, B{\"u}low, and
  Malik}]{frome_recognizing_2004}
\bibinfo{author}{A.~Frome}, \bibinfo{author}{D.~Huber},
  \bibinfo{author}{R.~Kolluri}, \bibinfo{author}{T.~B{\"u}low},
  \bibinfo{author}{J.~Malik},
\newblock \bibinfo{title}{Recognizing objects in range data using regional
  point descriptors},
\newblock in: \bibinfo{booktitle}{Computer Vision-{ECCV} 2004},
  \bibinfo{publisher}{Springer}, \bibinfo{year}{2004}, pp.
  \bibinfo{pages}{224--237}. \bibinfo{note}{00401}.
\bibitem[{Rusu et~al.(2008)Rusu, Marton, Blodow, and
  Beetz}]{rusu_persistent_2008}
\bibinfo{author}{R.~B. Rusu}, \bibinfo{author}{Z.~C. Marton},
  \bibinfo{author}{N.~Blodow}, \bibinfo{author}{M.~Beetz},
\newblock \bibinfo{title}{Persistent point feature histograms for 3d point
  clouds},
\newblock in: \bibinfo{booktitle}{Proc 10th Int Conf Intel Autonomous Syst
  ({IAS}-10), Baden-Baden, Germany}, \bibinfo{publisher}{IOS Press},
  \bibinfo{year}{2008}, pp. \bibinfo{pages}{119--128}.
\bibitem[{Rusu et~al.(2009)Rusu, Blodow, and Beetz}]{rusu_fast_2009}
\bibinfo{author}{R.~B. Rusu}, \bibinfo{author}{N.~Blodow},
  \bibinfo{author}{M.~Beetz},
\newblock \bibinfo{title}{Fast point feature histograms ({FPFH}) for 3d
  registration},
\newblock in: \bibinfo{booktitle}{Robotics and Automation, 2009. {ICRA}'09.
  {IEEE} International Conference on}, \bibinfo{publisher}{{IEEE}},
  \bibinfo{year}{2009}, pp. \bibinfo{pages}{3212--3217}. \bibinfo{note}{00320}.
\bibitem[{Wahl et~al.(2003)Wahl, Hillenbrand, and
  Hirzinger}]{wahl_surflet-pair-relation_2003}
\bibinfo{author}{E.~Wahl}, \bibinfo{author}{U.~Hillenbrand},
  \bibinfo{author}{G.~Hirzinger},
\newblock \bibinfo{title}{Surflet-pair-relation histograms: a statistical
  3d-shape representation for rapid classification},
\newblock in: \bibinfo{booktitle}{3-D Digital Imaging and Modeling, 2003. 3DIM
  2003. Proceedings. Fourth International Conference on},
  \bibinfo{publisher}{{IEEE}}, \bibinfo{year}{2003}, pp.
  \bibinfo{pages}{474--481}.
\bibitem[{Tombari et~al.(2010)Tombari, Salti, and
  Di~Stefano}]{tombari_unique_2010}
\bibinfo{author}{F.~Tombari}, \bibinfo{author}{S.~Salti},
  \bibinfo{author}{L.~Di~Stefano},
\newblock \bibinfo{title}{Unique signatures of histograms for local surface
  description},
\newblock in: \bibinfo{booktitle}{Computer Vision{\textendash}{ECCV} 2010},
  \bibinfo{publisher}{Springer}, \bibinfo{year}{2010}, pp.
  \bibinfo{pages}{356--369}. \bibinfo{note}{00160}.
\bibitem[{Othmani et~al.(2013)Othmani, Lew Yan~Voon, Stolz, and
  Piboule}]{othmani2013single}
\bibinfo{author}{A.~Othmani}, \bibinfo{author}{L.~Lew Yan~Voon},
  \bibinfo{author}{C.~Stolz}, \bibinfo{author}{A.~Piboule},
\newblock \bibinfo{title}{Single tree species classification from terrestrial
  laser scanning data for forest inventory},
\newblock \bibinfo{journal}{Pattern Recognition Letters} \bibinfo{volume}{34}
  (\bibinfo{year}{2013}) \bibinfo{pages}{2144--2150}.
\bibitem[{Leung and Malik(1996)}]{leung1996detecting}
\bibinfo{author}{T.~Leung}, \bibinfo{author}{J.~Malik},
\newblock \bibinfo{title}{Detecting, localizing and grouping repeated scene
  elements from an image},
\newblock in: \bibinfo{booktitle}{European Conference on Computer Vision},
  \bibinfo{publisher}{Springer}, \bibinfo{year}{1996}, pp.
  \bibinfo{pages}{546--555}.
\bibitem[{Csurka et~al.(2004)Csurka, Dance, Fan, Willamowski, and
  Bray}]{csurka2004visual}
\bibinfo{author}{G.~Csurka}, \bibinfo{author}{C.~Dance},
  \bibinfo{author}{L.~Fan}, \bibinfo{author}{J.~Willamowski},
  \bibinfo{author}{C.~Bray},
\newblock \bibinfo{title}{Visual categorization with bags of keypoints},
\newblock in: \bibinfo{booktitle}{Workshop on statistical learning in computer
  vision, ECCV}, \bibinfo{number}{1-22}, \bibinfo{publisher}{Prague},
  \bibinfo{year}{2004}, pp. \bibinfo{pages}{1--2}.
\bibitem[{Perronnin and Dance(2007)}]{perronnin2007fisher}
\bibinfo{author}{F.~Perronnin}, \bibinfo{author}{C.~Dance},
\newblock \bibinfo{title}{Fisher kernels on visual vocabularies for image
  categorization},
\newblock in: \bibinfo{booktitle}{2007 IEEE Conference on Computer Vision and
  Pattern Recognition}, \bibinfo{organization}{IEEE}, pp.
  \bibinfo{pages}{1--8}.
\bibitem[{Cimpoi et~al.(2016)Cimpoi, Maji, Kokkinos, and
  Vedaldi}]{cimpoi2016deep}
\bibinfo{author}{M.~Cimpoi}, \bibinfo{author}{S.~Maji},
  \bibinfo{author}{I.~Kokkinos}, \bibinfo{author}{A.~Vedaldi},
\newblock \bibinfo{title}{Deep filter banks for texture recognition,
  description, and segmentation},
\newblock \bibinfo{journal}{International Journal of Computer Vision}
  \bibinfo{volume}{118} (\bibinfo{year}{2016}) \bibinfo{pages}{65--94}.
\bibitem[{Frosini(1992)}]{frosini1992measuring}
\bibinfo{author}{P.~Frosini},
\newblock \bibinfo{title}{Measuring shapes by size functions},
\newblock in: \bibinfo{booktitle}{Intelligent Robots and Computer Vision X:
  Algorithms and Techniques}, \bibinfo{publisher}{International Society for
  Optics and Photonics}, \bibinfo{year}{1992}, pp. \bibinfo{pages}{122--133}.
\bibitem[{Verri et~al.(1993)Verri, Uras, Frosini, and Ferri}]{Verri1993}
\bibinfo{author}{A.~Verri}, \bibinfo{author}{C.~Uras},
  \bibinfo{author}{P.~Frosini}, \bibinfo{author}{M.~Ferri},
\newblock \bibinfo{title}{On the use of size functions for shape analysis},
\newblock \bibinfo{journal}{Biological Cybernetics} \bibinfo{volume}{70}
  (\bibinfo{year}{1993}) \bibinfo{pages}{99--107}.
\bibitem[{Edelsbrunner et~al.(2002)Edelsbrunner, Letscher, and
  Zomorodian}]{EdLeZo2002}
\bibinfo{author}{H.~Edelsbrunner}, \bibinfo{author}{D.~Letscher},
  \bibinfo{author}{A.~Zomorodian},
\newblock \bibinfo{title}{Topological persistence and simplification},
\newblock \bibinfo{journal}{Discrete and Computational Geometry}
  \bibinfo{volume}{28} (\bibinfo{year}{2002}) \bibinfo{pages}{511--533}.
\bibitem[{Li et~al.(2014)Li, Ovsjanikov, and Chazal}]{li2014persistence}
\bibinfo{author}{C.~Li}, \bibinfo{author}{M.~Ovsjanikov},
  \bibinfo{author}{F.~Chazal},
\newblock \bibinfo{title}{Persistence-based structural recognition},
\newblock in: \bibinfo{booktitle}{IEEE Conference on Computer Vision and
  Pattern Recognition (CVPR)}, \bibinfo{publisher}{IEEE}, \bibinfo{year}{2014},
  pp. \bibinfo{pages}{2003--2010}.
\bibitem[{Reininghaus et~al.(2015)Reininghaus, Huber, Bauer, and
  Kwitt}]{reininghaus2014stable}
\bibinfo{author}{J.~Reininghaus}, \bibinfo{author}{S.~Huber},
  \bibinfo{author}{U.~Bauer}, \bibinfo{author}{R.~Kwitt},
\newblock \bibinfo{title}{A stable multi-scale kernel for topological machine
  learning},
\newblock in: \bibinfo{booktitle}{2015 IEEE Conference on Computer Vision and
  Pattern Recognition (CVPR)}, pp. \bibinfo{pages}{4741--4748}.
\bibitem[{Adams et~al.(2017)Adams, Emerson, Kirby, Neville, Peterson, Shipman,
  Chepushtanova, Hanson, Motta, and Ziegelmeier}]{adams2015persistent}
\bibinfo{author}{H.~Adams}, \bibinfo{author}{T.~Emerson},
  \bibinfo{author}{M.~Kirby}, \bibinfo{author}{R.~Neville},
  \bibinfo{author}{C.~Peterson}, \bibinfo{author}{P.~Shipman},
  \bibinfo{author}{S.~Chepushtanova}, \bibinfo{author}{E.~Hanson},
  \bibinfo{author}{F.~Motta}, \bibinfo{author}{L.~Ziegelmeier},
\newblock \bibinfo{title}{Persistence images: A stable vector representation of
  persistent homology},
\newblock \bibinfo{journal}{Journal of Machine Learning Research}
  \bibinfo{volume}{18} (\bibinfo{year}{2017}) \bibinfo{pages}{1--35}.
\bibitem[{Ferri et~al.(1998)Ferri, Frosini, Lovato, and Zambelli}]{Ferri1997}
\bibinfo{author}{M.~Ferri}, \bibinfo{author}{P.~Frosini},
  \bibinfo{author}{A.~Lovato}, \bibinfo{author}{C.~Zambelli},
\newblock \bibinfo{title}{Point selection: A new comparison scheme for size
  functions (with an application to monogram recognition)},
\newblock in: \bibinfo{booktitle}{Computer Vision --- ACCV'98: Third Asian
  Conference on Computer Vision Hong Kong, China, January 8--10, 1998
  Proceedings, Volume I}, \bibinfo{publisher}{Springer}, \bibinfo{year}{1998},
  pp. \bibinfo{pages}{329--337}.
\bibitem[{Donatini et~al.(1998)Donatini, Frosini, and Lovato}]{Frosini1998}
\bibinfo{author}{P.~Donatini}, \bibinfo{author}{P.~Frosini},
  \bibinfo{author}{A.~Lovato},
\newblock \bibinfo{title}{Size functions for signature recognition},
\newblock in: \bibinfo{booktitle}{Proc. SPIE}, volume \bibinfo{volume}{3454},
  pp. \bibinfo{pages}{178--183}.
\bibitem[{Poincar\'e(1890)}]{Poinc1890}
\bibinfo{author}{H.~J. Poincar\'e},
\newblock \bibinfo{title}{Sur le probleme des trois corps et les \'equations de
  la dynamique},
\newblock \bibinfo{journal}{Acta Mathematica} \bibinfo{volume}{13}
  (\bibinfo{year}{1890}) \bibinfo{pages}{1--270}.
\bibitem[{Poincar\'e(1899)}]{Poinc1899}
\bibinfo{author}{H.~J. Poincar\'e},
\newblock \bibinfo{title}{Les m\'ethodes nouvelles de la m\'ecanique
  c\'eleste},
\newblock \bibinfo{journal}{Gauthiers-Villars, Paris}  (\bibinfo{year}{1892,
  1893, 1899}).
\bibitem[{Poincar\'e(1895)}]{Poinc1895}
\bibinfo{author}{H.~J. Poincar\'e},
\newblock \bibinfo{title}{Analysis situs},
\newblock \bibinfo{journal}{J. \'Ec. Polytech., ser. 2} \bibinfo{volume}{1}
  (\bibinfo{year}{1895}) \bibinfo{pages}{1--123}.
\bibitem[{Bubenik(2015)}]{bubenik2015statistical}
\bibinfo{author}{P.~Bubenik},
\newblock \bibinfo{title}{Statistical topological data analysis using
  persistence landscapes.},
\newblock \bibinfo{journal}{Journal of Machine Learning Research}
  \bibinfo{volume}{16} (\bibinfo{year}{2015}) \bibinfo{pages}{77--102}.
\bibitem[{Makarenko et~al.(2016)Makarenko, Kalimoldayev, Pak, and
  Yessenaliyeva}]{makarenko2016texture}
\bibinfo{author}{N.~Makarenko}, \bibinfo{author}{M.~Kalimoldayev},
  \bibinfo{author}{I.~Pak}, \bibinfo{author}{A.~Yessenaliyeva},
\newblock \bibinfo{title}{Texture recognition by the methods of topological
  data analysis},
\newblock \bibinfo{journal}{Open Engineering} \bibinfo{volume}{6}
  (\bibinfo{year}{2016}).
\bibitem[{Schmid(2001)}]{schmid2001constructing}
\bibinfo{author}{C.~Schmid},
\newblock \bibinfo{title}{Constructing models for content-based image
  retrieval},
\newblock in: \bibinfo{booktitle}{Computer Vision and Pattern Recognition,
  2001. CVPR 2001. Proceedings of the 2001 IEEE Computer Society Conference
  on}, volume~\bibinfo{volume}{2}, \bibinfo{organization}{IEEE}, pp.
  \bibinfo{pages}{II--39}.
\bibitem[{Geusebroek et~al.(2003)Geusebroek, Smeulders, and Van~de
  Weijer}]{geusebroek2003fast}
\bibinfo{author}{J.-M. Geusebroek}, \bibinfo{author}{A.~W. Smeulders},
  \bibinfo{author}{J.~Van~de Weijer},
\newblock \bibinfo{title}{Fast anisotropic gauss filtering},
\newblock \bibinfo{journal}{IEEE Transactions on Image Processing}
  \bibinfo{volume}{12} (\bibinfo{year}{2003}) \bibinfo{pages}{938--943}.
\bibitem[{Guo et~al.(2010)Guo, Zhang, and Zhang}]{guo2010completed}
\bibinfo{author}{Z.~Guo}, \bibinfo{author}{L.~Zhang},
  \bibinfo{author}{D.~Zhang},
\newblock \bibinfo{title}{A completed modeling of local binary pattern operator
  for texture classification},
\newblock \bibinfo{journal}{IEEE Transactions on Image Processing}
  \bibinfo{volume}{19} (\bibinfo{year}{2010}) \bibinfo{pages}{1657--1663}.
\bibitem[{Cohen-Steiner et~al.(2007)Cohen-Steiner, Edelsbrunner, and
  Harer}]{Cohen-Steiner2007}
\bibinfo{author}{D.~Cohen-Steiner}, \bibinfo{author}{H.~Edelsbrunner},
  \bibinfo{author}{J.~Harer},
\newblock \bibinfo{title}{Stability of persistence diagrams},
\newblock \bibinfo{journal}{Discrete {\&} Computational Geometry}
  \bibinfo{volume}{37} (\bibinfo{year}{2007}) \bibinfo{pages}{103--120}.
\bibitem[{Cohen-Steiner et~al.(2010)Cohen-Steiner, Edelsbrunner, Harer, and
  Mileyko}]{Cohen-Steiner2010}
\bibinfo{author}{D.~Cohen-Steiner}, \bibinfo{author}{H.~Edelsbrunner},
  \bibinfo{author}{J.~Harer}, \bibinfo{author}{Y.~Mileyko},
\newblock \bibinfo{title}{Lipschitz functions have lp-stable persistence},
\newblock \bibinfo{journal}{Foundations of Computational Mathematics}
  \bibinfo{volume}{10} (\bibinfo{year}{2010}) \bibinfo{pages}{127--139}.
\bibitem[{Garcia and Stoll(1984)}]{garcia1984monte}
\bibinfo{author}{N.~Garcia}, \bibinfo{author}{E.~Stoll},
\newblock \bibinfo{title}{Monte carlo calculation for electromagnetic-wave
  scattering from random rough surfaces},
\newblock \bibinfo{journal}{Physical review letters} \bibinfo{volume}{52}
  (\bibinfo{year}{1984}) \bibinfo{pages}{1798}.
\bibitem[{Seiffert et~al.(2010)Seiffert, Khoshgoftaar, Van~Hulse, and
  Napolitano}]{seiffert2010rusboost}
\bibinfo{author}{C.~Seiffert}, \bibinfo{author}{T.~M. Khoshgoftaar},
  \bibinfo{author}{J.~Van~Hulse}, \bibinfo{author}{A.~Napolitano},
\newblock \bibinfo{title}{Rusboost: A hybrid approach to alleviating class
  imbalance},
\newblock \bibinfo{journal}{Systems, Man and Cybernetics, Part A: Systems and
  Humans, IEEE Transactions on} \bibinfo{volume}{40} (\bibinfo{year}{2010})
  \bibinfo{pages}{185--197}.
\bibitem[{Freund and Schapire(1997)}]{freund1997decision}
\bibinfo{author}{Y.~Freund}, \bibinfo{author}{R.~E. Schapire},
\newblock \bibinfo{title}{A decision-theoretic generalization of on-line
  learning and an application to boosting},
\newblock \bibinfo{journal}{Journal of computer and system sciences}
  \bibinfo{volume}{55} (\bibinfo{year}{1997}) \bibinfo{pages}{119--139}.
\bibitem[{Fumera and Roli(2002)}]{fumera2002cost}
\bibinfo{author}{G.~Fumera}, \bibinfo{author}{F.~Roli},
\newblock \bibinfo{title}{Cost-sensitive learning in support vector machines},
\newblock \bibinfo{journal}{VIII Convegno Associazione Italiana per
  L’Intelligenza Artificiale}  (\bibinfo{year}{2002}).
\bibitem[{Vedaldi and Fulkerson(2010)}]{vedaldi2010vlfeat}
\bibinfo{author}{A.~Vedaldi}, \bibinfo{author}{B.~Fulkerson},
\newblock \bibinfo{title}{Vlfeat: An open and portable library of computer
  vision algorithms},
\newblock in: \bibinfo{booktitle}{Proceedings of the International Conference
  on Multimedia}, \bibinfo{publisher}{ACM}, \bibinfo{year}{2010}, pp.
  \bibinfo{pages}{1469--1472}.
\bibitem[{Juda et~al.(2015)Juda, Mrozek, Brendel, Wagner, and et~al.}]{RedHom}
\bibinfo{author}{M.~Juda}, \bibinfo{author}{M.~Mrozek},
  \bibinfo{author}{P.~Brendel}, \bibinfo{author}{H.~Wagner},
  \bibinfo{author}{et~al.}, \bibinfo{title}{Capd::redhom},
  \bibinfo{howpublished}{http://redhom.ii.uj.edu.pl},
  \bibinfo{year}{2010-2015}.
\bibitem[{Juda and Mrozek(2014)}]{juda2014capd}
\bibinfo{author}{M.~Juda}, \bibinfo{author}{M.~Mrozek},
\newblock \bibinfo{title}{Capd:: Redhom v2-homology software based on reduction
  algorithms},
\newblock in: \bibinfo{booktitle}{Mathematical Software--ICMS 2014},
  \bibinfo{publisher}{Springer}, \bibinfo{year}{2014}, pp.
  \bibinfo{pages}{160--166}.
\bibitem[{Bauer et~al.(2013)Bauer, Kerber, and Reininghaus}]{url:PHAT}
\bibinfo{author}{U.~Bauer}, \bibinfo{author}{M.~Kerber},
  \bibinfo{author}{J.~Reininghaus}, \bibinfo{title}{Phat - persistent homology
  algorithms toolbox},
  \bibinfo{howpublished}{https://bitbucket.org/phat-code/phat},
  \bibinfo{year}{2013}.
\bibitem[{Bauer et~al.(2014)Bauer, Kerber, Reininghaus, and Wagner}]{PHAT}
\bibinfo{author}{U.~Bauer}, \bibinfo{author}{M.~Kerber},
  \bibinfo{author}{J.~Reininghaus}, \bibinfo{author}{H.~Wagner},
\newblock \bibinfo{title}{Phat - persistent homology algorithms toolbox},
\newblock in: \bibinfo{editor}{H.~Hong}, \bibinfo{editor}{C.~Yap} (Eds.),
  \bibinfo{booktitle}{Mathematical Software - ICMS 2014}, volume
  \bibinfo{volume}{8592} of \textit{\bibinfo{series}{Lecture Notes in Computer
  Science}}, \bibinfo{publisher}{Springer Berlin Heidelberg},
  \bibinfo{year}{2014}, pp. \bibinfo{pages}{137--143}.
\bibitem[{Zhenhua et~al.(2010)Zhenhua, Zhang, and Zhang}]{CLBPcode}
\bibinfo{author}{G.~Zhenhua}, \bibinfo{author}{L.~Zhang},
  \bibinfo{author}{D.~Zhang}, \bibinfo{title}{Clbp implementation for matlab},
  \bibinfo{howpublished}{\url{http://www.comp.polyu.edu.hk/~cslzhang/code/CLBP.rar}},
  \bibinfo{year}{2010}.
\bibitem[{Breiman(2001)}]{breiman2001random}
\bibinfo{author}{L.~Breiman},
\newblock \bibinfo{title}{Random forests},
\newblock \bibinfo{journal}{Machine learning} \bibinfo{volume}{45}
  (\bibinfo{year}{2001}) \bibinfo{pages}{5--32}.
\bibitem[{Duda et~al.(1973)Duda, Hart et~al.}]{duda1973pattern}
\bibinfo{author}{R.~O. Duda}, \bibinfo{author}{P.~E. Hart}, et~al.,
  \bibinfo{title}{Pattern classification and scene analysis},
  volume~\bibinfo{volume}{3}, \bibinfo{publisher}{Wiley New York},
  \bibinfo{year}{1973}.
\bibitem[{ISO-IEC(2002)}]{mpeg7standard}
\bibinfo{author}{ISO-IEC}, \bibinfo{title}{Information Technology - Multimedia
  Content Description Interface}, 15938, \bibinfo{publisher}{ISO/IEC, Moving
  Pictures Expert Group}, \bibinfo{edition}{1st} edition, \bibinfo{year}{2002}.
\bibitem[{Lowe(2004)}]{lowe2004distinctive}
\bibinfo{author}{D.~G. Lowe},
\newblock \bibinfo{title}{Distinctive image features from scale-invariant
  keypoints},
\newblock \bibinfo{journal}{International journal of computer vision}
  \bibinfo{volume}{60} (\bibinfo{year}{2004}) \bibinfo{pages}{91--110}.
\bibitem[{Ojala et~al.(1996)Ojala, Pietik{\"a}inen, and
  Harwood}]{ojala1996comparative}
\bibinfo{author}{T.~Ojala}, \bibinfo{author}{M.~Pietik{\"a}inen},
  \bibinfo{author}{D.~Harwood},
\newblock \bibinfo{title}{A comparative study of texture measures with
  classification based on featured distributions},
\newblock \bibinfo{journal}{Pattern recognition} \bibinfo{volume}{29}
  (\bibinfo{year}{1996}) \bibinfo{pages}{51--59}.
\bibitem[{Dalal and Triggs(2005)}]{dalal2005histograms}
\bibinfo{author}{N.~Dalal}, \bibinfo{author}{B.~Triggs},
\newblock \bibinfo{title}{Histograms of oriented gradients for human
  detection},
\newblock in: \bibinfo{booktitle}{Computer Vision and Pattern Recognition,
  2005. CVPR 2005. IEEE Computer Society Conference on},
  volume~\bibinfo{volume}{1}, \bibinfo{organization}{IEEE}, pp.
  \bibinfo{pages}{886--893}.
\bibitem[{Haralick et~al.(1973)Haralick, Shanmugam, and
  Dinstein}]{haralick1973textural}
\bibinfo{author}{R.~M. Haralick}, \bibinfo{author}{K.~Shanmugam},
  \bibinfo{author}{I.~H. Dinstein},
\newblock \bibinfo{title}{Textural features for image classification},
\newblock \bibinfo{journal}{Systems, Man and Cybernetics, IEEE Transactions on}
   (\bibinfo{year}{1973}) \bibinfo{pages}{610--621}.
\bibitem[{Vedaldi and Fulkerson(2008)}]{vedaldi08vlfeat}
\bibinfo{author}{A.~Vedaldi}, \bibinfo{author}{B.~Fulkerson},
  \bibinfo{title}{{VLFeat}: An open and portable library of computer vision
  algorithms}, \bibinfo{howpublished}{http://www.vlfeat.org/},
  \bibinfo{year}{2008}.
\bibitem[{Heikkilae and Ahonen(2014)}]{LBPcode}
\bibinfo{author}{M.~Heikkilae}, \bibinfo{author}{T.~Ahonen}, \bibinfo{title}{A
  general local binary pattern (lbp) implementation for matlab},
  \bibinfo{howpublished}{\url{http://www.cse.oulu.fi/CMV/Downloads/LBPMatlab}},
  \bibinfo{year}{2014}.
\bibitem[{Uppuluri(2008)}]{GLCMcode}
\bibinfo{author}{A.~Uppuluri}, \bibinfo{title}{Glcm texture features},
  \bibinfo{howpublished}{\url{http://de.mathworks.com/matlabcentral/fileexchange/22187-glcm-texture-features}},
  \bibinfo{year}{2008}.
\bibitem[{Long et~al.(2015)Long, Shelhamer, and Darrell}]{Long2015cvpr}
\bibinfo{author}{J.~Long}, \bibinfo{author}{E.~Shelhamer},
  \bibinfo{author}{T.~Darrell},
\newblock \bibinfo{title}{Fully convolutional networks for semantic
  segmentation},
\newblock in: \bibinfo{booktitle}{Proceedings of the International Conference
  on Computer Vision and Pattern Recognition (CVPR)}, pp.
  \bibinfo{pages}{3431--3440}.
\bibitem[{Nguyen et~al.(2015)Nguyen, Manzanera, and Kropatsch}]{Nguyen2015}
\bibinfo{author}{T.~P. Nguyen}, \bibinfo{author}{A.~Manzanera},
  \bibinfo{author}{W.~G. Kropatsch},
\newblock \bibinfo{title}{Impact of topology-related attributes from local
  binary patterns on texture classification},
\newblock in: \bibinfo{booktitle}{Computer Vision - ECCV 2014 Workshops:
  Zurich, Switzerland, September 6-7 and 12, 2014, Proceedings, Part II},
  \bibinfo{publisher}{Springer International Publishing}, \bibinfo{year}{2015},
  pp. \bibinfo{pages}{80--93}.

\end{thebibliography}

\end{document}